\documentclass[11pt,logo,copyright]{nvidiatechreport}
\usepackage{multicol}
\usepackage{amsmath}
\usepackage[numbers, sort&compress]{natbib}
\usepackage[skins, breakable]{tcolorbox}
\usepackage{colortbl}

\usepackage[ruled, vlined, linesnumbered]{algorithm2e}
\SetKwFor{ForPar}{for}{do in parallel}{end}
\usepackage{tikz}
\usetikzlibrary{positioning,shapes,arrows,patterns,calc}
\usetikzlibrary{shadows}
\usetikzlibrary{positioning, calc, patterns, backgrounds, decorations.pathmorphing, decorations.markings, fit}
\usetikzlibrary{arrows.meta, positioning, fit}
\usepackage{environ}

\newif\ifskiptikz

\ifskiptikz
  \RenewEnviron{tikzpicture}[1][]{\fbox{\scriptsize TikZ skipped}}
\fi
\usepackage{xcolor}

\usepackage{pgfplots}
\usepackage{subcaption}
\pgfplotsset{compat=1.17}
\pgfplotsset{every tick label/.append style={font=\scriptsize}}

\definecolor{Gray}{gray}{0.9}
\definecolor{lightgrey}{HTML}{f0f0f0} %
\definecolor{darkgreen}{RGB}{0,127,0}
\definecolor{darkred}{RGB}{200,0,0}
\definecolor{newgreen}{HTML}{55ff55} %
\definecolor{bargreen}{HTML}{9fffcb}
\definecolor{bargreenline}{HTML}{25a18e} %
\definecolor{barorange}{HTML}{888888} %
\definecolor{orangeding}{HTML}{ff8c00} %
\definecolor{barorangeline}{HTML}{888888}
\definecolor{bargreenlight4}{HTML}{bbffda}
\definecolor{bargreenlight2}{HTML}{d5ffe8}
\definecolor{bargreenlight3}{HTML}{e2ffef}
\definecolor{bargreenlight}{HTML}{ceffff}
\definecolor{lightgreen}{HTML}{e7f0eb}  %
\definecolor{lightgreenline}{HTML}{8fe5b6}
\definecolor{blockborder}{HTML}{B4B4B4} %
\definecolor{phaseblock}{HTML}{F2EDEB} %
\definecolor{simpleblock}{HTML}{FDF8FF} %
\definecolor{problemblock}{HTML}{FFEBE2} %
\definecolor{productblock}{HTML}{B6F094} %
\definecolor{solution2block}{HTML}{C9C9C9} %
\definecolor{solutionblock}{HTML}{ECFFBC} %
\definecolor{evalskipblock}{HTML}{FDC3C6} %
\newcommand{\bbox}[2]{\colorbox{#1!20}{$#2$}}

\usepackage{amssymb}
\DeclareFontFamily{U}{mathb}{\hyphenchar\font45}
\DeclareFontShape{U}{mathb}{m}{n}{
      <5> <6> <7> <8> <9> <10> gen * mathb
      <10.95> mathb10 <12> <14.4> <17.28> <20.74> <24.88> mathb12
      }{}
\DeclareSymbolFont{mathb}{U}{mathb}{m}{n}
\DeclareFontSubstitution{U}{mathb}{m}{n}
\let\tikzdot\dot
\let\dot\relax
\DeclareMathAccent{\dot}{0}{mathb}{"39}
\let\dddot\relax
\DeclareMathAccent{\dddot}{0}{mathb}{"3B}
\let\ddddot\relax
\DeclareMathAccent{\ddddot}{0}{mathb}{"3C}
\let\dot\tikzdot

\newcommand{\cmark}{\textcolor{darkgreen}{\ding{51}}} %
\newcommand{\xmark}{\textcolor{darkred}{\ding{55}}} %

\newcommand\numberthis[1]{\addtocounter{equation}{1}\tag{\theequation}\label{#1}}
\DeclareMathOperator*{\argmin}{arg\,min}

\newcommand{\cuRobo}{\textsf{cuRobo}}
\newcommand{\cuRoboVone}{\textsf{cuRoboV1}}
\newcommand{\project}{\textsf{cuRoboV2}} %
\newcommand{\nvblox}{\textsf{nvblox}}
\newcommand{\MoveIt}{\textsf{MoveIt}}

\newcommand{\mink}{\textsf{mink}}
\newcommand{\PyRoki}{\textsf{PyRoki}}
\newcommand{\Newton}{\textsf{Newton}}
\newcommand{\NewtonDyn}{\textsf{NewtonDyn}}
\newcommand{\GRiD}{\textsf{GRiD}}
\newcommand{\Warp}{\textsf{Warp}}
\newcommand{\CUDA}{\textsf{CUDA}}
\newcommand{\NumPy}{\textsf{NumPy}}
\newcommand{\PyTorch}{\textsf{PyTorch}}
\newcommand{\pybindeleven}{\textsf{pybind11}}
\newcommand{\Viser}{\textsf{Viser}}

\newcommand{\R}{\mathbb{R}}  %

\newcommand{\loss}{\mathcal L}

\usepackage{libertine}
\usepackage[libertine]{newtxmath}

\usepackage[format=plain,
            labelfont={bf},
            textfont=sf]{caption}

\newcommand{\figloc}[1]{\textit{#1}}
\definecolor{bg}{rgb}{0.95,0.95,0.95}
\usepackage{soul}
\sethlcolor{yellow}

\usepackage{titletoc}
\usetikzlibrary{patterns}
\usepackage{tabularx}
\usepackage{multicol}
\usepackage{multirow}
\usepackage{booktabs}
\usepackage{wrapfig}

\usepackage[color=green!20, textsize=footnotesize]{todonotes}

\definecolor{commentcolor}{HTML}{00e268}%

\SetKwComment{tcp}{$\triangleright$\ }{}
\SetCommentSty{mycommfont}

\usepackage{fancyhdr}
\usepackage{enumitem}

\title{cuRoboV2: Dynamics-Aware Motion Generation with Depth-Fused Distance Fields
for High-DoF Robots
}

\author{\normalsize  Balakumar Sundaralingam \quad Adithyavairavan Murali  \quad Stan Birchfield}
\pgfplotsset{compat=1.18}
\begin{abstract}
Effective robot autonomy requires motion generation that is safe, feasible, and reactive.
Current methods are fragmented: fast planners output physically unexecutable trajectories, reactive controllers struggle with high-fidelity perception, and existing solvers fail on high-DoF systems.
We present \project{}, a unified framework with three key innovations: (1) B-spline trajectory optimization that enforces smoothness and torque limits;
(2) a GPU-native TSDF/ESDF perception pipeline that generates \emph{dense} signed distance fields covering the full workspace, unlike existing methods that only provide distances within sparsely allocated blocks, up to 10$\times$ faster and in 8$\times$ less memory than the state-of-the-art at manipulation scale, with up to 99\% collision recall;
and (3) scalable GPU-native whole-body computation, namely topology-aware kinematics, differentiable inverse dynamics, and map-reduce self-collision, that achieves up to 61$\times$ speedup while also extending to high-DoF humanoids (where previous GPU implementations fail).
On benchmarks, \project{} achieves 99.7\% success under 3\,kg payload (where baselines achieve only 72--77\%), 99.6\% collision-free IK on a 48-DoF humanoid (where prior methods fail entirely), and 89.5\% retargeting constraint satisfaction (\emph{vs.}\ 61\% for \PyRoki{}); these collision-free motions yield locomotion policies with 21\% lower tracking error than \PyRoki{} and 12$\times$ lower cross-seed variance than GMR.
A ground-up codebase redesign for discoverability enabled LLM coding assistants to author up to 73\% of new modules, including hand-optimized \CUDA{} kernels, demonstrating that well-structured robotics code can unlock productive human--LLM collaboration.
Together, these advances provide a unified, dynamics-aware motion generation stack that scales from single-arm manipulators to full humanoids. Code is available at \url{https://github.com/NVlabs/curobo}.
\end{abstract}

\makeatletter
\renewcommand{\fps@figure}{!tbh}
\renewcommand{\fps@table}{!tbh}
\makeatother

\begin{document}
\hypersetup{colorlinks=true,allcolors=nvidiagreen,pdfborder={0 0 0}}

\maketitle
\abscontent

\clearpage
{\small
\tableofcontents
}
\clearpage

\section{Introduction}
\label{sec:intro}
Effective robot autonomy requires motion generation to navigate and interact with an environment safely, efficiently, and responsively. Research on this problem has historically been split into two sub-domains: (1) Global Motion Planning, which computes global paths from a static start state to a goal~\cite{thomason2024vamp,curobo_icra23,mishani2025srmp}, and (2) Reactive Motion Generation, which performs high-frequency tracking of a moving target~\cite{capt_2024, storm2021,dalal2024neural, fishman2022mpinets}.
While these tasks differ in scope, they share the same fundamental requirements of satisfying robot hardware limits while avoiding collisions (both with the scene and with the robot itself).
In both domains, however, current methods struggle to meet these requirements simultaneously, leaving critical gaps in functionality.
What prevents existing methods from achieving unified, high-performance autonomy?  We identify three fundamental barriers:

\paragraph*{1) The Feasibility Gap} While collision-free motion planning can be done quickly~\cite{curobo_icra23,capt_2024}, such speed often comes at the cost of ignoring dynamics, leading to physically unexecutable motions. Most planners output piecewise-linear joint paths that assume infinite torque, ignoring the robot's mass and momentum. Consequently, ``time-optimal'' plans often violate torque limits, especially under heavy payloads, requiring aggressive post-processing that invalidates the original plan's safety guarantees. Conversely, dynamic trajectory optimization methods respect these limits but struggle to scale to the non-convex constraints of mesh or depth-based collision avoidance.

\paragraph*{2) The Perception-Reactivity Trade-off} Current reactive methods force a choice between safety and high-fidelity perception. Analytic controllers (e.g., RMPs \cite{cheng2018rmp, Ratliff2018RiemannianMP}, MPC \cite{williams2017mppi, storm2021}) provide safety guarantees but are often too slow to process raw depth data, restricting them to simplified geometric primitives. Conversely, learning-based approaches~\cite{danielczuk2021object,murali2023cabinet, fishman2022mpinets} process visual observations rapidly, but they fail to provide the strict collision guarantees required for safe deployment. Furthermore, they lack the ability to generalize, thus requiring extensive retraining for new environments.

\paragraph*{3) The Scalability Wall} Methods that succeed for single-arm platforms often fail to scale when applied to high-DoF systems like bimanual manipulators or humanoids. We find that state-of-the-art planners~\cite{curobo_icra23} converge slowly or not at all in these higher-dimensional spaces, particularly for collision-free inverse kinematics (IK) in cluttered scenes, an important requirement for real-world bimanual manipulation and humanoid motion retargeting.

\paragraph{Contributions} To overcome these barriers, we present \project{}, a unified motion generation framework with three core algorithmic contributions:

\begin{enumerate}

\item \textbf{B-Spline Optimization Formulation} addresses the \textit{feasibility gap}. By optimizing B-spline control points, our approach automatically enforces smoothness, allowing the solver to satisfy torque limits for global planning and handle non-static boundary conditions for reactive control, thus providing a unified representation for motion generation.

\item \textbf{GPU-native Perception Pipeline} solves the \textit{perception-reactivity trade-off}. Diverse inputs (depth, meshes, cuboids) are fused into a millimeter-resolution block-sparse TSDF using a voxel-centric projection strategy that eliminates atomic contention. Unlike existing GPU libraries such as \nvblox{}~\cite{nvblox}, which only compute distances within sparsely allocated blocks and leave the rest of the workspace without distance information, we generate a \emph{dense} ESDF covering the full workspace on-demand via the Parallel Banding Algorithm (PBA+) with a gather-based seeding stage that enables full CUDA graph capture. This provides $O(1)$ distance queries at any point in the workspace, up to 10$\times$ faster and in 8$\times$ less memory than \nvblox{} at manipulation scale with up to 99\% collision recall, enabling millisecond-rate updates for reactive control.

\item \textbf{Scalable Kinematics, Dynamics \& Self-Collision} handles the \textit{scalability wall}. We design GPU-native building blocks, namely topology-aware kinematics with sparse Jacobian computation, differentiable inverse dynamics, and map-reduce self-collision, that scale to humanoids where prior GPU implementations fail, with up to 40$\times$ speedup. This throughput accelerates optimization iterations, enabling solvers to converge quickly for high-DoF bimanual and humanoid robots while enforcing torque limits.
\end{enumerate}
Together, these GPU-native innovations enable unified, dynamics-aware motion generation that scales from single-arm manipulators to full humanoids across planning, reactive control, and retargeting.
The remainder of the paper is organized as follows. Sec.~\ref{sec:related-work} surveys related work, and Sec.~\ref{sec:problem} formulates the motion optimization problem. We then present the three core modules: B-spline trajectory optimization (Sec.~\ref{sec:bspline}), TSDF/ESDF perception pipeline (Sec.~\ref{sec:esdf}), and scalable kinematics, dynamics, and self-collision for high-DoF robots (Sec.~\ref{sec:high_dof_engine}). Sec.~\ref{sec:results} evaluates these contributions on benchmarks and real-world manipulation, and Sec.~\ref{sec:llm_dev} reports on LLM-assisted development enabled by a ground-up codebase redesign. We conclude in Sec.~\ref{sec:conclusion}.

\begin{table}
    \centering
    \caption{Feature comparison of \project{} with other motion generation frameworks.}
    \label{tab:features_tools}
    {
    \small %
    \setlength{\tabcolsep}{5pt} %
    \begin{tabular}{l c >{\columncolor{lightgrey}}c c c >{\columncolor{lightgrey}}c}
    \toprule
    & \MoveIt{} & \cuRobo{} & \mink{} & \PyRoki{} & \project{} \\
    \toprule
    \multicolumn{6}{l}{\textit{Collision Checking}} \\
    \quad Primitive & \cmark & \cmark & \cmark &  \cmark & \cmark \\
    \quad Mesh & \cmark & \cmark & \cmark &  \xmark & \cmark \\
    \quad Depth & \cmark & \cmark & \xmark &  \xmark & \cmark \\ \midrule
    \multicolumn{6}{l}{\textit{Optimization}} \\
    \quad Trajectory Space & Position & Position & --- & Position & \textcolor{darkgreen}{B-Spline} \\
    \quad Custom Costs & \xmark & \,\,\,\cmark$^{1}$ & \cmark & \cmark & \cmark \\
    \quad Kinematic Jacobian & \xmark & \xmark & \cmark & \cmark & \cmark \\
    \quad Center of Mass & \xmark & \xmark & \cmark & \xmark & \cmark \\
    \quad Torque-Limits & \xmark & \xmark & \xmark & \xmark & \cmark \\
    \midrule
    \multicolumn{6}{l}{\textit{Numerical Solvers}} \\
    \quad QP & \xmark & \xmark & \cmark & \xmark & \xmark \\
    \quad NLLS (LM) & \xmark & \xmark & \xmark & \cmark & \cmark \\
    \quad Quasi-Newton (L-BFGS) & \xmark & \cmark & \xmark & \xmark & \cmark \\
    \quad Particle-based (MPPI) & \xmark & \cmark & \xmark & \xmark & \cmark \\
    \midrule
    \multicolumn{6}{l}{\textit{Collision-Aware Plan Modes}} \\
    \quad Graph/Roadmap Planner &  \cmark & \cmark & - & \xmark & \cmark \\
    \quad Trajectory Optimization & Single-Arm & Single-Arm & - & Single-Arm$^3$ & \textcolor{darkgreen}{Whole-Body} \\
    \quad Global Motion Plan & \cmark & Single-Arm & - & \xmark & \textcolor{darkgreen}{Whole-Body}\\
    \quad MPC & \xmark & \,\,Single-Arm$^{2}$ & - &  \xmark & \textcolor{darkgreen}{Whole-Body} \\
    \midrule
    \multicolumn{6}{l}{\textit{IK Modes}} \\
    \quad Local-IK & Single-Arm & Single-Arm & \textcolor{darkgreen}{Whole-Body} & \textcolor{darkgreen}{Whole-Body} & \textcolor{darkgreen}{Whole-Body} \\
    \quad \textit{w/ Collision Avoidance} & Single-Arm & Single-Arm & \textcolor{darkgreen}{Whole-Body} & Single-Arm$^3$ & \textcolor{darkgreen}{Whole-Body} \\
    \addlinespace
    \quad Global-IK & Single-Arm & Single-Arm & \xmark & \textcolor{darkgreen}{Whole-Body} & \textcolor{darkgreen}{Whole-Body} \\
    \quad \textit{w/ Collision Avoidance} & Single-Arm & Single-Arm & \xmark & Single-Arm$^3$ & \textcolor{darkgreen}{Whole-Body} \\
    \addlinespace
    \quad Humanoid Retargeting & \xmark & \xmark & \textcolor{darkgreen}{Whole-Body} & \textcolor{darkgreen}{Whole-Body} & \textcolor{darkgreen}{Whole-Body}\\
    \quad \textit{w/ Collision Avoidance} & \xmark & \xmark & \textcolor{darkgreen}{Whole-Body} & \xmark$^3$ & \textcolor{darkgreen}{Whole-Body} \\
    \midrule
    \multicolumn{6}{l}{\textit{Perception}} \\
    \quad Extrinsic Camera Calib. & \cmark & \xmark & - & \xmark & \cmark \\
    \quad Voxel/ESDF Integrator & Voxel & \xmark & - & \xmark & \textcolor{darkgreen}{ESDF} \\ \midrule
    \multicolumn{6}{l}{\textit{Usability \& Integration}} \\
    \quad Installation & ros & \textcolor{darkred}{compile} & \textcolor{darkgreen}{pip} & \textcolor{darkgreen}{pip} & \textcolor{darkgreen}{pip} \\
    \quad Startup Time & \textcolor{darkgreen}{$<5s$} & \textcolor{darkgreen}{$<5s$} & \textcolor{darkgreen}{$<5s$} & \textcolor{darkred}{$>30s$} & \textcolor{darkgreen}{$<5s$} \\
    \quad Easy \& Accurate Robot Import & \cmark & \xmark & \cmark & \xmark & \cmark \\ \midrule
    Batched Scene Interface & \xmark & \cmark & \xmark & \xmark & \cmark \\
    \bottomrule
    \end{tabular}
    }
    {\footnotesize
    \bigskip

    \begin{tabular}{l p{0.9\columnwidth}}
          $[1]$ & Custom cost terms can be implemented in \cuRobo{} as shown by \cite{chen2025bodex}, but
          it's fairly complex and requires extensive working knowledge of the framework.\\
          $[2]$ & \cuRobo{} implements MPPI~\citep{storm2021}, providing reactivity but not high-quality motions that can be obtained with L-BFGS.\\
          $[3]$ & \PyRoki{} claims features for Collision-Free planning, but fails to converge on Global-IK problems and is also extremely slow due to an inefficient collision checking implementation as we show in Sec.~\ref{sec:results}.
    \end{tabular}
    }
\end{table}

\section{Related Work}
\label{sec:related-work}

\paragraph{Global Planning} Motion planning algorithms fall into search, sampling, and optimization-based methods. \textit{Search-based} planners~\cite{Hart1968AFB, Likhachev2003ARAAA, mishani2025srmp} discretize the state space but struggle with high-dimensional spaces. \textit{Sampling-based} planners~\cite{LaValle1998RapidlyexploringRT, Karaman2011SamplingbasedAF, Gammell2015BatchIT, RobotMotionPlanningChipRSS2016} are probabilistically complete but cannot easily accommodate cost terms and require post-processing that may violate dynamics constraints.
\textit{Trajectory optimization}~\cite{ratliff2009chomp, kalakrishnan2011stomp, Schulman2014MotionPW} jointly optimizes smoothness, collision avoidance, and custom costs but parameterizes trajectories as joint-position waypoints, relying on acceleration regularization for smoothness. STOMP~\cite{kalakrishnan2011stomp} was among the earliest to incorporate torque constraints into trajectory optimization, but this capability was never ported beyond ROS\,1 and is absent from modern frameworks. As a local optimizer rather than a global planner, STOMP achieves only 60\% kinematic success on our benchmark even without dynamics constraints. B-splines offer an alternative, used for trajectory smoothing~\cite{pan2011collision} and path planning for mobile~\cite{chiaravalli2018bspline} and aerial~\cite{koyuncu2008bspline} robots. Existing B-spline formulations for manipulators are restricted to convex constraints like polyhedral obstacles~\cite{ji2023convex, drake} and do not address non-convex constraints from depth-image-based collision representations, triangle meshes, or torque limits.
\project{} introduces a B-spline formulation for trajectory optimization for high-DoF manipulators that can accommodate additional constraints like torque limits without sacrificing convergence.

\paragraph{Reactive Motion Generation} Practical applications require reacting to dynamic environments from visual observations. \textit{Model-based} approaches like potential fields~\cite{khatib1986real}, Riemannian motion policies~\cite{Ratliff2018RiemannianMP, cheng2018rmp}, geometric fabrics~\cite{VanWyk2022GeometricFG}, and MPC~\cite{storm2021, williams2017mppi} provide safety guarantees but get stuck in local optima and require simplified geometric primitives. \textit{Learning-based} methods~\cite{fishman2022mpinets, dalal2024neuralmp, fishman2025avoideverything, huang2024diffusionseeder, yang2025deepreactive} distill classical planners into neural policies or warm-start optimizers~\cite{ichnowski2020deep, yoon2023learning, Ichter2018LearningSD, Chamzas2021LearningSD}; RL has also trained multi-arm reaching policies~\cite{lai2025roboballet}. However, these methods lack out-of-distribution generalization, require per-embodiment retraining, and cannot guarantee collision-free execution.

\paragraph{Hardware Acceleration} FPGAs/ASICs~\cite{RobotMotionPlanningChipRSS2016, wan2021fpga, neuman2021robomorphic} and SIMD~\cite{thomason2024vamp} accelerate sampling-based planning. GPUs have been applied to MPC~\cite{storm2021, hyatt2020parameterized} and optimization~\cite{ichter2017group, curobo_icra23}. \cuRobo{}~\cite{curobo_icra23} introduced GPU-parallelized trajectory optimization and collision-aware IK, while \PyRoki{}~\cite{kim2025pyroki} provides cross-platform kinematic optimization. These methods target single-arm robots; scaling to high-DoF robots degrades convergence and increases compute time.

\paragraph{Visual Representations} \textit{Geometry-based} methods like KinectFusion~\cite{KinectFusion2011} represent scenes as TSDFs but lack collision queries for planning.
\nvblox{}~\cite{nvblox} provides GPU-accelerated ESDF for gradient-based planning but only computes ESDF in allocated blocks rather than a dense field, and incurs high memory overhead at millimeter resolution. CPU-based methods~\cite{oleynikova2017voxblox, fiesta} lack real-time capability.
\textit{Neural} representations~\cite{Danielczuk2021ObjectRU, kew2019neural, murali2023cabinet} enable fast collision checking, while learned encoders~\cite{Qi2017PointNetDH, huang2024diffusionseeder, fishman2022mpinets, dalal2024neuralmp, fishman2025avoideverything, yang2025deepreactive} often fail on out-of-distribution scenes. \project{} contributes a GPU-native perception pipeline with millimeter-resolution collision queries for manipulation.

\paragraph{Human-to-Humanoid Motion Retargeting} Transferring human motion to humanoids is critical for locomotion policy training. End-to-end learning approaches~\cite{he2024h2o, he2025omnih2o, fu2024humanplus, ji2024exbody2} train RL policies directly from retargeted human motion data, filtering infeasible motions during training but relying on the policy to implicitly handle self-collision and joint limits. Optimization-based retargeting frameworks such as GMR~\cite{joao2025gmr} use \mink{}~\cite{Zakka_Mink_Python_inverse_2025} for per-frame local IK with only joint-limit constraints. \PyRoki{}~\cite{kim2025pyroki} supports global collision-free IK but fails to converge on high-DoF robots. Consequently, retargeted motions often contain self-penetrations and joint-limit violations that propagate into policy training, causing instability and high cross-seed variance. \project{} addresses this gap by providing collision-free, joint-limit-satisfying IK for high-DoF robots, producing feasible reference motions that directly improve policy quality without changes to the learning pipeline.

\section{Problem Formulation}
\label{sec:problem}
We formalize motion generation as a trajectory optimization problem, where the goal is to find a sequence of actions that drives the robot from a start state to a goal pose (or, more specifically, to target poses for one or more links). The resulting trajectory should minimize an objective function (typically related to motion smoothness and energy) while satisfying a set of hard constraints such as avoiding self-collisions and collisions with the environment, as well as respecting the robot's kinematic and dynamic limits.

Let $\theta_t \in \R^d$ be the joint angles at time $t$, and let $\Theta_t = [\theta_t, \dot{\theta}_t, \ddot{\theta}_t, \dddot{\theta}_t]$ be the robot's state (angles, velocities, acceleration, and jerk), where $d$ is the number of joints.  Given an initial state $\Theta_0$ and one or more link goal poses $X_g^m \in {\it{\text{SE}}}(3)$, $m=1,\ldots,M$, we seek a trajectory $U$ that solves the following optimization problem,\footnote{Green highlights components where we introduce new formulations. Notation: $\odot$ is the Hadamard product.}
\begin{align*}
    \argmin_{U} \quad &  \sum_{t=0}^T \gamma_{s} \underbrace{||\ddot{\theta}_t||^2}_\text{smooth}   + \gamma_\ell \underbrace{||\dot{\theta}_t||^2}_\text{length} +
    \bbox{newgreen}{\gamma_e \underbrace{||(\dot{\theta}_t \odot \tau_t) dt||^2}_\text{energy}}  \numberthis{eq:cost_motion_opt}\\
\text{s.t.} \quad
& \bbox{newgreen}{\Theta_t = \operatorname{BsplineEval}(U, \Theta_0, t)} \numberthis{eq:forward_interp}\\
& \Theta^- \preceq \Theta_t \preceq \Theta^+ \numberthis{eq:joint-limit}\\
& \bbox{newgreen}{\mathbf{C}_\text{scene}(\operatorname{FwdKin}(\theta_t)) \preceq 0}\numberthis{eq:scene-cost}\\
& \bbox{newgreen}{\mathbf{C}_\text{robot}(\operatorname{FwdKin}(\theta_t)) \preceq 0} \numberthis{eq:self-cost}\\
& \bbox{newgreen}{||X_g^m - \operatorname{FwdKin}(\theta_T)|| < \eta_m} \numberthis{eq:goal}\\
& \bbox{newgreen}{\tau_t = \operatorname{InvDyn}(\Theta_t)} \numberthis{eq:inverse-dynamics}\\
& \bbox{newgreen}{\tau^- \preceq \tau_t \preceq \tau^+} \numberthis{eq:torque-limit}\\
& \dot{\theta}_T, \ddot{\theta}_T = 0 \numberthis{eq:stop-constraint}
\end{align*}
where the loss in Eq.~(\ref{eq:cost_motion_opt}) encourages smooth and efficient motion; Eq.~(\ref{eq:forward_interp}) evaluates the B-spline at a certain time; Eq.~(\ref{eq:joint-limit}) enforces limits on joint angles, velocities, \emph{etc.}; Eqs.~(\ref{eq:scene-cost}) and (\ref{eq:self-cost}) ensure a collision-free path with both the scene and the robot itself (via forward kinematics); Eq.~(\ref{eq:goal}) ensures the final state is close to the target poses, where $\eta_m$ is a threshold; Eq.~(\ref{eq:inverse-dynamics}) computes the torques $\tau_t \in \R^d$ via inverse dynamics; Eq.~(\ref{eq:torque-limit}) ensures torque limits $\tau^-$ and $\tau^+$ are respected; and Eq.~(\ref{eq:stop-constraint}) ensures that the robot comes to a complete stop.  Constraints in Eqs.~(\ref{eq:forward_interp})--(\ref{eq:torque-limit}) are applied to all timesteps $t$ and links $m$.

This formulation aligns with prior trajectory optimization methods~\cite{ratliff2009chomp,curobo_icra23} and subsumes several core robotics problems.
Variations of this optimization problem have been used primarily for obtaining collision-free trajectories,
starting with local trajectory optimization~\cite{ratliff2009chomp,schulman2014motion} which uses a rough initial trajectory as a seed and locally converges to a collision-free trajectory.
\citet{curobo_icra23} generalized this to global collision-free motion planning by running many seeds of trajectory optimization in parallel and also leveraging a collision-free graph-based planner as a seed.
Special cases include inverse kinematics, where a single state is optimized without temporal constraints. Re-solving the optimization after a short execution horizon yields reactive control strategies such as model predictive control (MPC).

\begin{figure}
    \centering
    \includegraphics{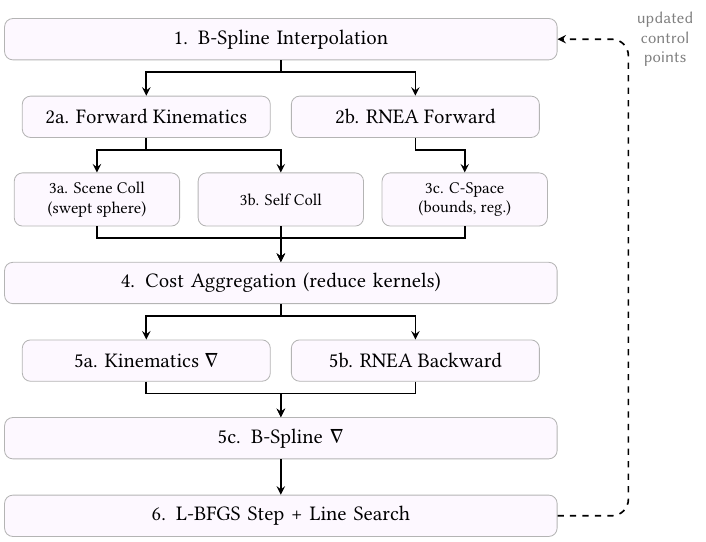}
    \caption[Trajectory optimization loop overview]{Trajectory optimization loop. Each iteration evaluates B-spline waypoints, computes kinematics and inverse dynamics in parallel, evaluates costs (scene collision, self-collision, configuration-space bounds) concurrently on separate \CUDA{} streams, aggregates them, backpropagates gradients through all forward operations, and updates the B-spline control points via L-BFGS.}
    \label{fig:trajopt-flow}
\end{figure}

Figure~\ref{fig:trajopt-flow} illustrates the trajectory optimization loop. Each iteration consists of six steps: (1) B-spline interpolation generates trajectory waypoints from control points; (2) forward kinematics computes link poses and Jacobians, with Recursive Newton-Euler Algorithm (RNEA) additionally computing joint torques; (3) cost evaluation runs in parallel for world collision, self-collision, and state costs; (4) cost aggregation reduces per-trajectory costs; (5) the backward pass computes gradients through all forward operations; and (6) the optimizer updates control points.

\section{B-Spline Basis as Optimization Space}
\label{sec:bspline}

\begin{figure}
  \centering
  \includegraphics[width=0.7\linewidth]{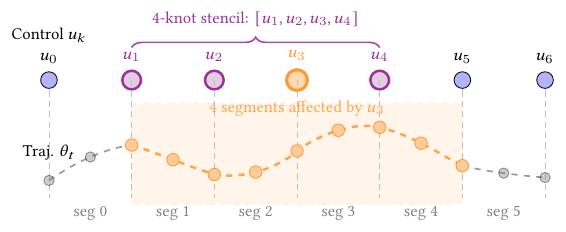}
  \caption[Local support property of cubic B-splines]{Local support property of cubic B-splines. Perturbing knot $u_3$ only affects 4 neighboring curve segments (orange region). Interpolation points (dots) are uniformly spaced in time; during optimization, each point within this region contributes a gradient to $u_3$ (accumulated with GPU warp-level reductions). (For simplicity, $N_{\text{interp}}=2$ is shown.)}
  \label{fig:bspline_local_support}
\end{figure}

Each joint trajectory $U \in \R^{d \times K}$ is represented as a uniform cubic B-spline over control points $u_k \in \R, k=0,\ldots,K-1$ for each of the $d$ joints; these control points serve as the optimization variables in Eq.~\eqref{eq:forward_interp}.
B-splines of degree three and higher yield $C^2$-continuous trajectories, and their local support ensures that changing a single knot only influences a few neighboring segments as shown in Fig.~\ref{fig:bspline_local_support}.
As a result, this choice leads to a basis that is both smooth and compact.
This implicit smoothness allows us to satisfy the acceleration and velocity regularizers in Eq.~\eqref{eq:cost_motion_opt} with far fewer decision variables than a typical per-timestep position parameterization, while keeping the problem well-conditioned for gradient-based solvers.

Given the control points $u_k, k=0,\ldots,K-1$ for a particular joint, the spline is evaluated within a segment (between consecutive knots) at a point specified by an interpolation parameter $\alpha \in [0, 1]$.
We sample at uniform time intervals, with each segment containing the same number of points, ensuring velocity, acceleration, and jerk are computed at consistent timesteps.
For the cubic case, the state vector~$\Theta_t=[\theta_t, \dot{\theta}_t, \ddot{\theta}_t, \dddot{\theta}_t]^\top$ for a segment with control points
$U_k=[u_{k-1}, u_k, u_{k+1}, u_{k+2}]^\top$ is given by
\begin{align*}
\Theta_t
&=
\underbrace{\mathbf{T}\mathbf{P}(\alpha)\mathbf{C}}_{\mathbf{B}(\alpha)} U_k,
\numberthis{eq:bspline-cubic}
\end{align*}
where $\mathbf{T} = \operatorname{diag}\!\left(1, dt_u^{-1}, dt_u^{-2}, dt_u^{-3}\right)$ encodes time scaling with $dt_u\in\R$ being a fixed time interval,
\begin{equation}
\mathbf{P}(\alpha) =
\begin{bmatrix}
\alpha^3   & \alpha^2 & \alpha & 1 \\
3\alpha^2 & 2\alpha  & 1      & 0 \\
6\alpha   & 2        & 0      & 0 \\
6         & 0        & 0      & 0
\end{bmatrix}
\end{equation}
evaluates cubic monomials and their derivatives at $\alpha$ (each row differentiates the preceding row), and where the cubic B-spline coefficients are given by
\begin{equation}
\mathbf{C} =
\frac{1}{6}
\begin{bmatrix}
-1 & 3 & -3 & 1 \\
 3 & -6 & 3 & 0 \\
 -3 & 0 & 3 & 0 \\
 1 & 4 & 1 & 0
\end{bmatrix}
\end{equation}

\subsection{Gradients with Respect to Spline Control Points}
Solving the optimization problem requires $\nabla_U \loss$, where $\loss$ is the loss in Eq.~(1). By the chain rule, the per-sample gradient contribution for a local control point vector $U_k$ is given by
\begin{align}
    \frac{\partial \loss}{\partial U_k} = \mathbf{B}(\alpha)^\top g,
    \label{eq:backward_grad}
\end{align}
where $g = \frac{\partial \loss}{\partial \Theta_t}$ is the upstream gradient on the state. The backward pass is simply the transpose of the forward pass in Eq.~\eqref{eq:bspline-cubic}, with time scaling handled implicitly by $\mathbf{T}$ inside $\mathbf{B}(\alpha)$.
Since a control point $u_k$ influences multiple segments (Fig.~\ref{fig:bspline_local_support}), its total gradient sums contributions from all relevant $N_{\text{interp}}$ interpolation points,
\begin{align}
    \nabla_{u_k} \loss = \sum_{i=1}^{N_{\text{interp}}} \frac{\partial \loss}{\partial u_{k,i}},
\end{align}
where we set $N_{\text{interp}}=4$ in our implementation.
Since this summation dominates runtime, we implement it as a highly parallel reduction on the GPU. Our kernel assigns threads to knots in an interleaved pattern (Fig.~\ref{fig:bspline_local_support}): each thread accumulates a local sum, and a warp-level reduction produces the final gradient. This design fuses computation and reduction into a single pass, avoiding intermediate writes to global memory.

\subsection{Boundary Conditions}
A key advantage of the spline parameterization is that it satisfies boundary states implicitly, thus eliminating constraint coupling and the oscillatory corrections that typically follow.
Static goals require only that we repeat the final knot four times, which automatically satisfies Eq.~\eqref{eq:stop-constraint}.
When the initial state $\Theta_0$ is non-static, we introduce three fictitious knots with corresponding ghost/virtual control points $u_{-3}, u_{-2}, u_{-1}$, computed from Eq.~\eqref{eq:bspline-cubic} at $\alpha=0$ with $\dddot{\theta}_0=0$. By anchoring these knots to the initial state, we eliminate conflicting boundary constraints and let the optimizer concentrate on the remaining objectives,
\begin{align*}
    u_{-3} &= -\frac{1}{6} \ddot{\theta}_0 dt_u^2 + \theta_0, \numberthis{eq:fixed_3} \\
    u_{-2} &= \frac{1}{3} \ddot{\theta}_0 dt_u^2 + \theta_0 + \dot{\theta}_0 dt_u, \numberthis{eq:fixed_2}\\
    u_{-1} &= \frac{11}{6} \ddot{\theta}_0 dt_u^2 + \theta_0 + 2 \dot{\theta}_0 dt_u. \numberthis{eq:fixed_1}
\end{align*}

\section{ESDF for Scene Collision Avoidance}
\label{sec:esdf}
\begin{figure}
  \centering
  \includegraphics{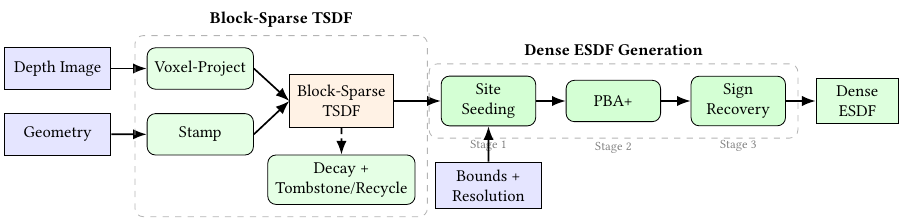}
  \caption[TSDF-ESDF pipeline overview]{TSDF-ESDF pipeline. Depth images are fused into a block-sparse TSDF via voxel-centric projection (Sec.~\ref{sec:tsdf_integration}), and known geometry is stamped analytically. After each update, TSDF weights are decayed (time + frustum factors), and blocks below a weight threshold are tombstoned and recycled. On demand, a dense ESDF is generated in three stages: (1) seeding surface sites from zero-crossings, (2) propagating distances via PBA+, and
  (3) recovering signs for geometry-layer voxels beyond the truncation band.}
  \label{fig:tsdf_esdf_pipeline}
\end{figure}

Collision query is a key computational bottleneck in motion generation. Recent methods reduce this cost by approximating the robot as spheres~\cite{curobo_icra23,thomason2024vamp,capt_2024}, converting signed distance to point queries. However, these approaches face two challenges: (1) fusing depth observations and known geometric primitives into a persistent world model, and (2) generating task-specific distance fields efficiently.
We address these difficulties with a two-stage representation as shown in Fig.~\ref{fig:tsdf_esdf_pipeline}:
(1) a block-sparse TSDF that fuses depth and primitives into a persistent world model via lock-free integration kernels (Sec.~\ref{sec:sparse_tsdf}); and (2) a dense ESDF generated on-demand at task-appropriate resolution with interior-sign recovery for watertight geometry (Sec.~\ref{sec:esdf_gen}).

\subsection{Block-Sparse TSDF Storage}
\label{sec:sparse_tsdf}

\begin{figure}
    \centering
    \includegraphics{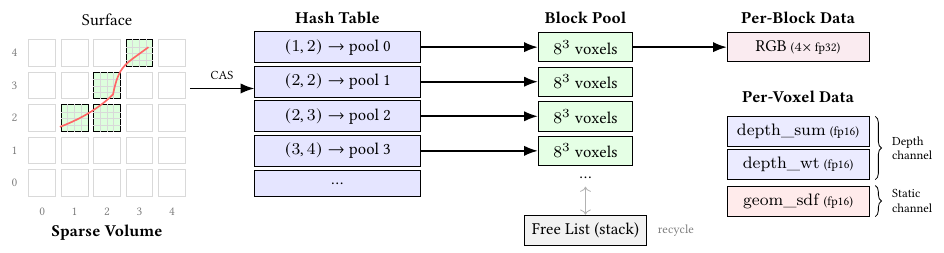}
    \caption[Block-sparse TSDF storage with dual-channel voxels]{Block-sparse TSDF storage. Only blocks near observed surfaces are allocated. A hash table maps block coordinates to pool indices via CAS. Each voxel stores two independent float16 channels (depth and geometry) whose minimum is returned at query time. Recycled blocks are managed through a free-list stack.}
    \label{fig:block_sparse_storage}
\end{figure}

For many common tasks, the TSDF needs to represent millimeter-resolution geometry over meter-scale environments. We use a block-sparse representation (Fig.~\ref{fig:block_sparse_storage}), partitioning the volume into $8^3$-voxel blocks and allocating only those near observed surfaces. A hash table maps block coordinates to pool indices~\cite{niessner2013real}, with Compare-And-Swap (CAS) handling concurrent insertions. Blocks whose voxel weights decay below a threshold are returned to a free list for reuse, bounding peak memory regardless of scene duration.

Each voxel maintains two independent signed-distance channels stored in float16: a running weighted average (\code{depth\_sum}, \code{depth\_wt}) for depth observations (4 bytes per voxel) and a \texttt{geom\_sdf} for analytic primitives (2 bytes per voxel).
At query time, the effective signed distance is $\min(\text{depth}, \text{geom})$, so depth-fused surfaces and known geometry contribute jointly to collision avoidance.
For a typical indoor scene with 100K active blocks, total GPU memory is ${\sim}350$\,MB, roughly 11$\times$ less than the equivalent dense grid.

\subsection{Depth and Primitive Integration}
\label{sec:tsdf_integration}

\begin{figure}
    \centering
    \includegraphics[width=0.98\linewidth]{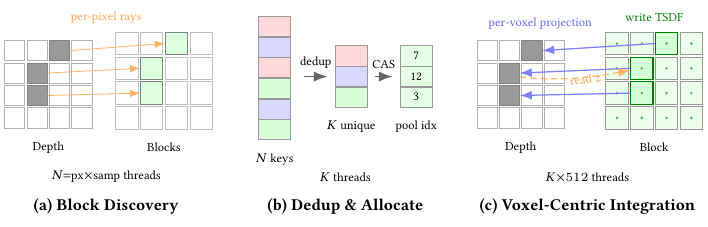}
    \caption[Voxel-Project depth integration pipeline]{Voxel-Project depth integration. (a)~Per-pixel rays discover blocks touched by the current depth frame. (b)~Duplicate keys are filtered and surviving blocks are allocated via CAS, yielding $K$ pool indices. (c)~Phase~4 reverses the mapping: each voxel projects itself into the image, reads the depth at the projected pixel, and writes the signed distance directly, eliminating all atomic contention.}
    \label{fig:voxel_project_pipeline}
\end{figure}

Depth integration uses a \emph{Voxel-Project} strategy (Fig.~\ref{fig:voxel_project_pipeline}).
The core idea is to organize GPU work around \emph{voxels}: each voxel is assigned its own thread, projects itself into the depth image, and writes its signed distance in a single, contention-free store.
The pipeline proceeds in four phases:
\begin{enumerate}[leftmargin=*,itemsep=1pt,topsep=2pt]
    \item \textbf{Block discovery.} For each pixel, ray samples within the truncation band identify which voxel blocks are touched by the current depth frame, launching $N{=}\text{pixels}{\times}\text{samples}$ threads.
    \item \textbf{Deduplication.} Keys from pixels with no valid depth (e.g., out-of-range or missing readings) are discarded and the remainder are deduplicated, yielding only the $K$ unique block keys visible in this frame.
    \item \textbf{Block allocation.} Each unique key is inserted into the hash table via compare and swap (CAS), reusing recycled blocks from the free list when available. The output is a compact array of pool indices for the visible blocks.
    \item \textbf{Voxel-centric integration.} One thread is launched per voxel across all visible blocks ($K \times 512$ threads). Each thread computes its voxel center in world coordinates, projects it onto the image plane using the camera intrinsics, reads the depth at the projected pixel, and writes the signed distance and weight directly. Because each voxel is owned by exactly one thread, no atomic operations are required.
\end{enumerate}

Organizing work around voxels yields coalesced memory writes (adjacent threads write adjacent voxels) and avoids hash-table lookups in the integration kernel, since pool indices are already resolved in Phase~3.
The trade-off is that Phase~4 has a data-dependent launch dimension ($K{\times}512$), requiring a device-to-host synchronization to read $K$.
Because multiple image pixels may map to the same nearby voxel, we scale the integration weight by the projected voxel area in pixels, $c = (f_x \, v / z)(f_y \, v / z)$, where $f_x, f_y$ are the focal lengths, $v$ is the voxel size, and $z$ is the camera-frame depth. The per-observation weight is $w=\max(c,1)$, ensuring at least unit weight.

Cuboids and meshes are stamped directly into the geometry channel.
For each primitive, we compute its axis-aligned bounding box in voxel coordinates and select candidate blocks whose analytical SDF falls within the truncation band.
After deduplication and hash-table allocation, we update each voxel with the minimum signed distance to the primitive.

\subsection{Frustum-Aware Decay and Block Recycling}

To track dynamic scenes, we apply a two-tier multiplicative weight decay after each integration step. A \emph{time decay} factor $\alpha_t$ (e.g., $0.99$) is applied to every allocated block, gradually fading old observations. A \emph{frustum decay} factor $\alpha_f$ (e.g., $0.5$) is applied additionally to blocks within the current camera frustum, enabling rapid adaptation to objects that move or disappear in view. The combined per-frame weight update is
\begin{equation}
w \leftarrow
\begin{cases}
w \cdot \alpha_t \cdot \alpha_f, & \text{block in frustum}, \\
w \cdot \alpha_t, & \text{otherwise}.
\end{cases}
\end{equation}
Frustum membership is determined at block granularity using a conservative bounding-sphere test (one thread per block), achieving $\pm$half-block precision. The decay itself is a single fused multiplication over the block data tensor, requiring zero atomic operations. After decay, the per-block weight sum is computed via a PyTorch reduction; blocks whose total weight falls below a threshold are recycled by tombstoning their hash slot and pushing the block index onto a free list for reuse.

\subsection{On-Demand ESDF with Sign Recovery}
\label{sec:esdf_gen}
For a robot to interact and navigate in the world without collision, it needs to be able to query signed distance to the nearest surface from any point in space.
This requirement motivates the need for a dense ESDF over the workspace which, due to memory limitations, leads to storing the ESDF at a coarse resolution.
Furthermore, since the robot is approximated by collision spheres, refining the ESDF well beyond the smallest sphere radius yields diminishing returns: the trilinear interpolation error is already a small fraction of the sphere radius, so collision checks remain reliable. This motivates generating the ESDF at a task-specific resolution coarser than that of the TSDF. For manipulation with fine spheres (1--5\,cm), a 5\,mm ESDF suffices; for mobile-base planning with coarser spheres (e.g., 50\,cm) over larger workspaces, an even coarser ESDF is appropriate. This decoupling also benefits performance: a finer TSDF avoids atomic contention during integration, while a coarser ESDF reduces memory and computation. Given workspace bounds and resolution, we generate a dense ESDF in three stages, enabling $O(1)$ trilinear distance queries.

\begin{figure*}
    \centering
    \begin{minipage}[c]{0.45\textwidth}
        \begin{subfigure}[t]{\textwidth}
            \centering
            \includegraphics[width=\linewidth]{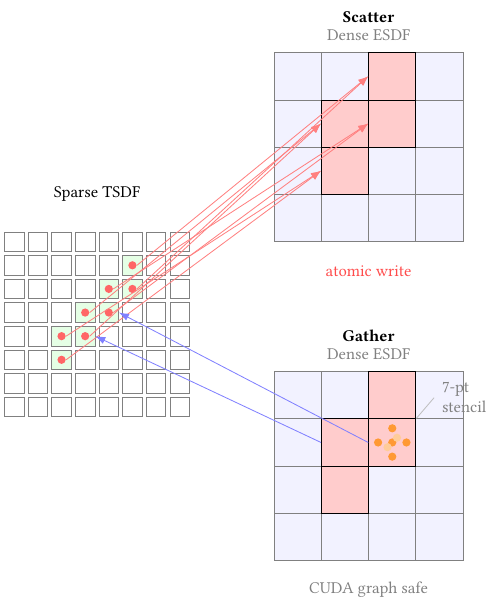}
            \caption[Scatter vs.\ gather site seeding]{Site seeding: scatter (top) vs.\ gather (bottom).}
            \label{fig:esdf_site_seeding}
        \end{subfigure}
    \end{minipage}%
    \hfill
    \begin{minipage}[c]{0.52\textwidth}
        \begin{subfigure}[t]{\textwidth}
            \centering
            \includegraphics[width=\linewidth]{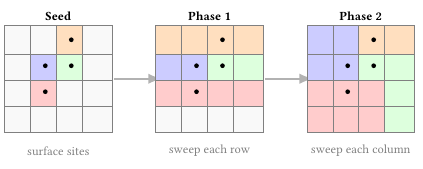}
            \caption[PBA+ distance propagation in 2D]{PBA+ propagation (2D): Phase~1 sweeps rows, Phase~2 sweeps columns to produce the exact Voronoi diagram.}
            \label{fig:esdf_pba}
        \end{subfigure}

        \vspace{0.5em}

        \begin{subfigure}[t]{\textwidth}
            \centering
            \includegraphics[width=0.9\linewidth]{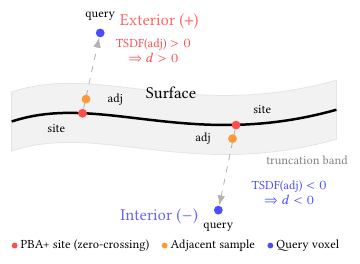}
            \caption[ESDF sign recovery via TSDF sampling]{Sign recovery via adjacent TSDF sampling.}
            \label{fig:esdf_sign_recovery}
        \end{subfigure}
    \end{minipage}
    \caption[ESDF generation pipeline with seeding, propagation, and sign recovery]{ESDF generation pipeline. (a)~Site seeding: \emph{scatter} launches one thread per TSDF voxel and writes to the ESDF (requires atomics); \emph{gather} launches one thread per ESDF voxel and probes the TSDF via a 7-point stencil (no atomics, CUDA graph safe). (b)~PBA+ propagates site ownership via separable sweeps: Phase~1 sweeps rows via bidirectional flooding, Phase~2 builds Maurer stacks along columns to produce the exact Voronoi diagram; in 3D a third phase reuses the same kernels on transposed data to resolve the remaining axis. (c)~For voxels beyond the truncation band, an adjacent TSDF sample resolves interior vs.\ exterior sign.}
    \label{fig:esdf_pipeline}
\end{figure*}

\paragraph{Stage 1: Site Seeding}
We identify \emph{surface sites}, i.e., voxels near zero-crossings in the sparse TSDF, that seed the distance transform (Fig.~\ref{fig:esdf_site_seeding}). This sparse-to-dense transfer is the computational bottleneck of ESDF generation, as the sparse TSDF must be read and the dense ESDF written for every surface voxel.

Two strategies exist for this transfer: \emph{scatter} and \emph{gather} (Fig.~\ref{fig:esdf_site_seeding}).
\emph{Scatter} iterates over allocated TSDF blocks ($O(B \cdot 8^3)$ work, where $B$ is the number of allocated blocks), projects each surface voxel (with $|\text{sdf}| < 0.9\,v_{\text{tsdf}}$) into ESDF coordinates, and writes the site.
When the ESDF is coarser than the TSDF, multiple TSDF voxels map to the same ESDF cell, requiring atomic writes to resolve write conflicts.
Moreover, the work dimension depends on $B$, which varies per frame, preventing CUDA graph capture.

\emph{Gather} reverses the mapping: each ESDF cell probes the sparse TSDF via $O(1)$ hash lookups to determine whether it contains a surface. We sample seven positions per cell (the center and six axis-aligned face centers, each offset by $\tfrac{1}{2}v_{\text{esdf}}$) and mark the cell as a site if any sample satisfies $|\text{sdf}| < 0.9\,v_{\text{tsdf}}$.
Since each cell is written by exactly one thread, no atomic operations are needed and memory writes are fully coalesced.
The work dimension is fixed at $|\mathcal{G}_{\text{esdf}}| = D{\times}H{\times}W$, independent of scene content, making the kernel compatible with CUDA graph capture and enabling the entire ESDF pipeline (seed $\to$ PBA+ $\to$ sign recovery) to run as a single captured graph.
The trade-off is coverage: when $v_{\text{esdf}} \gg v_{\text{tsdf}}$, thin surfaces falling between the seven sample points may be missed; in practice, the 7-point stencil covers the cell volume adequately for typical resolution ratios ($2$--$4\times$).

\paragraph{Stage 2: Distance Propagation}
We compute exact nearest-site assignments using the Parallel Banding Algorithm (PBA+)~\cite{cao2010parallel}, which exploits the separability of squared Euclidean distance to decompose the 3D Voronoi diagram into three independent axis-aligned passes (Fig.~\ref{fig:esdf_pba}).
Phase~1 solves the 1D nearest-site problem along each Z-column via a bidirectional flood: a forward sweep propagates the most recently seen site, and a backward sweep keeps whichever of the forward or backward candidate is closer.
The resulting partial Voronoi diagram is then extended to 2D and 3D by Phases~2 and~3, which sweep along the Y and X axes respectively.

Each phase applies Maurer's parabola intersection test~\cite{maurer2003linear}: given candidate sites $A,B,C$ ordered along the sweep axis, site $B$ is \emph{dominated} and discarded whenever the intersection of the distance paraboloids $d^2(A,\cdot)$ and $d^2(C,\cdot)$ lies before $B$'s position.
Building this compressed stack in a single linear pass per column and then walking it to assign the nearest site to every voxel yields an $O(N)$-work, $O(1)$-error distance transform per axis.

Phase~3 reuses the identical sweep and coloring logic on axis-transposed data, so only three unique kernels are needed.
The entire propagation produces the \emph{exact} Euclidean Voronoi diagram, unlike approximate iterative methods such as JFA~\cite{rong2006jump}, which require $\lceil \log_2 N \rceil + 3$ passes and can miss sites in adversarial configurations.
Because the number of kernel launches is fixed (five) and independent of scene content, the pipeline is fully compatible with CUDA graph capture. After propagation, the Euclidean distance at each voxel is computed from the stored site coordinates.

\paragraph{Stage 3: Sign Recovery}
PBA+ yields unsigned distances; the challenge is recovering sign for voxels beyond the TSDF truncation band (Fig.~\ref{fig:esdf_sign_recovery}). For depth-based TSDF, we assume positive sign (exterior) outside the truncation band, since depth observations only see the front surface. For watertight geometry (analytic primitives), surface sites lie at zero-crossings where the TSDF is ambiguous; we resolve this by sampling an \emph{adjacent} voxel, offset from the site toward the query along the site-to-query direction. Because this adjacent point lies on the same side of the surface as the query, its TSDF sign is unambiguous: negative indicates interior and positive indicates exterior.
Crucially, the adjacent lookup samples only the \emph{static} (geometry) SDF channel, since only watertight primitives have reliable sign when stepping away from the surface; depth-fused surfaces are open and their sign beyond the truncation band is undefined.
If the adjacent lookup misses (e.g., outside allocated blocks), we fall back to the combined SDF at the query voxel itself; voxels with no observation default to positive (exterior).

Once we have the ESDF, we calculate the collision constraint as an approximate swept-volume checks between timesteps, stepping along the trajectory by signed distance. The cost uses the CHOMP speed metric~\citep{ratliff2009chomp}, scaling gradients by velocity to accelerate escape from collision.

\section{Scaling to High-DoF Robots}
\label{sec:high_dof_engine}

\begin{figure}
  \centering
  \begin{minipage}[b]{0.23\textwidth}
      \centering
      \includegraphics{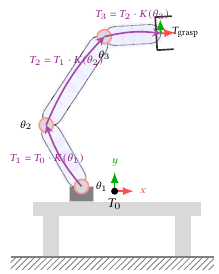}
      \subcaption{Serial chain}
      \label{fig:fk_single_arm}
  \end{minipage}%
  \hfill
  \begin{minipage}[b]{0.22\textwidth}
      \centering
      \includegraphics{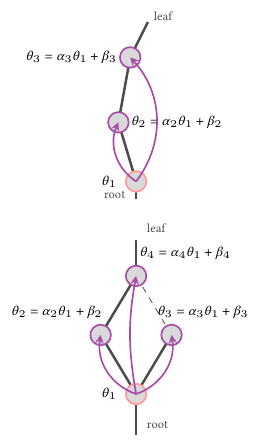}
      \subcaption{Mimic Joints}
      \label{fig:mimic_joint}
  \end{minipage}%
  \hfill
  \begin{minipage}[b]{0.5\textwidth}
      \centering
      \includegraphics{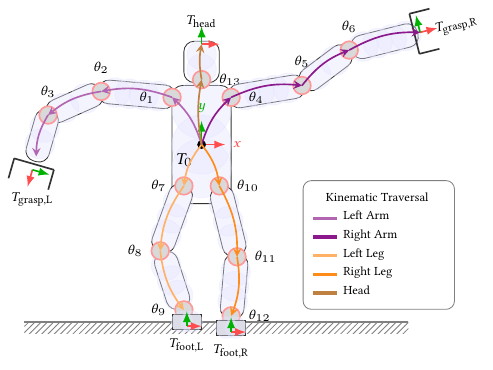}
      \subcaption{Branching tree of a humanoid}
      \label{fig:fk_humanoid}
  \end{minipage}

  \caption[Complex Kinematic Structures]{Kinematic structures that challenge parallelized computation.
  (a) A single-arm serial chain, the traditional case.
  (b) Mimic joints, where actuating $\theta_1$ mechanically constrains $\theta_2$ and $\theta_3$. (c) A humanoid with branching kinematic chains. We introduce precomputed topology caches for $O(1)$ ancestor lookups and two-stage Jacobian filtering to efficiently parallelize gradient and Jacobian computation.}
  \label{fig:fk_comparison}
\end{figure}

Scaling motion optimization to high-DoF robots such as humanoids introduces challenges absent in simpler manipulators. In particular, \emph{branching kinematic trees} create multiple end-effectors whose Jacobians must be computed in parallel (Fig.~\ref{fig:fk_comparison}); and \emph{self-collision pairs} grow quadratically with the number of links, making naive pairwise queries memory-bound.  Furthermore, enforcing \emph{actuator torque limits} requires differentiable inverse dynamics within the optimization loop; and \emph{mimic joints} mechanically couple multiple links through a single actuator.

We address these challenges with five key innovations:
(1)~an adaptive kernel dispatch for forward kinematics that scales from fused single-kernel execution to a dual-kernel split for complex robots (Sec.~\ref{subsec:kinematicsimprovfwdkin});
(2)~a precomputed topology cache that enables parallel gradient backpropagation by exploiting kinematic sparsity (Sec.~\ref{subsec:kinematicsimprovbp});
(3)~a two-stage filtering scheme that computes sparse Jacobians for branching trees and mimic joints (Sec.~\ref{subsec:kinematicsimprovjac});
(4)~a fused map-reduce kernel for self-collision that converts memory-bound pairwise queries into compute-bound reductions (Sec.~\ref{sec:self_collision});
and (5)~a differentiable inverse dynamics engine that enforces torque limits directly within optimization (Sec.~\ref{sec:torque_approach}).

\subsection{Kinematics Improvements}
\label{sec:kinematicsimprov}

\subsubsection{Forward Kinematics}
\label{subsec:kinematicsimprovfwdkin}
\begin{figure}
  \centering
  \includegraphics{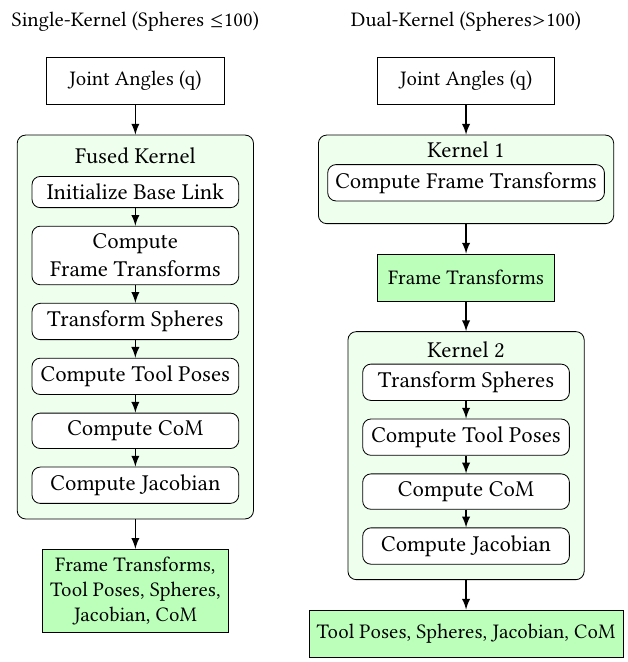}
  \caption[Adaptive Forward Kinematics Execution]{Adaptive forward kinematics execution. For simple robots, a single fused kernel computes frame transforms, sphere positions, and Jacobians. For complex robots, this is split into two kernels: the first computes frame transforms, and the second computes sphere positions and Jacobians.}
  \label{fig:fk_kernels}
\end{figure}

A single forward kinematics call must produce frame transforms, tool poses, collision-sphere positions, centers of mass, and Jacobians. \cuRobo{}~\cite{curobo_icra23} introduced a fused single-kernel approach that computes frame transforms and sphere positions in one launch, but it parallelizes across only four threads and lacks support for center of mass and Jacobian computation. We extend this with an adaptive kernel dispatch based on robot complexity (Fig.~\ref{fig:fk_kernels}). For simple robots with few collision spheres ($\leq 100$), the fused kernel now computes all five outputs within the same four-thread design. For complex robots with many spheres ($>100$), such as bimanual and humanoids, the computation splits into two kernels (Fig.~\ref{fig:fk_kernels}): the first computes frame transforms from joint angles and writes them to global memory; the second reads these transforms and computes sphere positions, tool poses, centers of mass, and Jacobians in parallel across many threads.
In contrast to the fused kernel, where all stages are serialized across four threads, this split allows the second kernel to scale thread count with the number of spheres and links, greatly reducing the compute time.

\subsubsection{Parallel Gradient Backpropagation}
\label{subsec:kinematicsimprovbp}
The adaptive kernel dispatch described above accelerates forward kinematics, but backpropagating gradients to joints remains a bottleneck: each link's gradient must be pushed back through all ancestor joints. We parallelize this at the link level by partitioning links across threads within a GPU warp (Algorithm~\ref{alg:backward_grad}). A gradient on a humanoid's right hand, for instance, only affects the right arm and torso joints. To exploit this kinematic sparsity, we use a precomputed topology cache that provides the ordered list of ancestor links for any link in $O(1)$ time, allowing each thread to traverse only the relevant kinematic chain. This data-driven design also implicitly handles mimic joints by mapping each link's gradient to the correct actuated joint.

\begin{algorithm}
    \small
    \caption{Backprop with Topology Cache}
    \label{alg:backward_grad}
    \KwIn{Link gradients $\nabla_{\mathbf{p}_\ell} \loss, \nabla_{\mathbf{q}_\ell} \loss$; Frame transforms $\mathbf{T}_\ell$}
    \KwOut{Joint gradients $\nabla_{\mathbf{q}} \loss$}
    $\texttt{local\_grad} \gets \mathbf{0}$ \tcp*{init thread-local gradient for all joints}
    \For{each link $\ell$ (parallel threads in warp)}{
        \If{link gradient is non-zero }{
            \For(\tcp*[f]{\textcolor{darkgreen}{loop only through topology cache}}){each ancestor $j_{\text{link}} \in \texttt{linkChain}[\ell]$ \textbf{in reverse}}
            {
                $j \gets \texttt{jointMap}[j_{\text{link}}]$ \tcp*{\textcolor{darkgreen}{handle mimic joints}}
                $g \gets B$($\nabla_{\mathbf{p}_\ell} \loss$, $\mathbf{T}_\ell$) \tcp*{projecting link $\ell$ gradient using frame transform and joint model}
                $\texttt{local\_grad}[j] \mathrel{+}= g$ \tcp*{\textcolor{darkgreen}{adding to account for mimic joints}}
            }
        }
    }
     $\nabla_{\mathbf{q}} \gets \texttt{WarpReduceSum}(\texttt{local\_grad})$\;
\end{algorithm}

\subsubsection{Sparse Jacobian Computation}
\label{subsec:kinematicsimprovjac}
For solvers like Levenberg-Marquardt, we compute the full kinematic Jacobian by parallelizing each column (Algorithm~\ref{alg:jacobian_compute}). Each thread, responsible for a single joint $j$, uses a two-stage filtering process. First, a precomputed $\texttt{affects}[j, e]$ cache provides an $O(1)$ coarse check to prune entire subtrees that cannot affect the target link $e$. For the remaining joints, including mimic joints that control multiple links, a second fine-grained filter checks if each mechanically coupled link is a direct ancestor of $e$. This ensures each Jacobian column accumulates contributions only from relevant links in the kinematic chain.

\begin{algorithm}
  \small
  \caption{Jacobian for Trees and Mimic Joints}
  \label{alg:jacobian_compute}
  \KwIn{Joint angles $\mathbf{q}$, target link index $e$, frame transforms $\mathbf{T}_\ell$}
  \KwOut{Jacobian matrix $\mathbf{J}^e$}
  \For(\tcp*[f]{compute column per joint}){each joint $j \in \{1, \ldots, n_{\text{joints}}\}$ \textbf{in parallel}}
  {
      $\mathbf{jac\_col} \gets \mathbf{0}$\;
      \If(\tcp*[f]{\textcolor{darkgreen}{Coarse filter: does joint $j$ affect target link $e$?}})
      {$\texttt{affects}[j, e]$}
      {
          \For(\tcp*[f]{\textcolor{darkgreen}{A joint can control multiple links (e.g., mimic)}}){each link $\ell \in \texttt{connectedLinks}[j]$}{
              \If(\tcp*[f]{\textcolor{darkgreen}{Fine filter: is this specific link in chain to $e$?}}){$\ell \in \texttt{linkChain}[e]$}{
                  $\mathbf{jac\_col}$ += Compute Jacobian contribution for $\ell$\;
              }
          }
      }
      $\mathbf{J}^e[:, j] \gets \mathbf{jac\_col}$ \tcp*{write to global memory}

  }
\end{algorithm}

Our sparse Jacobian (Alg.~\ref{alg:jacobian_compute}) enables a Levenberg-Marquardt (LM) solver for inverse kinematics (IK),
a natural fit since IK is inherently a nonlinear least-squares problem, minimizing the squared pose error $\|\mathbf{r}\|^2$ with residual $\mathbf{r} = \mathbf{T}_{goal} \ominus \operatorname{FwdKin}(\mathbf{q})$.
 LM approximates the Hessian as $\mathbf{J}^\top \mathbf{J}$ and uses trust-region damping for global robustness (Alg.~\ref{alg:lm}), converging faster than first-order methods such as L-BFGS on this least-squares structure. We adapt \Newton{}'s~\cite{newton} \Warp{}-based~\cite{warp} LM implementation to use our Jacobian, supporting high-DoF robots with multiple tool frames.

For collision-free IK, we use a two-stage strategy: LM first converges to the target pose without collision constraints, then L-BFGS refines the solution under self-collision and environment constraints. LM provides a good seed near the goal, so L-BFGS only needs small adjustments to resolve collisions rather than searching from scratch.

\begin{algorithm}
  \small
  \SetAlgoNoLine\SetAlgoNoEnd
  \caption{Levenberg-Marquardt with Trust-Region}
  \label{alg:lm}
  \KwIn{Initial state $S_k=(\mathbf q_k,\lambda_k,\mathbf J_k,\mathbf g_k,E_k)$, goals $\mathbf T_{goal}$}
  \KwOut{Updated state $S_{k+1}$}
  $\mathbf H\leftarrow \mathbf J_k^\top \mathbf J_k$ \tcp*{Approximate the Hessian}
  $\mathbf A\leftarrow \mathbf H+\lambda_k\mathbf I$ \tcp*{Apply damping to the Hessian}
  Solve $\mathbf A\boldsymbol\delta=-\mathbf g_k$ for $\boldsymbol\delta$ \tcp*{Compute the update step}
  $\mathbf q^+\leftarrow \mathbf q_k+\boldsymbol\delta$\tcp*{Apply update to joint configuration}
  Evaluate at $\mathbf q^+$ to get: $\mathbf r^+,E^+,\mathbf J^+,\mathbf g^+$\tcp*{Evaluate error at new configuration}
  $\rho_{\text{pred}}\leftarrow \tfrac12\boldsymbol\delta^\top(\lambda_k\boldsymbol\delta-\mathbf g_k)$\tcp*{Predict the error reduction}
  $\rho_{\text{trust}}\leftarrow (E_k-E^+)/(\rho_{\text{pred}}+\epsilon)$\tcp*{Compute trust region ratio}
  \eIf{$\rho_{\text{trust}} \ge \rho_{\min}$}
  {
      $\lambda_{k+1}\leftarrow \mathrm{clamp}\big(\lambda_k/\gamma,\lambda_{\min},\lambda_{\max}\big)$\tcp*{Accept step, decrease damping}
      $(\mathbf q_{k+1}, E_{k+1}, \dots)\leftarrow (\mathbf q^+, E^+, \dots)$
  }{
      $\lambda_{k+1} \leftarrow \mathrm{clamp}\big(\lambda_k\gamma,\lambda_{\min},\lambda_{\max}\big)$\tcp*{Reject step, increase damping}
      $(\mathbf q_{k+1}, E_{k+1}, \dots) \leftarrow (\mathbf q_k, E_k, \dots)$
  }
  Check for convergence against position and orientation tolerances\tcp*{Verify if goal is reached}
  \Return{$S_{k+1}$}
\end{algorithm}

\begin{figure}
  \centering
  \resizebox{0.8\textwidth}{!}{
  \includegraphics[width=0.7\textwidth]{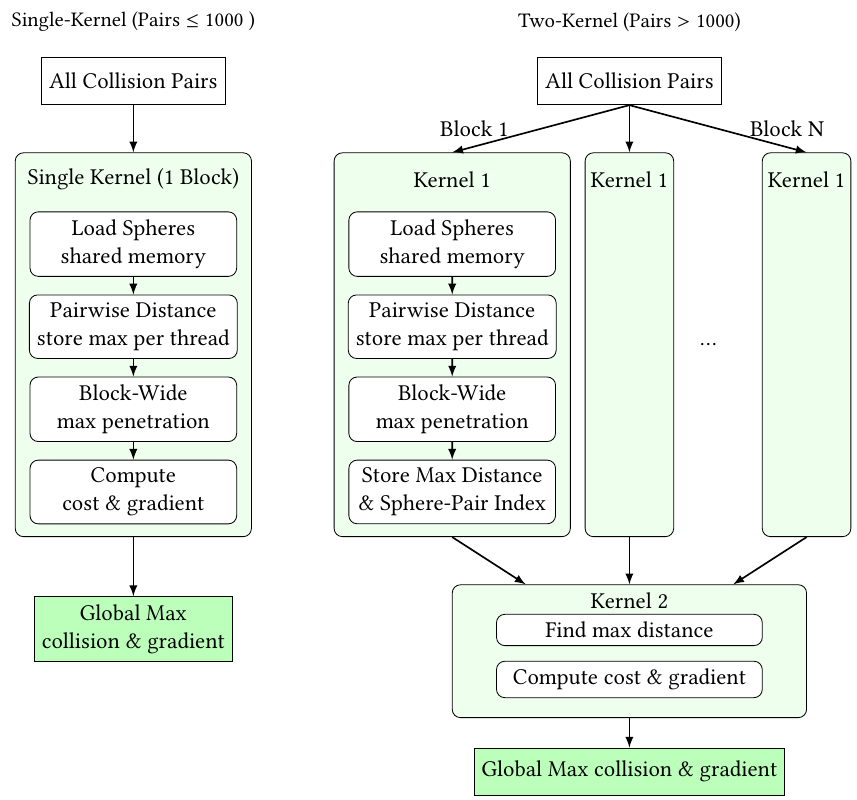}
  }
  \caption[Two-Stage Self-Collision Checking]{Two-stage self-collision checking. For complex robots, collision pairs are partitioned across GPU blocks (map). A first kernel finds local maxima, and a second kernel reduces these to find the global maximum.}
  \label{fig:self_collision_reduction}
\end{figure}

\subsection{Self-Collision via Map-Reduce}
\label{sec:self_collision}
\cuRobo{}~\cite{curobo_icra23} computes self-collision cost within a single thread block by loading spheres into shared memory. For high-DoF robots, however, the number of pairs grows quadratically, from 818 pairs on a 7-DoF Franka to 162k pairs on a 48-DoF humanoid, forcing each thread to loop over many pairs and serializing the computation. We address this with a two-stage map-reduce: (1) partition pairs across blocks, each loading its spheres into shared memory and reducing to find its most-penetrating pair; (2) a final kernel reduces across block maxima to identify the global maximum as shown in Fig.~\ref{fig:self_collision_reduction}.

\subsection{Torque Limits via Differentiable Inverse Dynamics}
\label{sec:torque_approach}
Trajectories that violate actuator limits cannot be executed. While most planners treat torque constraints as a post-hoc check, we enforce them directly via a differentiable inverse dynamics engine based on the Recursive Newton-Euler Algorithm (RNEA)~\cite{featherstone2000robot}.
RNEA computes the joint torques $\tau$ required to produce a given motion $(q, \dot{q}, \ddot{q})$ in three passes over the kinematic tree: (1)~a forward pass from base to tips that propagates link velocities and accelerations, (2)~a force pass that computes the net wrench on each link from its inertia and motion, and (3)~a backward pass from tips to base that accumulates these forces into joint torques. The $O(n)$ complexity makes RNEA the standard choice for inverse dynamics, but its sequential tree traversals pose a challenge for GPU parallelization.

\begin{table}
  \centering
  \caption{Comparison of GPU-accelerated RNEA implementations.}
  \label{tab:rnea_comparison}
  \small
  \begin{tabular}{@{}lccc@{}}
  \toprule
  \textbf{Aspect} & \textbf{\Newton{}} & \textbf{\GRiD{}} & \textbf{Ours} \\
  \midrule
  Robot binding & Runtime & CodeGen & Runtime \\
  Spatial transform & 6$\times$6 & 6$\times$6 & $R$+$p$ (12 fl.) \\
  Spatial inertia & 6$\times$6 & 6$\times$6 & Compact (12 fl.) \\
  Backward pass & Recompute & Jacobian $O(n^2)$ & cache+VJP $O(n)$ \\ \midrule
  Parallelism & None & Vector comp. & Tree-level \\
  Sync primitive & None & Block & Warp \\ \midrule
  Runtime payload & \cmark & \xmark & \cmark \\
  Mimic joints & \cmark & \xmark & \cmark \\
  High-DoF robots & \cmark & \xmark & \cmark \\
  \PyTorch{} interop & \cmark & \xmark & \cmark \\
  \bottomrule
  \end{tabular}
\end{table}

\begin{figure}
  \begin{minipage}[t]{0.48\textwidth}
  \begin{algorithm}[H]
    \small
    \caption{RNEA Forward}
    \label{alg:rnea_forward}
    \KwIn{$q, \dot{q}, \ddot{q}$, $f_{\text{ext}}$, robot params}
    \KwOut{$\tau$, cache $\{v, a, f\}$}
    \tcp{Pass 1: Forward (base $\to$ tips) propagate velocities and accelerations}
    \For{level $\ell = 0$ \KwTo $L-1$}{
      \ForPar{link $k$ at level $\ell$}{
        \tcp{\textcolor{darkgreen}{recompute transforms to save memory}}
        $R_k, p_k \gets \texttt{localRp}(T_k^{\text{fixed}}, q)$\;
        \eIf{$k$ is root}{
          $v_k \gets S_k \dot{q}_k$\;
          $a_k \gets X_k g + S_k \ddot{q}_k$\;
        }{
          $v_k \gets X_k v_{\text{pa}} + S_k \dot{q}_k$\;
          $a_k \gets X_k a_{\text{pa}} + S_k \ddot{q}_k + v_k \!\times_m\! S_k \dot{q}_k$\;
        }
      }
      \texttt{syncwarp()}\;
    }
    \tcp{Pass 2: Forces (parallel)}
    \ForPar{link $k = 0$ \KwTo $n-1$}{
      $f_k \gets I_k a_k + v_k \!\times^*\! (I_k v_k) - f_{\text{ext},k}$\;
    }
    \tcp{Pass 3: Backward (tips $\to$ base) accumulate forces into joint torques}
    \For{level $\ell = L-1$ \KwTo $0$}{
      \ForPar{link $k$ at level $\ell$}{
        \tcp{\textcolor{darkgreen}{$+=$ handles mimic joints}}
        $\tau_{j_k} \mathrel{+}= m_k S_k^\top f_k$\;
        \If{$k \neq$ root}{$f_{\text{pa}} \mathrel{+}= X_k^\top f_k$}
      }
      \texttt{syncwarp()}\;
    }
  \end{algorithm}
  \end{minipage}%
  \hfill
  \begin{minipage}[t]{0.48\textwidth}
  \begin{algorithm}[H]
    \small
    \caption{RNEA Backward (VJP)}
    \label{alg:rnea_backward}
    \KwIn{$\bar{\tau}$, $q, \dot{q}$, cached $\{v, a, f\}$}
    \KwOut{$\bar{q}, \bar{\dot{q}}, \bar{\ddot{q}}, \bar{f}_{\text{ext}}$}
    \tcp{Pass 1: Adjoint of backward (base $\to$ tips)}
    \For{level $\ell = 0$ \KwTo $L-1$}{
      \ForPar{link $k$ at level $\ell$}{
        $\bar{f}_k \gets S_k (m_k \bar{\tau}_{j_k})$\;
        \If{$k \neq$ root}{
          $\bar{f}_k \mathrel{+}= X_k \bar{f}_{\text{pa}}$\;
          $\bar{q}_{j_k} \mathrel{+}= m_k (X_k \bar{f}_{\text{pa}})^\top \text{crf}(S_k) f_k$\;
        }
      }
      \texttt{syncwarp()}\;
    }
    \tcp{Pass 2: Adjoint of external force}
    \ForPar{link $k = 0$ \KwTo $n-1$}{
      $\bar{f}_{\text{ext},k} \gets -\bar{f}_k$\;
    }
    \tcp{Pass 3: Adjoint of forward (tips $\to$ base)}
    Init $\bar{a}_k, \bar{v}_k \gets 0$ $\forall k$\;
    \For{level $\ell = L-1$ \KwTo $0$}{
      \ForPar{link $k$ at level $\ell$}{
        $\bar{a}_k \mathrel{+}= I_k \bar{f}_k$\;
        $\bar{v}_k \mathrel{-}= \text{crf}(\bar{f}_k) I_k v_k + I_k \text{crm}(v_k) \bar{f}_k$\;
        $\bar{\ddot{q}}_{j_k} \mathrel{+}= m_k S_k^\top \bar{a}_k$\;
        $\bar{\dot{q}}_{j_k} \mathrel{+}= m_k S_k^\top \bar{v}_k - m_k (\text{crf}(v_k) \bar{a}_k)_{s_k}$\;
        \If{$k \neq$ root}{
          $\bar{a}_{\text{pa}} \mathrel{+}= X_k^\top \bar{a}_k$; ~
          $\bar{v}_{\text{pa}} \mathrel{+}= X_k^\top \bar{v}_k$\;
          $\bar{q}_{j_k} \mathrel{-}= m_k \bar{a}_k^\top \text{crm}(S_k) X_k a_{\text{pa}}$\;
          $\bar{q}_{j_k} \mathrel{-}= m_k \bar{v}_k^\top \text{crm}(S_k) X_k v_{\text{pa}}$\;
        }
      }
      \texttt{syncwarp()}\;
    }
  \vspace{0.35cm}
  \end{algorithm}
  \end{minipage}
  \caption[RNEA Forward and Backward Kernels]{RNEA forward and VJP backward kernels. We use Featherstone's spatial algebra~\cite{featherstone2000robot}: $X_k$ is the spatial transform from parent to child frame, $I_k$ the spatial inertia, and $S_k$ the joint motion subspace (a unit 6-vector for 1-DoF joints). The operators $\text{crm}(\cdot)$ and $\text{crf}(\cdot)$ denote the $6\times 6$ motion and force cross-product matrices, with shorthand $v \times_m u = \text{crm}(v) \cdot u$ and $v \times^* f = \text{crf}(v) \cdot f$. Bar notation ($\bar{v}$, $\bar{f}$) indicates adjoint variables, i.e., gradients with respect to that quantity. The scalar $m_k$ is the mimic joint multiplier, and $j_k$ maps link $k$ to its actuated joint index.}
  \label{fig:rnea_algorithms}
\end{figure}

\paragraph{Limitations of Existing GPU Implementations}
Two GPU-accelerated RNEA implementations exist: \Newton{}~\cite{newton} and \GRiD{}~\cite{plancher2022grid}. \Newton{} supports runtime inertial parameter changes (e.g., payloads) but launches six kernels per call due to its modular architecture, recomputes forward quantities during backpropagation, and becomes a bottleneck at our trajectory optimization batch sizes. \GRiD{} generates robot-specific \CUDA{} code with fully unrolled loops and hardcoded indices, achieving high throughput at small batch sizes. However, it requires recompilation for each robot, does not support runtime inertial changes, and its heavy shared memory usage (36 floats per link for 6$\times$6 spatial transforms and inertias) exceeds SM limits on high-DoF robots like humanoids. We summarize these shortcomings in Table~\ref{tab:rnea_comparison}.

\paragraph{Our RNEA Implementation}
We implement RNEA as robot-generic GPU kernels that are parameterized by the number of links $n$, degrees of freedom $d$, and batch size $B$. The kinematic tree topology, spatial inertias $I_k$, and joint types are supplied at runtime, so a single compiled binary supports any robot described by a URDF. Three design choices drive performance: compact spatial representations that reduce shared memory pressure, a VJP-based backward pass that avoids $O(n^2)$ Jacobian materialization, and tree-level parallelism with warp-level synchronization.

\emph{Compact spatial representations.}
Rather than materializing full 6$\times$6 spatial transforms and inertias (72 floats/link), we store transforms in factored Featherstone form, specifically a 3$\times$3 rotation $R$ plus translation $p$ (12 floats), and recompute spatial products on-the-fly via the identity $X \cdot [\omega; v] = [R^\top \omega;\, R^\top(v + \omega \times p)]$. Spatial inertias use a compact 12-float representation (mass, CoM, rotational inertia tensor), with products computed analytically rather than via 6$\times$6 matrix multiplication.

\emph{VJP-based backward pass.}
\GRiD{} computes the full $n \times n$ Jacobians $\partial \tau / \partial q$ and $\partial \tau / \partial \dot{q}$ per Carpentier~\cite{carpentier2018analytical}, incurring $O(n^2)$ cost. For gradient-based optimization, we instead compute the vector-Jacobian product (VJP) directly: given $\partial \loss / \partial \tau$, we obtain $\partial \loss / \partial q$, $\partial \loss / \partial \dot{q}$, $\partial \loss / \partial \ddot{q}$ in $O(n)$ time via an adjoint RNEA pass. When external forces $f_{\text{ext}}$ are provided, the backward pass also computes $\partial \loss / \partial f_{\text{ext}} = -\bar{f}$, enabling gradient flow through contact or interaction forces.

\emph{Memory layout.}
The forward kernel caches velocities and accelerations packed without padding (12 floats) plus forces with padding for alignment (8 floats), totaling 20 floats/link. Transforms are recomputed from $q$ rather than cached, trading ${\sim}48$ arithmetic instructions per link for 12 floats of global memory bandwidth. The backward kernel loads robot inertia parameters (mass, CoM, rotational inertia) into block-shared memory once per thread block, eliminating redundant global loads across batches.

\emph{Tree-level parallelism.}
Multiple threads within a warp cooperate on links at the same tree depth, synchronized via \texttt{\_\_syncwarp()} rather than block-level \texttt{\_\_syncthreads()}, reducing synchronization overhead.
Global memory loads are issued as \texttt{float4} vectors to ensure coalesced access.

Algorithms~\ref{alg:rnea_forward}--\ref{alg:rnea_backward} detail the forward and VJP kernels. Our design achieves 14$\times$ speedup over \Newton{} and is within 1.5--2$\times$ of \GRiD{} at trajectory optimization batch sizes, while supporting runtime payload changes, \PyTorch{} interoperability, and humanoid-scale robots. Importantly, RNEA constitutes only 24\% of trajectory optimization time and 29\% of total motion planning time (Sec.~\ref{sec:computation-time}), so the end-to-end overhead of dynamics-aware planning remains modest.

\section{Experimental Results}
\label{sec:results}

We evaluate \project{} on dynamics-aware motion planning benchmarks and profile the planning pipeline (Secs.~\ref{sec:res-bspline-motion-plan}--\ref{sec:computation-time}), benchmark the key computational components, namely inverse dynamics (Sec.~\ref{sec:res-rnea}), high-DoF inverse kinematics (Sec.~\ref{sec:res-ik}), and ESDF generation (Sec.~\ref{sec:res-esdf}), and demonstrate two applications: humanoid motion retargeting and locomotion policy training (Sec.~\ref{sec:res-retarget}), and real-world manipulation with live depth fusion (Sec.~\ref{sec:res-real}).
Unless stated otherwise, all experiments were run on an NVIDIA RTX 4090 GPU.

\subsection{Dynamics-Aware Planning}
\label{sec:res-bspline-motion-plan}

\begin{figure}
    \centering
    \includegraphics[width=0.99\linewidth]{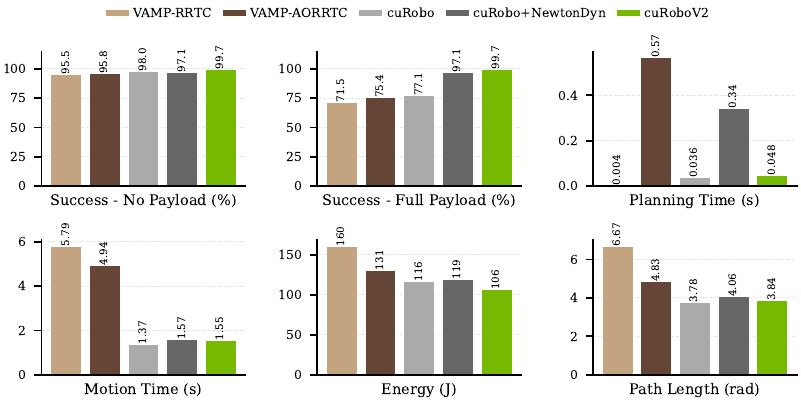}
    \caption[Benchmark Comparison]{Motion Planning Results (75th percentile). Top: Kinematic success (solid) vs.\ dynamics success with 3~kg payload (hatched). Methods without dynamics constraints fail when torque limits are checked, even at zero payload. Middle/Bottom: Our \project{} achieves the highest payload success (99.7\%) and lowest energy (106~J), without sacrificing other trajectory quality metrics.
    }
    \label{fig:mp_metrics}
\end{figure}

We evaluate our B-spline trajectory optimization (Sec.~\ref{sec:bspline}) on the MotionBenchMaker~\cite{chamzas2022motionbenchmaker} and M$\pi$Nets~\cite{fishman2022mpinets} datasets, comprising 2600 problems in cluttered environments for the Franka Panda. We compare against VAMP~\cite{thomason2024vamp}, a state-of-the-art sampling-based planner with post-hoc time-optimal parameterization (RRT-Connect (RRTC) and asymptotically-optimal AORRTC~\cite{aorrtc_2025} variants), and \cuRobo{}~\cite{curobo_icra23}, a GPU-accelerated trajectory optimizer. We also evaluate \cuRobo{} extended with inverse dynamics constraints.

We report three success metrics: kinematic success (collision-free with position, velocity, and acceleration limits), zero-payload dynamics success (additionally checking torque limits without payload), and payload dynamics success (torque limits with 3\,kg end-effector payload). Fig.~\ref{fig:mp_metrics} summarizes the results. While all optimization-based methods achieve $>$99\% kinematic success, the gap emerges when dynamics are checked: even without payload, \cuRobo{} drops from 99.8\% to 97.9\% and sampling-based methods fall from 98.9\% to 95.5--95.8\%, while \project{} maintains 99.7\%. Under full payload, the difference is stark: \project{} maintains 99.7\% success while cuRobo drops to 77.1\% and sampling-based methods fall to 72--75\%. This validates our core claim: trajectories planned without dynamics constraints become infeasible even without payload, and the problem is exacerbated under realistic loads.

The B-spline formulation also yields superior trajectory quality. At 75th percentile (Fig.~\ref{fig:mp_metrics}),
\project{} achieves the lowest energy consumption (106\,J vs.\ 116\,J for \cuRobo{} and 131--160\,J for VAMP),
since the B-spline formulation produces inherently smooth trajectories and the optimizer explicitly minimizes an energy cost.
\project{} requires only 48\,ms planning time (vs.\ 36\,ms for \cuRobo{}), a modest overhead for guaranteed executable trajectories that eliminate costly re-planning at execution time. \project{} solves 66 more payload problems than \cuRobo{}+\NewtonDyn{} (2591 vs.\ 2525 of 2600), showing that B-splines' implicit smoothness yields a better-conditioned optimization landscape than per-timestep parameterizations.

\begin{figure}
    \centering
    \includegraphics[width=0.45\linewidth]{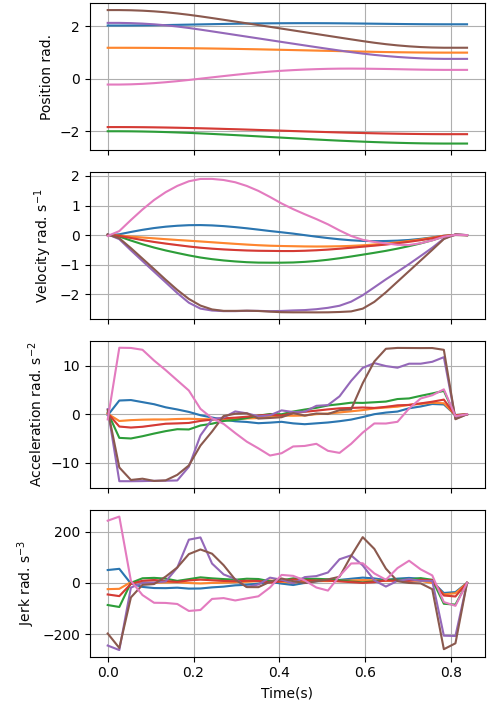}
    \includegraphics[width=0.45\linewidth]{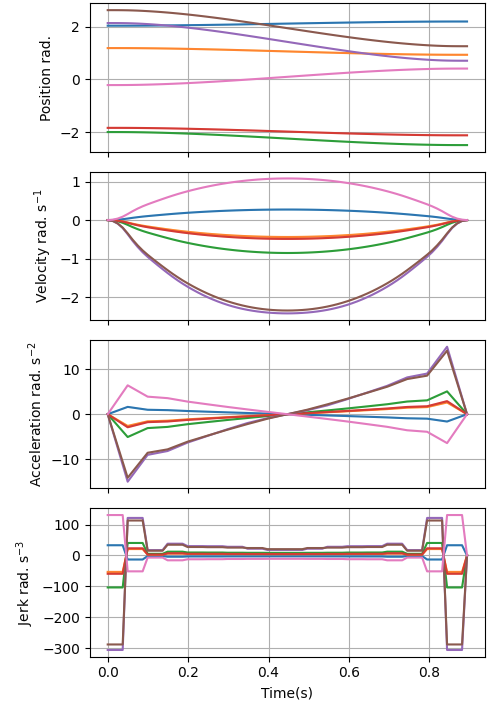}
    \caption[B-spline vs.\ per-timestep trajectory profiles]{Trajectory profiles for the same motion planning problem (each color denotes one joint). Left: \cuRobo{}'s per-timestep optimization produces discontinuous acceleration and jerk. Right: \project{}'s B-spline optimization yields smooth, continuous derivatives suitable for real robot execution.}
    \label{fig:mp_visual}
\end{figure}

Fig.~\ref{fig:mp_visual} compares trajectories from \project{} (B-spline optimization) and \cuRobo{} (per-timestep joint position optimization). Both solvers start from the same seed and run for an equal number of iterations. Optimizing directly in joint position space produces non-smooth acceleration and jerk profiles, whereas B-spline optimization yields clean, continuous derivatives by construction.

\begin{tcolorbox}[
    colback=lightgrey!50!white,
    colframe=blockborder,
    boxrule=0.5pt,
    arc=4pt,
    left=6pt, right=6pt, top=6pt, bottom=6pt,
    drop shadow=black!15!white,
    enhanced, breakable
] Trajectories planned without dynamics constraints become infeasible under payload. \project{}'s B-spline formulation maintains $>$99\% success while naturally producing smooth, continuous derivatives suitable for real robot execution at full payload (3\,kg).
\end{tcolorbox}

\subsection{Runtime Breakdown}
\label{sec:computation-time}

We profile \project{}'s motion planning pipeline on an NVIDIA RTX 4090 GPU with \texttt{torch.compile} enabled, comparing execution with and without dynamics constraints. Table~\ref{tab:phase-timing} shows the kernel execution time breakdown across the three planning phases, along with end-to-end wall-clock times that include Python and \CUDA{} graph launch overhead.
Enabling dynamics adds 9.38\,ms (+44\%) to kernel execution time, distributed across IK solving (+3.70\,ms) and trajectory optimization (+5.69\,ms). The seed IK phase remains unaffected as it only computes an initial configuration without dynamics evaluation. The RNEA kernels add 5.12\,ms (23.8\%) to TrajOpt when dynamics is enabled. Including all overhead, the end-to-end planning wall-clock time is 35\,ms without dynamics and 42\,ms with dynamics. Notably, the end-to-end increase is only +7\,ms despite +9.38\,ms in kernel time: \CUDA{} streams run independent kernels in parallel (e.g., cost, kinematics, and dynamics kernels), so the added dynamics kernels partially overlap with existing computation rather than executing sequentially. \CUDA{} graphs further batch these launches into a single invocation with fixed overhead, and because dynamics adds more GPU work, a larger fraction of this fixed launch overhead is hidden behind kernel execution.

\begin{table}
\centering
\caption{Planning time per phase with and without dynamics constraints. \%Kernel shows fraction of total kernel execution time; \%Total shows fraction of end-to-end wall-clock time. The gap between Kernel Total and End-to-end is \CUDA{} graph launch and Python overhead.}
\label{tab:phase-timing}
\small
\setlength{\tabcolsep}{4pt}
\begin{tabular}{l rrr rrr r}
\toprule
& \multicolumn{3}{c}{\textbf{No Dynamics}} & \multicolumn{3}{c}{\textbf{With Dynamics}} & \\
\cmidrule(lr){2-4} \cmidrule(lr){5-7}
\textbf{Phase} & Time & \%Kernel & \%Total & Time & \%Kernel & \%Total & $\boldsymbol{\Delta}$ \\
\midrule
Seed IK (LM) & 0.74\,ms & 3.5 & 2.1 & 0.74\,ms & 2.4 & 1.8 & --- \\
IK Solver (L-BFGS) & 4.87\,ms & 22.7 & 13.9 & 8.57\,ms & 27.8 & 20.4 & +3.70\,ms \\
TrajOpt (MPPI+L-BFGS) & 15.82\,ms & 73.8 & 45.2 & 21.51\,ms & 69.8 & 51.2 & +5.69\,ms \\
\midrule
\textbf{Kernel Total} & \textbf{21.44\,ms} & 100 & \textbf{61.3} & \textbf{30.83\,ms} & 100 & \textbf{73.4} & \textbf{+9.38\,ms} \\
\midrule
\textbf{End-to-end (wall-clock)} & \textbf{35\,ms} & & \textbf{100} & \textbf{42\,ms} & & \textbf{100} & \textbf{+7\,ms} \\
\bottomrule
\end{tabular}
\end{table}

\begin{table}
    \centering
    \caption{Full planning time breakdown with dynamics enabled. Per-Iter is within TrajOpt (132.8\,$\mu$s/iter, 162 iterations). \%Kernel is relative to the kernel total (30.83\,ms); \%Total is relative to end-to-end wall-clock time (42\,ms).}
    \label{tab:kernel-breakdown}
    \small
    \setlength{\tabcolsep}{4pt}
    \begin{tabular}{l rrr rr}
    \toprule
    \textbf{Kernel} & \textbf{TrajOpt/Iter} & \textbf{TrajOpt} & \textbf{IK+TrajOpt} & \textbf{\%Kernel} & \textbf{\%Total} \\
    \midrule
    RNEA (fwd + bwd) & 31.6\,$\mu$s & 5.12\,ms & 8.82\,ms & 28.6 & 21.0 \\
    Swept-sphere collision & 21.0\,$\mu$s & 3.41\,ms & 3.41\,ms & 11.1 & 8.1 \\
    L-BFGS + line search & 15.2\,$\mu$s & 2.46\,ms & 3.31\,ms & 10.7 & 7.9 \\
    Kinematics (fwd + bwd) & 14.4\,$\mu$s & 2.33\,ms & 4.03\,ms & 13.1 & 9.6 \\
    Cspace + pose + speed costs & 9.5\,$\mu$s & 1.54\,ms & 2.07\,ms & 6.7 & 4.9 \\
    Self-collision & 5.1\,$\mu$s & 0.83\,ms & 0.83\,ms & 2.7 & 2.0 \\
    B-spline & 3.6\,$\mu$s & 0.58\,ms & 0.58\,ms & 1.9 & 1.4 \\
    \midrule
    Elementwise + reduce & 23.3\,$\mu$s & 3.77\,ms & 5.07\,ms & 16.4 & 12.1 \\
    Other & 9.1\,$\mu$s & 1.47\,ms & 1.97\,ms & 6.4 & 4.7 \\
    \midrule
    IK + TrajOpt & 132.8\,$\mu$s & 21.51\,ms & 30.09\,ms & 97.6 & 71.6 \\
    Seed IK (LM) & --- & --- & 0.74\,ms & 2.4 & 1.8 \\
    \midrule
    \textbf{Kernel Total} & & \textbf{21.51\,ms} & \textbf{30.83\,ms} & \textbf{100} & \textbf{73.4} \\
    \midrule
    \textbf{End-to-end (wall-clock)} & & & \textbf{42\,ms} & & \textbf{100} \\
    \bottomrule
    \end{tabular}
    \end{table}

\paragraph{Kernel Time Breakdown}
Table~\ref{tab:kernel-breakdown} shows the dominant GPU kernels with dynamics enabled, broken down by TrajOpt (162 iterations) and the full IK + TrajOpt pipeline.
RNEA dominates the compute budget at 29\% of the full pipeline when dynamics is enabled, growing from 5.12\,ms in TrajOpt alone to 8.82\,ms when IK is included. Despite this overhead, the total planning time of 30.83\,ms enables real-time motion generation at $>$30\,Hz, making dynamics-aware planning practical for reactive manipulation. At the per-iteration level, RNEA adds 31.6\,$\mu$s, increasing the per-iteration cost from 97.7\,$\mu$s to 132.8\,$\mu$s (+36\%).

\begin{tcolorbox}[
    colback=lightgrey!50!white,
    colframe=blockborder,
    boxrule=0.5pt,
    arc=4pt,
    left=6pt, right=6pt, top=6pt, bottom=6pt,
    drop shadow=black!15!white,
    enhanced, breakable
] Enabling dynamics constraints adds only +7\,ms to end-to-end planning time (35\,ms $\to$ 42\,ms), a modest overhead that prevents the 22--27\% payload success drop seen in methods that plan without dynamics constraints.
\end{tcolorbox}

\subsection{Inverse Dynamics Comparison}
\label{sec:res-rnea}

\begin{figure}
    \centering
    \includegraphics[width=0.98\linewidth]{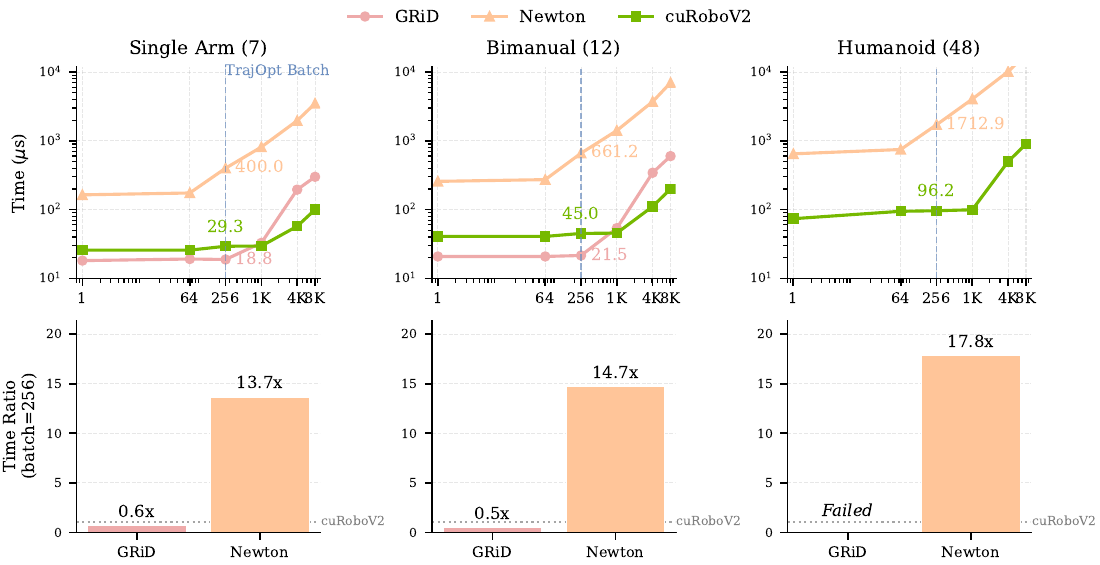}
    \caption[Inverse dynamics performance across batch sizes]{Inverse dynamics performance (forward + backward pass) across batch sizes. \project{} is 1.5--2$\times$ slower than \GRiD{}'s code-generated kernels on simple robots but 14--18$\times$ faster than \Newton{}. Critically, \GRiD{} fails on the 48-DoF humanoid due to shared memory limits, while \project{} scales gracefully.}
    \label{fig:rnea_comparison}
\end{figure}

We benchmark our inverse dynamics implementation against \GRiD{}~\cite{plancher2022grid}, which generates robot-specific \CUDA{} kernels via code synthesis, and \Newton{}~\cite{newton}, a general-purpose GPU simulation library. We evaluate on three robot configurations: a 7-DoF single arm, a 12-DoF bimanual system, and a 48-DoF humanoid, measuring forward and backward pass times across batch sizes from 1 to 8192 (Fig.~\ref{fig:rnea_comparison}).

On the 7-DoF arm at batch size 256, \GRiD{} achieves 18.8\,$\mu$s while \project{} requires 29.3\,$\mu$s, a 1.5$\times$ overhead compared to \GRiD{}'s code-generated kernels. However, \project{} outperforms \Newton{} (400.0\,$\mu$s) by 14$\times$. The gap widens on the 12-DoF bimanual robot, where \project{} (45.0\,$\mu$s) is 2$\times$ slower than \GRiD{} (21.5\,$\mu$s) but 15$\times$ faster than \Newton{} (661.2\,$\mu$s).
The critical advantage of \project{} emerges on the 48-DoF humanoid: \GRiD{} fails entirely due to shared memory limits in its code-generated kernels, while \project{} scales gracefully at 96.2\,$\mu$s (batch size 256). \Newton{} completes in 1712.9\,$\mu$s, making \project{} 18$\times$ faster. This scalability to high-DoF systems is essential for planning on complex robots.

\begin{tcolorbox}[
    colback=lightgrey!50!white,
    colframe=blockborder,
    boxrule=0.5pt,
    arc=4pt,
    left=6pt, right=6pt, top=6pt, bottom=6pt,
    drop shadow=black!15!white,
    enhanced, breakable
] \project{} scales gracefully from single-arm manipulators to full humanoids, running 14--18$\times$ faster than \Newton{}, a general-purpose physics engine. Code-generated kernels (\GRiD{}) are 1.5--2$\times$ faster on simple arms but hit shared memory limits and fail entirely on the 48-DoF humanoid, making \project{} the only viable option for high-DoF systems.
\end{tcolorbox}

\subsection{High-DoF Inverse Kinematics}
\label{sec:res-ik}

We evaluate the scalability of our kinematics and self-collision (Sec.~\ref{sec:high_dof_engine}) across three robots of increasing complexity: a 7-DoF Franka Panda, a 12-DoF dual-UR10e system, and a 48-DoF Unitree G1 humanoid. We compare against \cuRobo{}~\cite{curobo_icra23}, \Newton{}~\cite{newton}, and \PyRoki{}~\cite{kim2025pyroki}.

\begin{figure}
    \centering
    \begin{tabular}{c c}
        \includegraphics[width=0.49\linewidth]{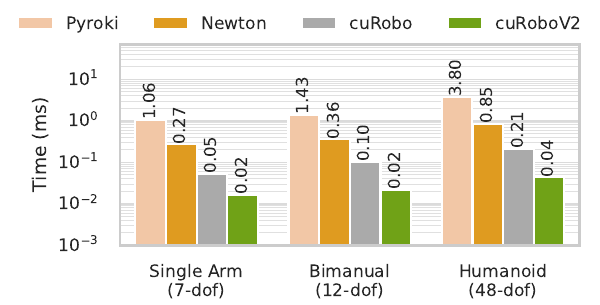} &
        \includegraphics[width=0.49\linewidth]{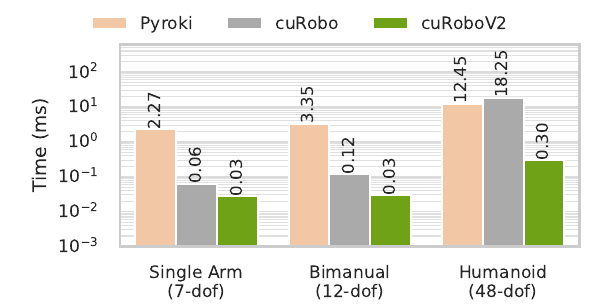}\\
        (a) Kinematics Fwd + Bwd &
        (b) Kinematics + Self-Collision Fwd + Bwd
    \end{tabular}
    \caption[Kinematics Comparison]{Kinematics and self-collision performance (batch size 1000). (a) Forward and backward pass without self-collision. (b) With self-collision, \project{} achieves 61$\times$ speedup over \cuRobo{} on the humanoid via map-reduce. These fast building blocks enable scalable IK and retargeting.}
    \label{fig:res_kin_pose}
\end{figure}

\paragraph{Fast Kinematics and Self-Collision}
Fig.~\ref{fig:res_kin_pose} benchmarks the forward and backward pass at batch size 1000. Without self-collision (a), our data-driven architecture delivers consistent speedups: 3--5$\times$ over \cuRobo{}, 16--19$\times$ over \Newton{}, and 62--86$\times$ over \PyRoki{}. The advantage grows with robot complexity, as our parallel branching kinematics avoids serial tree traversal. Adding self-collision (b) amplifies the gap: the G1 humanoid has 162k collision pairs (vs.\ 818 for Franka), and our map-reduce approach achieves 61$\times$ speedup over \cuRobo{} and 42$\times$ over \PyRoki{}. This throughput enables the applications below.

\begin{figure}
    \centering
    \begin{tabular}{cc}
    \includegraphics[width=0.48\linewidth, trim={0 0 0 0}, clip]{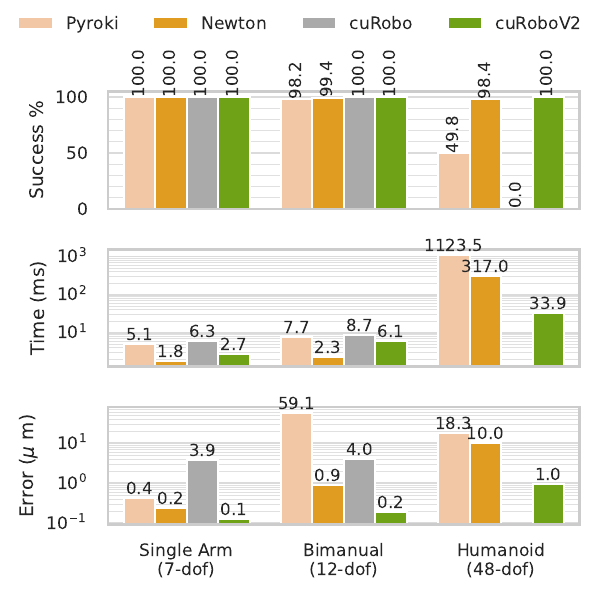}
    &
    \includegraphics[width=0.48\linewidth, trim={0 0 0 0}, clip]{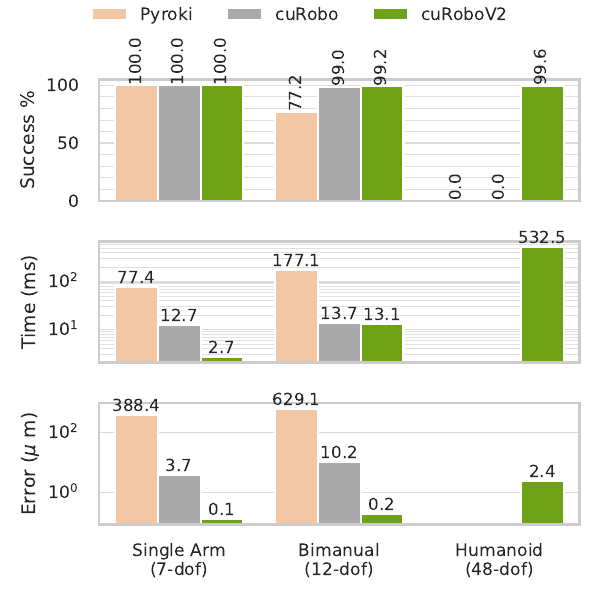}\\
    (a) IK & (b) Self-Collision-Free IK
    \end{tabular}
    \caption[IK Metrics]{IK results (batch size 100). (a)~Standard IK: all methods achieve near-perfect success on single-arm robots; on the 48-DoF humanoid, \project{} reaches 100\% success while \PyRoki{} drops to 49.8\%. (b)~Self-collision-free IK: \project{} achieves 99.6\% success on the humanoid via LM seeding followed by L-BFGS refinement; \cuRobo{} and \PyRoki{} both fail completely (0\%).}
    \label{fig:cfree_ik_metrics}
\end{figure}

\paragraph{Global Inverse Kinematics}

For standard IK without collision avoidance, all methods achieve near-perfect success on single-arm robots (Fig.~\ref{fig:cfree_ik_metrics}), though \project{} reaches lower position error (0.13\,$\mu$m vs.\ 0.43\,$\mu$m for \PyRoki{} and 3.9\,$\mu$m for \cuRobo{}) thanks to our efficient Jacobian computation enabling Levenberg-Marquardt optimization. The scalability advantage appears on the 48-DoF humanoid: \project{} achieves 100\% success with 0.96\,$\mu$m position error in 34\,ms, while \Newton{} reaches 98.4\% with 10\,$\mu$m error in 317\,ms (9$\times$ slower) and \PyRoki{} only 49.8\% with 18\,$\mu$m error in 1123\,ms (33$\times$ slower). Note: results from \PyRoki{}'s paper do not verify joint limit satisfaction.

\paragraph{Collision-Free IK}
When self-collision avoidance is enabled (Fig.~\ref{fig:cfree_ik_metrics}), the gap widens further: \project{} achieves 99.6\% success on the humanoid with 2.4\,$\mu$m position error in 533\,ms, while both \cuRobo{} and \PyRoki{} fail completely (0\%). This success stems from our two-stage approach: Levenberg-Marquardt first solves the pose objective (nonlinear least squares), providing a good initialization at the target; L-BFGS then refines under collision constraints, leveraging its strength on non-convex objectives. Our scalable kinematics and self-collision implementation make this seeding strategy computationally tractable for high-DoF systems.

\begin{tcolorbox}[
    colback=lightgrey!50!white,
    colframe=blockborder,
    boxrule=0.5pt,
    arc=4pt,
    left=6pt, right=6pt, top=6pt, bottom=6pt,
    drop shadow=black!15!white,
    enhanced, breakable
] For highly articulated systems like humanoids, a two-stage optimization strategy (LM seeding followed by L-BFGS refinement) is crucial for finding collision-free IK solutions. Our parallel branching kinematics and map-reduce collision checking make this strategy computationally tractable, achieving 99.6\% success where prior methods fail completely.
\end{tcolorbox}

\subsection{Humanoid Retargeting}
\label{sec:res-retarget}

\subsubsection{Motion Retargeting Quality}
We use the GMR library~\cite{joao2025gmr} to map over 70k frames of human motion from the LeFan dataset to the Unitree G1 humanoid. GMR natively uses \mink{} for per-frame local IK with only joint-limit constraints; we refer to this default configuration as GMR. We additionally evaluate \mink{} as a standalone solver with self-collision and ground-plane collision checking enabled, and compare against \PyRoki{} and \project{}-IK by swapping out GMR's default solver.

Motion retargeting is inherently ill-posed: the target robot's kinematic structure differs from the human source, and not all human poses lie within the robot's reachable workspace. Solvers must therefore balance tracking fidelity against feasibility, gracefully degrading on unreachable targets rather than diverging. This trade-off is typically mediated through weighted cost terms whose tuning significantly affects per-method performance. To isolate algorithmic differences, we evaluate each solver as a drop-in replacement within the GMR pipeline, which was developed and tuned for \mink{} without collision avoidance. We do not re-tune cost weights or constraint formulations for any method, including \project{}-IK. Dedicated per-solver tuning could improve absolute metrics for all methods; our goal is to compare algorithmic properties (collision-aware optimization, global seeding, constraint handling) under a controlled, common pipeline.

\begin{figure}
    \centering
    \includegraphics[width=0.99\linewidth]{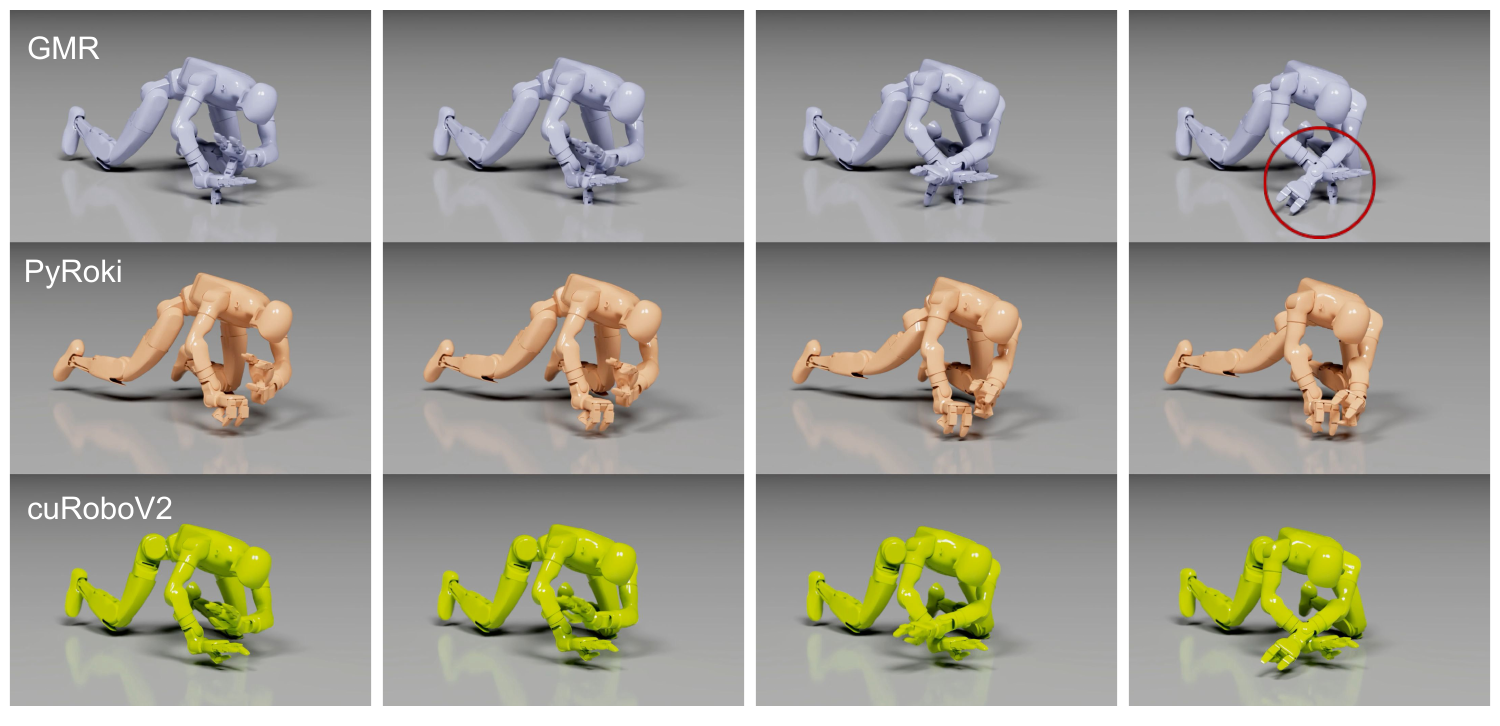}\\
    \caption[Retargeting visual comparison during crawling]{Retargeting visual comparison across GMR (top), \PyRoki{} (middle), and \project{} (bottom) during a crawling sequence. The sequence requires hands to cross: GMR produces self-collisions (see red circle) since collision avoidance is not activated, \PyRoki{} avoids collision but fails near contact boundaries, preventing the hands from crossing. \project{} resolves self-collisions gracefully, achieving accurate non-colliding retargeting.}
    \label{fig:visual_humanoid_retarget}
\end{figure}

\begin{figure}
    \centering
    \includegraphics[width=0.99\linewidth,
    trim= 0 0 0 0, clip]{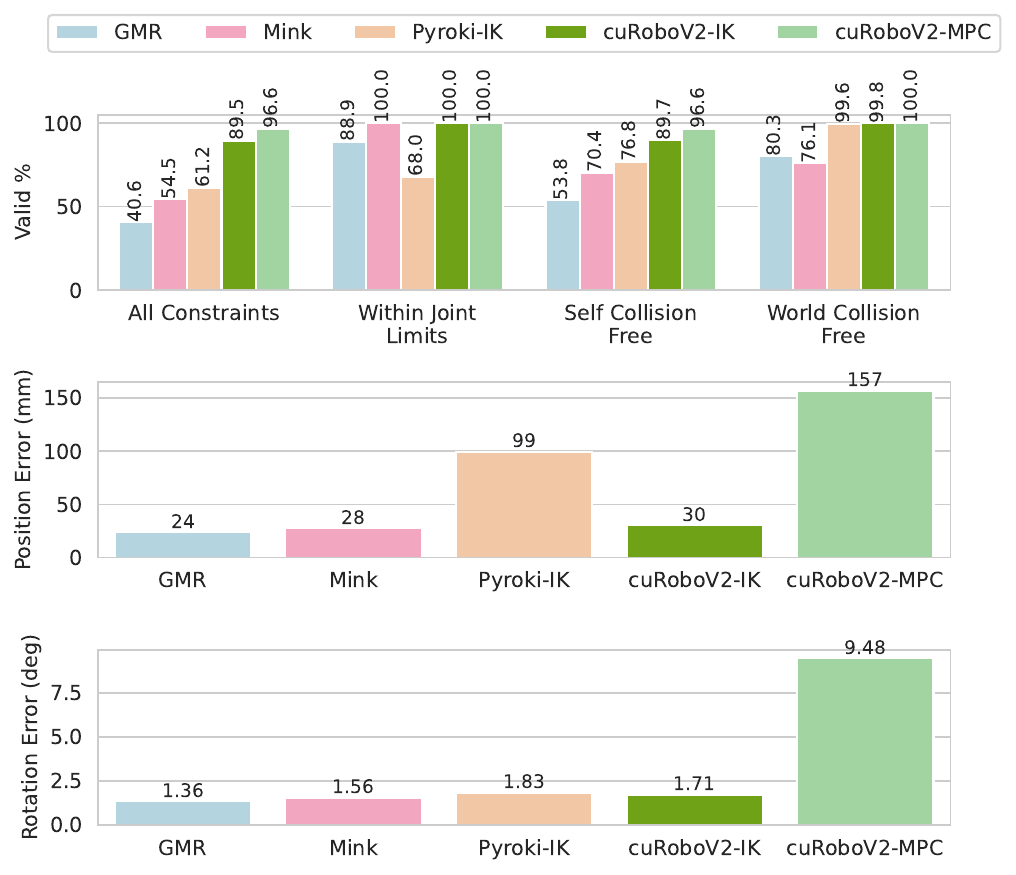}
    \caption[Retargeting Results]{Humanoid motion retargeting on the G1 (70k frames). \project{}-IK achieves 89.5\% constraint satisfaction vs.\ \PyRoki{} 61.2\%, \mink{} 54.5\%, and GMR 40.6\%, thanks to LM+L-BFGS seeding for collision-free solutions. MPC reaches 96.6\% by checking collisions between frames but at the cost of tracking accuracy.}
    \label{fig:retargeting_comparison}
\end{figure}

\begin{figure}
    \centering
    \includegraphics[width=0.99\linewidth]{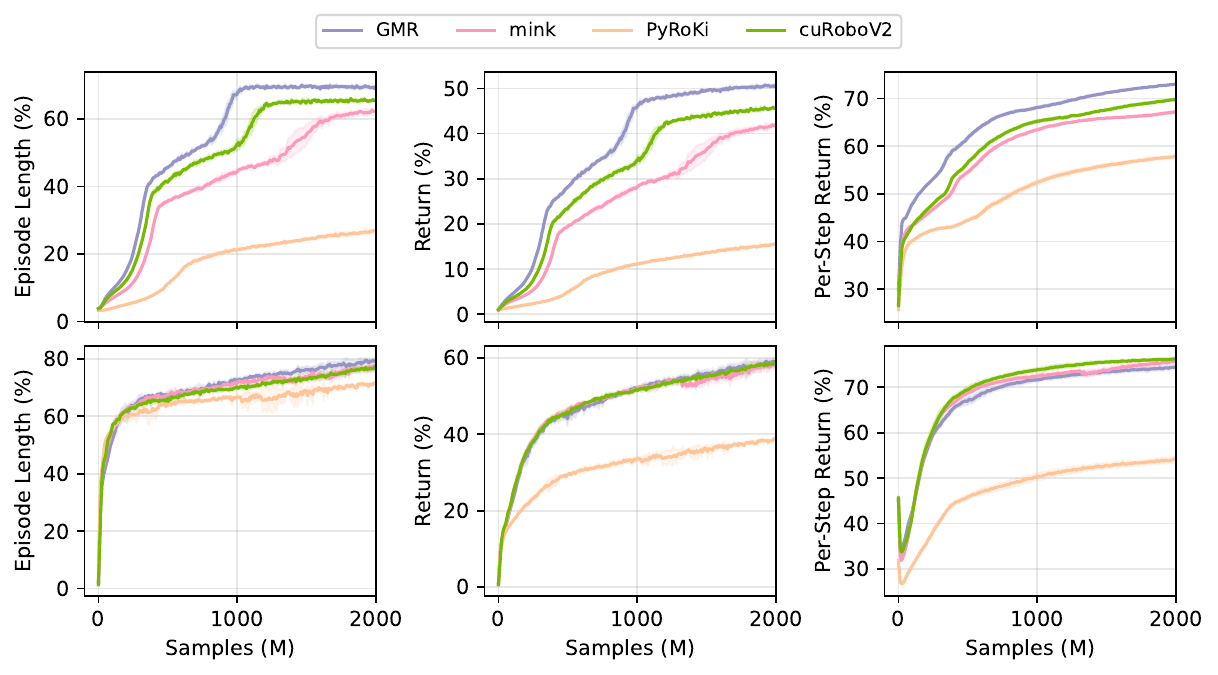}\\
    (a) Policy Training\\
    \includegraphics[width=0.99\linewidth]{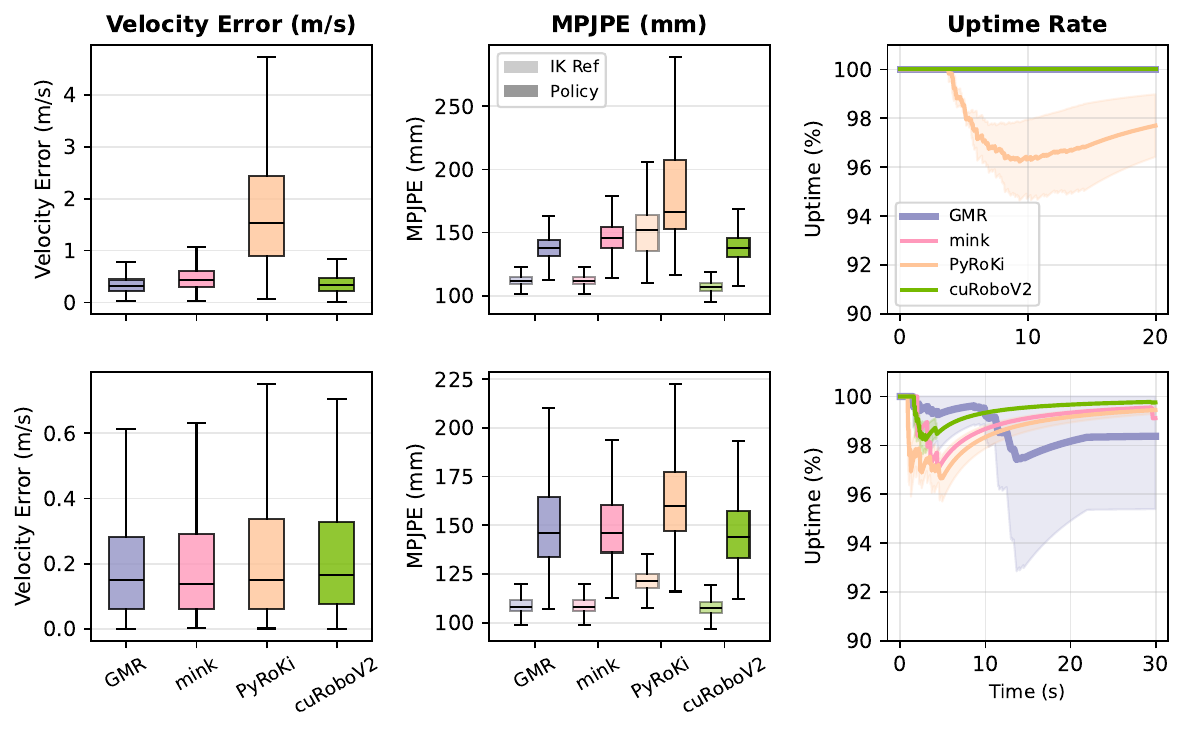}\\
    (b) Policy Evaluation
    \caption[Locomotion policy training and evaluation]{(a)~Policy training curves (normalized to \%). On running (top), \PyRoki{} plateaus at 31\% episode length due to repeated falls. On crawling (bottom), \project{}, GMR, and \mink{} all appear similar, masking the differences revealed by evaluation. (b)~Evaluation box plots across 5 seeds. \project{} achieves the lowest MPJPE with tight variance, while GMR exhibits 12$\times$ higher MPJPE variance on crawling.}
    \label{fig:retargeting_metrics}
\end{figure}

Fig.~\ref{fig:visual_humanoid_retarget} illustrates a crawling sequence where hands must cross: GMR produces self-collisions since collision avoidance is not activated, \PyRoki{} avoids them but fails near contact boundaries, and \project{} resolves the crossing gracefully. Trajectories are interpolated from 30\,fps to 120\,fps and validated for joint limits, self-collision, and ground contact (Fig.~\ref{fig:retargeting_comparison}). \project{}-MPC achieves 96.6\% constraint satisfaction, followed by \project{}-IK at 89.5\%, \PyRoki{} at 61.2\%, \mink{} at 54.5\%, and GMR at 40.6\%. Enabling collision checking in \mink{} improves over GMR's 40.6\% to 54.5\%, but \project{}-IK still leads among per-frame solvers (89.5\%) thanks to its LM+L-BFGS seeding strategy for finding collision-free solutions. The further gain from MPC stems from checking collisions \emph{between} frames via its trajectory representation, whereas IK methods only enforce constraints at each discrete frame; interpolation-induced collisions cause IK failures.

However, \project{}-MPC's pose tracking accuracy is worse than \project{}-IK's, as MPC solves a harder optimization problem that enforces smoothness and collision constraints across waypoints without the benefit of the LM optimizer. We therefore recommend \project{}-IK for retargeting, which already substantially outperforms existing methods by providing joint configurations that are not only accurate but also self-collision-free and scene-collision-free.
As shown in Sec.~\ref{sec:res-mimickit}, \project{}-IK already produces significantly better downstream policies than all baselines, making it our recommended retargeting method. Exploring MPC's inter-frame collision checking for policy learning is left to future work.

\subsubsection{Locomotion Policy Training}
\label{sec:res-mimickit}

\begin{figure}
    \centering
    \includegraphics[width=0.99\linewidth]{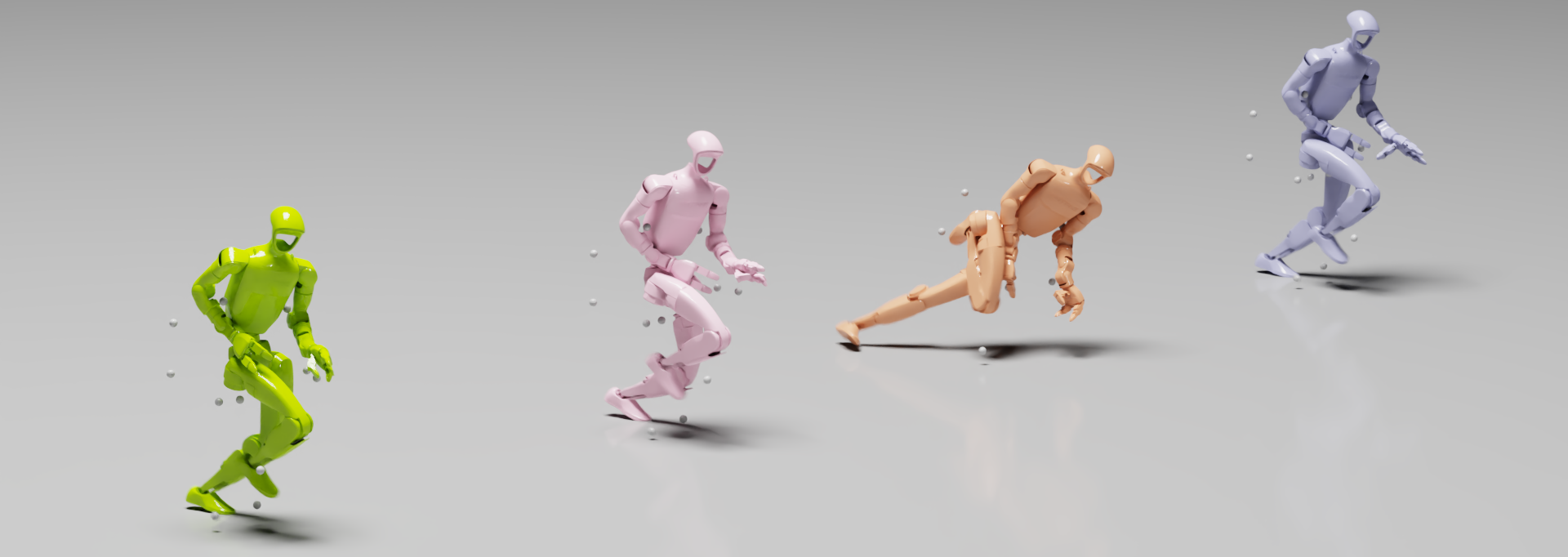}\\
    (a) Running\\
    \includegraphics[width=0.99\linewidth]{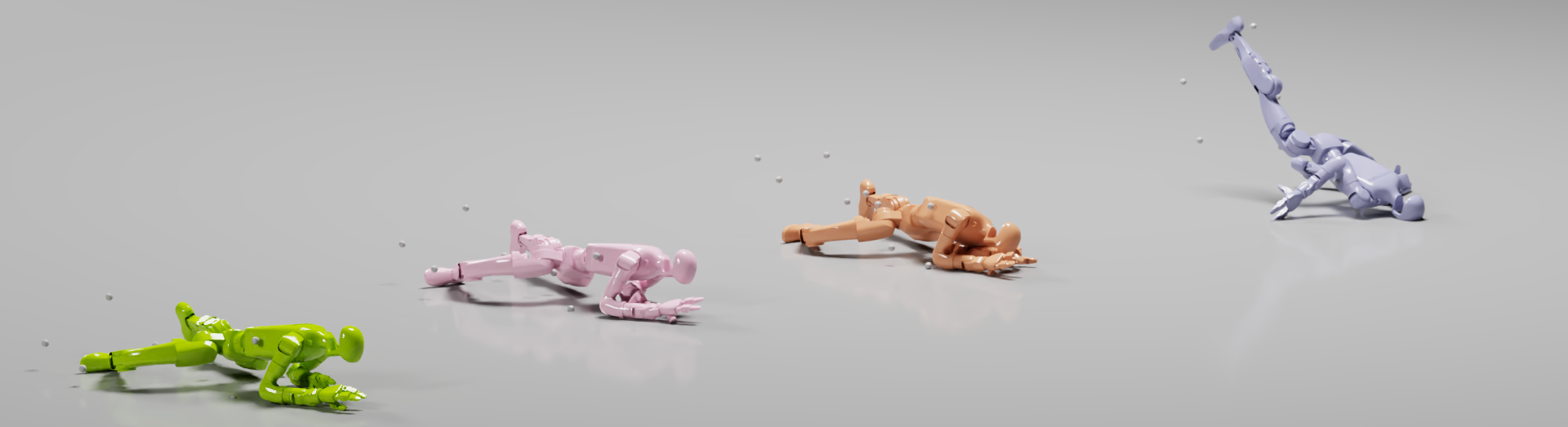}\\
    (b) Crawling \\
    \includegraphics[width=0.99\linewidth]{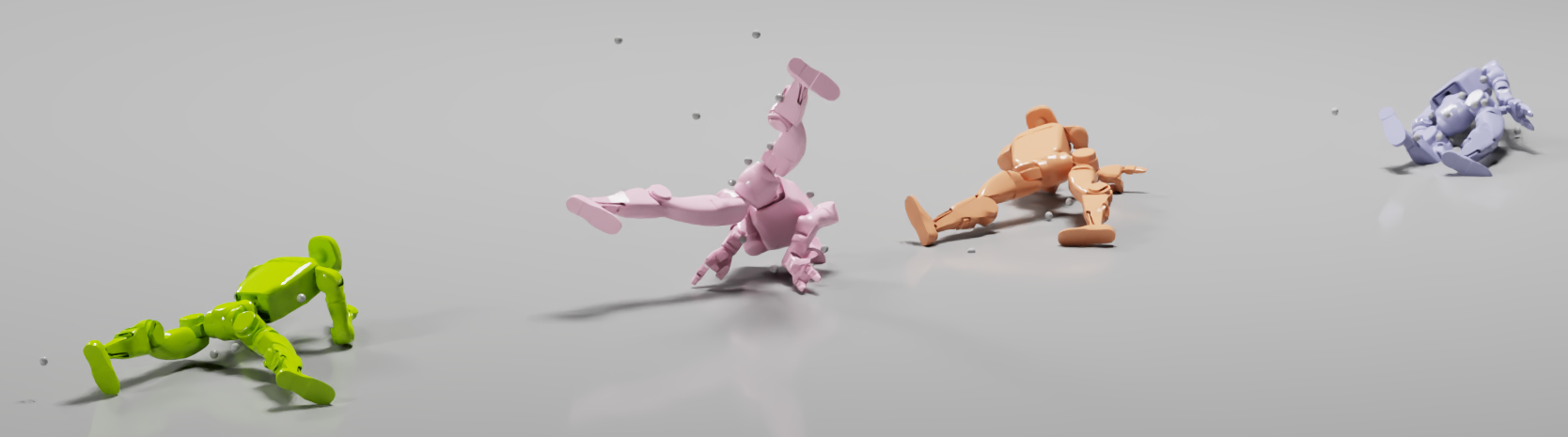}\\
    (c) Crawling (post-pivot sequence)
    \caption[Learned locomotion policies on the Unitree G1]{Learned locomotion policies on the Unitree G1 (green: \project{}, pink: \mink{}, orange: \PyRoki{}, purple: GMR). (a)~Running: \PyRoki{} falls at high velocity; the other three methods maintain stable gaits. (b,\,c)~Crawling across two sequences: GMR exhibits high seed-to-seed variability, tumbling in (b) and collapsing in (c), consistent with its 13$\times$ higher MPJPE variance. \PyRoki{} and \mink{} perform better with collision constraints. In (c), \mink{} fails after pivoting while \PyRoki{} ends with incorrect foot poses; both failure modes could be resolved with careful tuning. \project{} succeeds reliably across all conditions.}
    \label{fig:retargeted_policy}
\end{figure}

To validate that retargeting quality translates to downstream policy performance, we train locomotion policies for the Unitree G1 using MimicKit~\cite{MimicKitPeng2025}, which implements DeepMimic~\cite{2018-TOG-deepMimic} on Newton's Mujoco-Warp simulator. We retarget two motion segments (20\,s running, 30\,s crawling) using \project{}-IK, GMR, \mink{} (with self-collision and ground-plane collision checking enabled), and \PyRoki{}, then train each configuration for 2~billion samples across 5~seeds. We evaluate root velocity error, mean per-joint position error (MPJPE), alive rate, and episode resets (Fig.~\ref{fig:retargeting_metrics}).
The training curves and evaluation metrics reveal two qualitatively different failure modes depending on the motion and the retargeting method.

On \emph{running} (Fig.~\ref{fig:retargeted_policy}a), \project{}, GMR, and \mink{} all achieve $>$99.7\% alive rates with zero resets, since running involves minimal self-collision risk and all three solvers produce valid reference motions. \project{} and GMR achieve comparable tracking (MPJPE 139.6 vs.\ 138.9\,mm); \mink{} is slightly higher (149.2\,mm), likely because its collision constraints are too conservative for this non-contact motion and could benefit from careful retuning. GMR slightly outperforms \project{} on training metrics (PSR 74\% vs.\ 71\%, return 51\% vs.\ 46\%, episode length 70\% vs.\ 65\%), with \mink{} trailing slightly (PSR 67\%, episode length 62\%), confirming that when self-collision is not a factor, a standard local IK solver without collision handling is sufficient. In contrast, \PyRoki{}'s constraint-violating reference motions (61.2\% constraint satisfaction, Sec.~\ref{sec:res-retarget}) cause the policy to fall repeatedly, resulting in an average episode length of only 31\% compared to 65\% for \project{} and 70\% for GMR. These short episodes starve the policy of learning signal: the PSR plateaus at 59\% while the other methods reach 67--74\%. The consequence is a cycle in which the policy never learns to sustain the high velocities required for running, accumulating 13.8~resets and 33\% higher MPJPE (185.8\,mm).

On \emph{crawling} (Fig.~\ref{fig:retargeted_policy}b), where limbs must cross and self-collision handling is critical, the training curves tell a subtler and more informative story. The return and episode length for \project{}, GMR, and \mink{} appear nearly identical throughout training (return 60\% vs.\ 60\% vs.\ 58\%; episode length 79\% vs.\ 80\% vs.\ 77\% at 2B samples), and the PSR gap is modest (77\% vs.\ 75\% vs.\ 76\%). A na\"ive reading of the training curves would suggest comparable performance. However, the evaluation metrics (Fig.~\ref{fig:retargeting_metrics}b) reveal a stark difference: \project{} achieves an MPJPE of 151.4$\pm$4.5\,mm, while GMR reaches 178.0$\pm$58.3\,mm, exhibiting 12$\times$ higher variance across seeds. Because GMR does not activate collision avoidance, its reference motions contain poses where limbs interpenetrate. The reinforcement learning loss drives the policy to imitate these infeasible configurations, pushing it toward self-colliding and ground-penetrating states that trigger resets. Whether a given seed converges to a reasonable policy depends on whether the optimizer finds a path that avoids the worst constraint violations, explaining the large seed-to-seed variability. Enabling self-collision and ground-plane collision checking in \mink{} substantially reduces this variance: MPJPE drops to 156.7$\pm$9.5\,mm with 3.4 resets, confirming that collision-aware retargeting improves policy consistency. \PyRoki{} accumulates twice the resets of \project{} (4.6 vs.\ 2.2).

Across both motions, \project{} yields the best overall policy: lowest MPJPE (145.5\,mm), highest alive rate (99.6\%), and fewest resets (1.1), followed by \mink{} (153.0\,mm, 1.7 resets), GMR (158.5\,mm, 1.8 resets), and \PyRoki{} (177.6\,mm, 9.2 resets). This analysis highlights an important methodological point: standard training metrics (return, episode length, and PSR) are necessary but insufficient for evaluating retargeting quality. All three can appear healthy while the policy tracks physically infeasible reference poses. Downstream pose accuracy and cross-seed consistency are the metrics that reveal the true impact of reference motion quality. The same RL framework (MimicKit) and the same policy architecture are used across all experiments; the only variable is the IK solver used for retargeting. The results therefore isolate the effect of retargeting quality on policy performance: collision-free, joint-limit-satisfying reference motions from \project{}-IK produce policies that are not only more accurate but also more reliably trainable.

\begin{tcolorbox}[
    colback=lightgrey!50!white,
    colframe=blockborder,
    boxrule=0.5pt,
    arc=4pt,
    left=6pt, right=6pt, top=6pt, bottom=6pt,
    drop shadow=black!15!white,
    enhanced, breakable
] \project{}-IK's collision-free retargeting (89.5\% constraint satisfaction vs.\ 61\% \PyRoki{}, 55\% \mink{}, 41\% GMR) directly improves policy training: 25\% lower MPJPE than \PyRoki{} on running, 8$\times$ fewer resets overall, and 12$\times$ lower cross-seed MPJPE variance than GMR on crawling.
\end{tcolorbox}

\subsection{Fast and Safe ESDF at mm-Resolution}
\label{sec:res-esdf}

\begin{figure}
    \centering
    \includegraphics[width=0.99\linewidth]{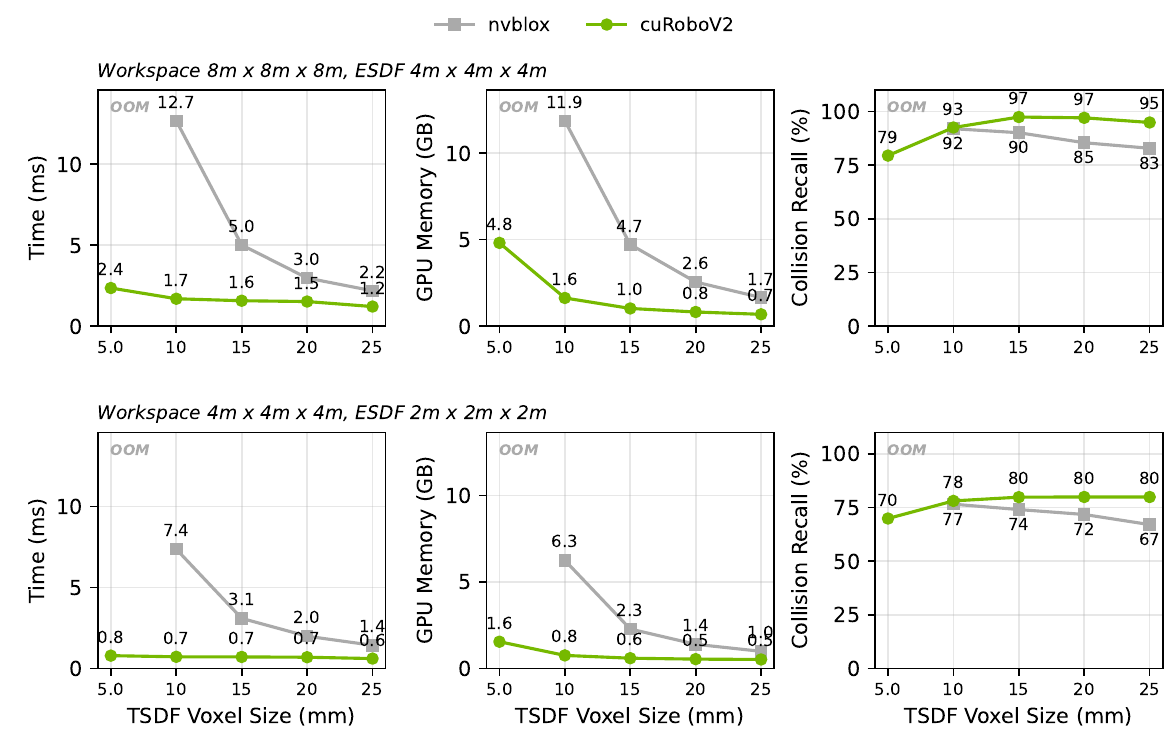}
    \caption[ESDF benchmark across TSDF resolutions]{%
    ESDF benchmark across TSDF resolutions (10--25\,mm) at 50\% and 100\% workspace coverage. \project{} achieves 2--10$\times$ lower total time and 2--8$\times$ less memory than \nvblox{} across all configurations, while maintaining higher collision recall. TSDF integration time is comparable across methods; the speedup stems from PBA+ distance propagation. Recall is computed only over the region where \nvblox{} produces valid distances.}
    \label{fig:esdf-result}
\end{figure}

We evaluate our ESDF generation pipeline against \nvblox{}~\cite{nvblox}, a GPU-accelerated TSDF/ESDF library.
Both methods process the same depth images from the Redwood bedroom scene~\cite{park2017}.
\project{} uses a fixed 20\,mm ESDF voxel size across all TSDF resolutions; when the TSDF voxel size exceeds 20\,mm, the ESDF resolution matches the TSDF. \nvblox{} requires matching TSDF and ESDF resolutions.
We measure depth image to ESDF generation time, GPU memory, and collision recall (Fig.~\ref{fig:esdf-result}). To ensure a fair comparison, recall is computed only over the region where \nvblox{} produces valid distance values, so both methods are evaluated on the same set of query points.

Fig.~\ref{fig:esdf-result} compares \project{} against \nvblox{} across TSDF resolutions (10--25\,mm) at two workspace coverage levels. At 10\,mm TSDF resolution with full workspace coverage, \project{} generates the complete ESDF in 1.69\,ms vs.\ 12.68\,ms for \nvblox{}, a 7$\times$ speedup, while using 7$\times$ less memory (1.63\,GB vs.\ 11.87\,GB) and matching recall (92.5\% vs.\ 92.0\%). At 20\,mm, the advantage persists: \project{} achieves 1.52\,ms vs.\ 2.97\,ms with 97.0\% recall (vs.\ 85.4\% for \nvblox{}) in 3$\times$ less memory (0.82\,GB vs.\ 2.55\,GB).

Since the benchmark plots report only total time, we detail the TSDF/ESDF breakdown here. TSDF integration time is comparable across methods, 0.50--0.55\,ms for \project{} vs.\ 0.30--0.76\,ms for \nvblox{}, confirming that integration is not the bottleneck. The speedup comes entirely from ESDF generation: at 10\,mm (100\% coverage), our PBA+ propagation completes in 1.15\,ms vs.\ 11.93\,ms for \nvblox{}'s iterative block-based ESDF updates~\cite{nvblox}, a 10$\times$ improvement. ESDF accounts for 94\% of \nvblox{}'s total time but only 68\% of ours. Although our dense PBA+ recomputes the full distance field each frame, this is faster than incremental approaches because the computation is fully separable, requires a fixed number of kernel launches, and is \CUDA{}-graph compatible. The recall gap is partly attributable to TSDF weight thresholding: \nvblox{}'s distance-based weighting couples noise rejection and surface fidelity into a single parameter that is difficult to tune, whereas our pixel-count-based voting decouples the two, making it easier to set a threshold that retains valid surfaces without amplifying noise.

\begin{figure}
    \centering
    \includegraphics[width=0.99\linewidth]{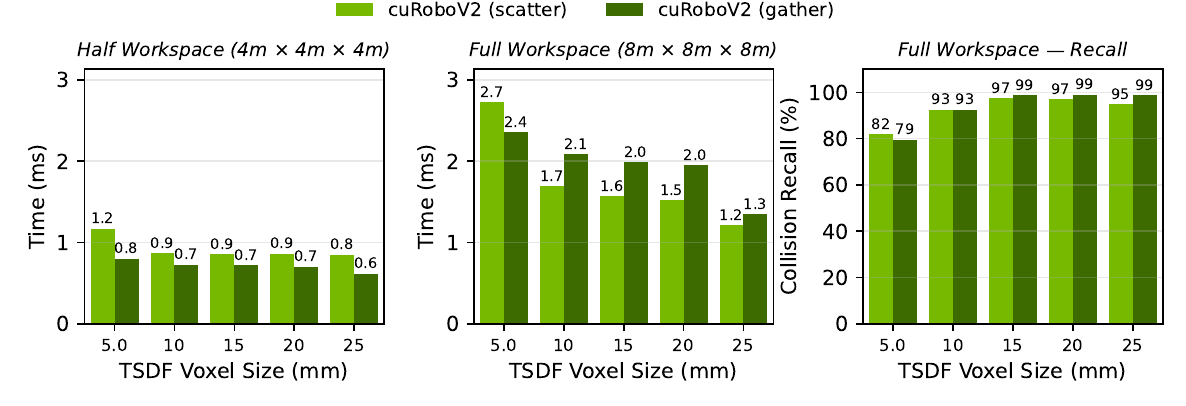}
    \caption[Scatter vs.\ gather seeding strategy comparison]{%
    Scatter vs.\ gather seeding strategies. Gather achieves lower total time (e.g., 0.73\,ms vs.\ 0.87\,ms at 10\,mm, 50\% coverage) and enables CUDA graph capture due to its fixed launch dimension.}
    \label{fig:esdf-scatter-gather}
\end{figure}

Fig.~\ref{fig:esdf-scatter-gather} compares our two seeding strategies. With scatter seeding, the ESDF stage takes 0.45\,ms at 10\,mm (50\% coverage) vs.\ 0.30\,ms with gather, a 1.5$\times$ reduction from eliminating atomic contention. At full workspace coverage, gather's ESDF time increases to 1.54\,ms (vs.\ 1.15\,ms for scatter) because the gather kernel iterates over all ESDF voxels regardless of occupancy, while scatter only visits allocated TSDF blocks. Despite this, gather's total time remains competitive (2.08\,ms vs.\ 1.69\,ms) and it enables CUDA graph capture of the entire ESDF pipeline, which is critical for achieving deterministic latency in real-time control loops.

The recall difference between the two strategies is governed by the TSDF-to-ESDF resolution ratio $r = v_{\text{esdf}} / v_{\text{tsdf}}$. Scatter seeds exactly those ESDF cells containing a TSDF surface voxel center, producing a surface band exactly one ESDF voxel thick.
Gather's 7-point stencil (center plus six face centers at $\pm\tfrac{1}{2}v_{\text{esdf}}$) probes positions at cell boundaries, reading TSDF voxels that belong to neighboring ESDF cells.
This effectively dilates the seed band to ${\sim}1.5$ voxels, giving PBA+ more surface sites near boundaries and reducing distance errors at surface-adjacent voxels.
When the TSDF is much finer than the ESDF ($r \geq 4$), scatter checks ${\sim}r^3$ TSDF voxels per ESDF cell, far exceeding gather's 7 probes, and achieves slightly higher recall (e.g., +2.5\% at $r{=}4$).
When the TSDF and ESDF resolutions are comparable ($r \approx 1$), scatter checks only 1--2 TSDF voxels per cell, and gather's denser sampling dominates (e.g., +2--4\% recall at $r{=}1$).
The crossover occurs near $r \approx 2$, where both strategies sample at similar density.

\begin{figure}
    \centering
    \includegraphics[width=0.99\linewidth]{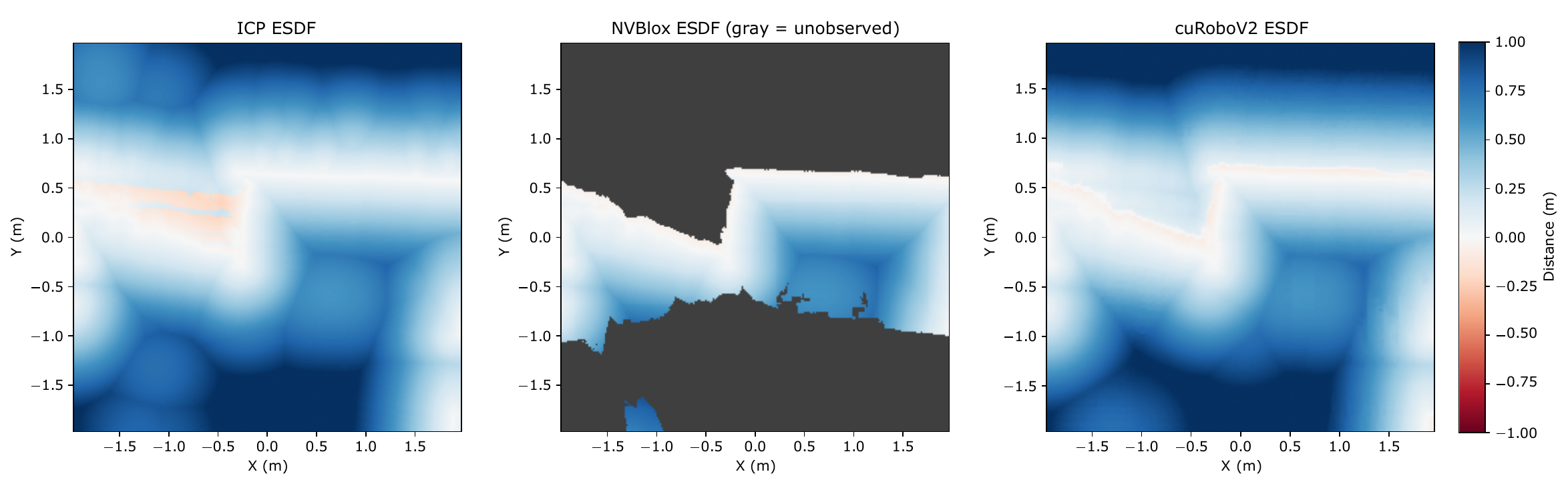}
    \caption[Signed distance field cross-section comparison]{%
    Signed distance field cross-section (4\,m $\times$ 4\,m slice) at 100\% workspace coverage. \nvblox{} (middle) produces distances only within allocated blocks, leaving gaps in unobserved regions. \project{} (right) covers the full workspace, closely matching the ground-truth SDF (left).}
    \label{fig:esdf-comparison}
\end{figure}

Unlike \nvblox{}, which requires matching TSDF and ESDF resolutions, our method supports heterogeneous resolutions: we maintain a fine TSDF (e.g., 5\,mm) for reconstruction fidelity while generating the ESDF at a task-appropriate resolution (e.g., 10--20\,mm). Since the robot is approximated by collision spheres (1--5\,cm diameter), 10\,mm ESDF resolution is sufficient, and finer grids yield diminishing returns. \nvblox{} cannot adopt this strategy, as a 5\,mm TSDF forces a 5\,mm ESDF, inflating both memory and compute.

Fig.~\ref{fig:esdf-comparison} shows a qualitative cross-section of the distance field from the Redwood dataset. \nvblox{} produces signed distances only within allocated blocks, leaving gaps in unobserved regions. \project{} generates a dense ESDF covering the full workspace via PBA+ propagation, providing $O(1)$ distance queries everywhere. Minor surface discrepancies (missed or spurious surfaces) remain, as the TSDF weight threshold requires tuning to balance noise rejection and surface retention.

When known geometry (e.g., tables, shelves) is stamped into the TSDF, collision queries become $O(1)$ trilinear lookups in the ESDF rather than per-cuboid analytical SDF evaluations. On the motion planning benchmark, this reduces mean planning time by 19\% (47\,ms to 38\,ms) with negligible impact on success rate ($\Delta = -0.12\%$) compared to querying individual obstacles, represented as geometries.

\begin{tcolorbox}[
    colback=lightgrey!50!white,
    colframe=blockborder,
    boxrule=0.5pt,
    arc=4pt,
    left=6pt, right=6pt, top=6pt, bottom=6pt,
    drop shadow=black!15!white,
    enhanced, breakable
] By decoupling TSDF and ESDF resolutions, \project{} builds manipulation-scale distance fields up to 10$\times$ faster with 8$\times$ less memory than \nvblox{}. These fast $O(1)$ distance queries directly translate to a 19\% reduction in motion planning time.
\end{tcolorbox}

\subsection{Real-World Manipulation}
\label{sec:res-real}

\begin{figure}
    \centering
    \includegraphics[width=0.95\linewidth]{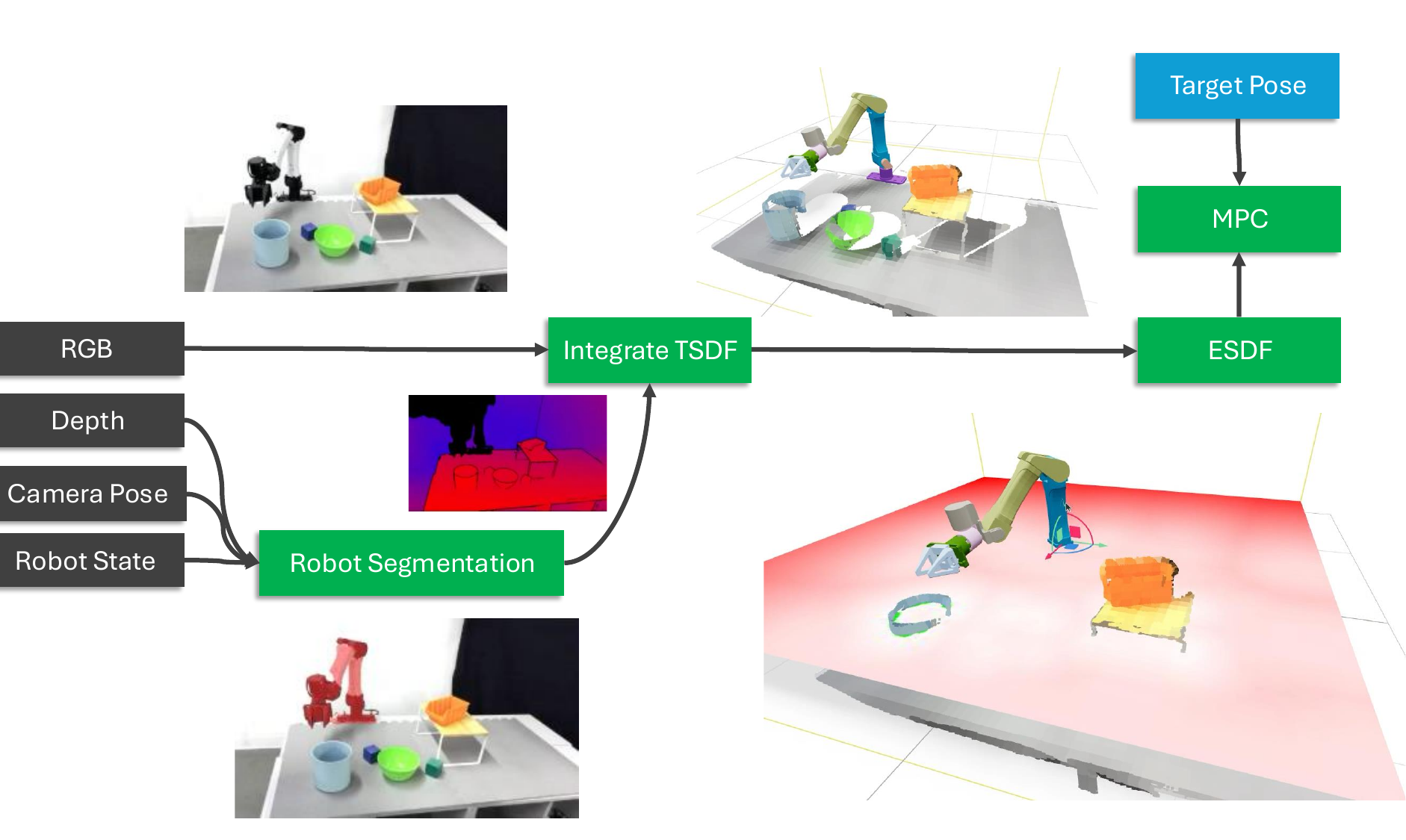}
    \caption[Real-world deployment pipeline with live depth fusion]{Real-world deployment pipeline. Depth from a ZED Mini stereo camera is segmented to remove the robot, fused into a TSDF, and converted to an ESDF for real-time collision avoidance in the MPC trajectory optimizer.
    }
    \label{fig:real-world}
\end{figure}

\begin{figure}
    \centering
    \includegraphics[width=0.98\linewidth]{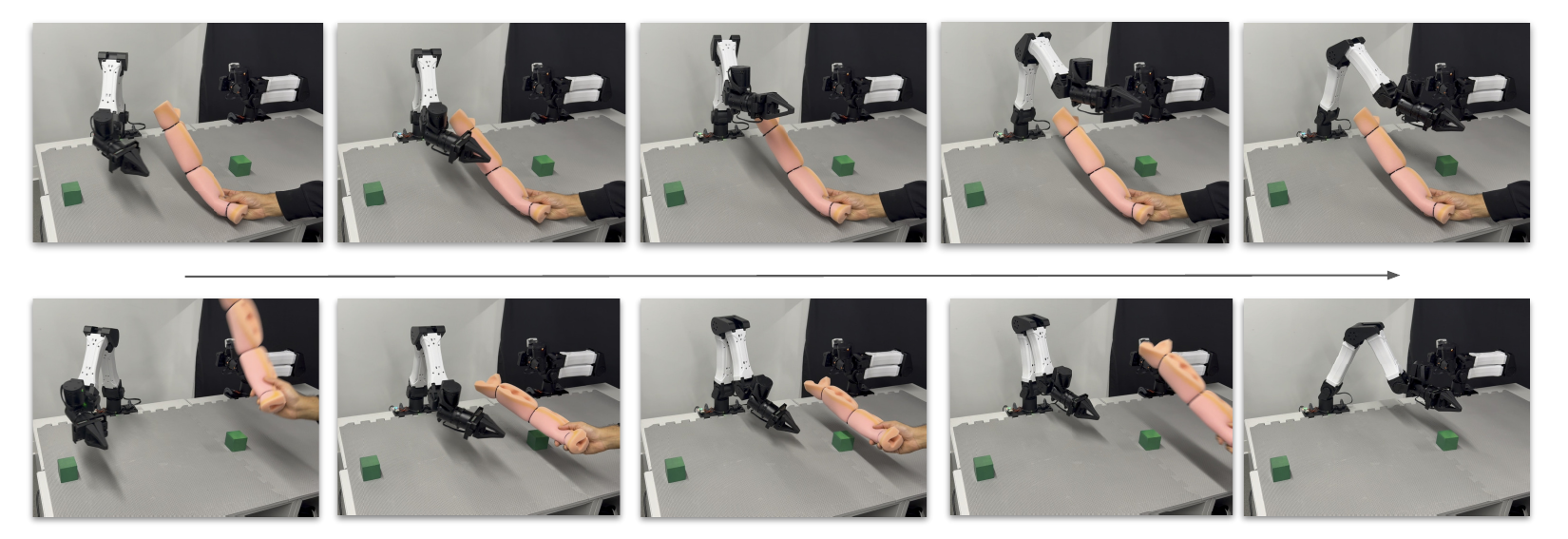}
    \caption[Real-world collision avoidance on the YAM robot]{Real-world collision avoidance on the I2RT YAM robot. The robot plans a motion from left to right while \project{}'s MPC dynamically avoids an obstacle using live ESDF updates from a ZED Mini stereo camera.}
    \label{fig:real_robot}
\end{figure}

We deploy \project{} on an I2RT YAM robot with a calibrated ZED Mini stereo camera (Figs.~\ref{fig:real-world},~\ref{fig:real_robot}). Depth images are generated using ZED's Neural Plus model at 1080p, 15\,Hz; higher frame rates incur prohibitive latency from the neural depth estimator. The robot is removed from each depth image by projecting the kinematic sphere model into image space, and the segmented depth feeds a 2.5\,mm TSDF from which we build a 2\,cm ESDF. This ESDF drives real-time collision avoidance in the MPC trajectory optimizer, which tracks pose commands while avoiding dynamically-observed obstacles. The full pipeline, comprising depth integration, ESDF generation, and trajectory optimization, runs within a single control loop, demonstrating the benefit of our GPU-native architecture.

\section{LLM-assisted Development}
\label{sec:llm_dev}

We hypothesize that a well-structured codebase is a prerequisite for productive LLM-assisted
development, and that the investment pays for itself.
By restructuring \project{} for discoverability, we enabled an LLM coding assistant
(Cursor~IDE with Claude models) to author
many parts of the final codebase, from high-level API wrappers to hand-optimized \CUDA{} kernels.

\subsection{Codebase Discoverability}
\label{sec:discoverability}

Early attempts to use LLM-based coding assistants with the v1 codebase revealed systematic
failure modes.
Each failure pointed to a specific property that, once addressed, enabled reliable
LLM-generated code:

\begin{enumerate}[nosep, leftmargin=*]
    \item \textbf{Config hidden in YAML $\rightarrow$ Config visible in code.}
          The LLM could not discover v1's API surface because configuration lived in YAML
          files loaded via dict mutation at runtime.
          Replacing these with typed dataclasses with documented defaults
          (\code{IKSolverCfg.create(num\_seeds=32)}) made every parameter inspectable in source code.

    \item \textbf{Opaque naming $\rightarrow$ Names predict location.}
          Deep paths like \code{curobo.wrap.reacher.ik\_solver.IKSolver} and opaque class names (\code{CudaRobotModel}) caused the LLM to hallucinate imports.
          Short, stable imports (\code{from curobo import IKSolver}), domain nouns
          (\code{Robot}, \code{Kinematics}), consistent conventions
          (\code{*Cfg}/\code{*Result}/\code{.create()}), and \code{\_\_all\_\_} on every public module eliminated this.

    \item \textbf{Large files $\rightarrow$ Files fit in one context window.}
          v1 averaged 460 lines/file (max 4{,}427) with sparse docstrings, forcing the LLM to work with incomplete context.
          Reducing files to 254 lines on average (max 1{,}643) with docstrings and inline usage examples ensures the LLM can read an entire module at once.

    \item \textbf{Opaque tests $\rightarrow$ Tests as executable documentation.}
          v1's 264 tests in a flat directory with no docstrings gave the LLM no usage examples to learn from.
          3{,}978 tests in a source-mirrored layout with 813 documented test classes provide concrete input/output examples for every public API.

    \item \textbf{Ambiguous signatures $\rightarrow$ Self-documenting interfaces.}
          v1 signatures like \code{start\_state: JointState} provided no shape or batch semantics, causing the LLM to generate incorrectly shaped tensors.
          Documenting shape, dtype, and indexing on every public method lets the LLM generate
          correct code from the docstring alone:
{\small
\begin{verbatim}
  start_state:     Shape (-1, dof). Indexed as start_state[start_state_idx[batch_idx]].
  start_state_idx: Shape (batch_size,). dtype: torch.int32.
\end{verbatim}
}
\end{enumerate}

\noindent These changes benefit human researchers equally, but they are \emph{necessary} for
LLMs, which cannot compensate with intuition or IDE tooling.

\paragraph{Cost of the investment}
\label{sec:codebase_glance}

\begin{figure}
    \centering
    \includegraphics{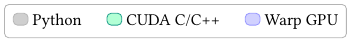}

    \vspace{0.2cm}

    \begin{minipage}[b]{0.65\columnwidth}
    \centering
    \includegraphics[width=\linewidth]{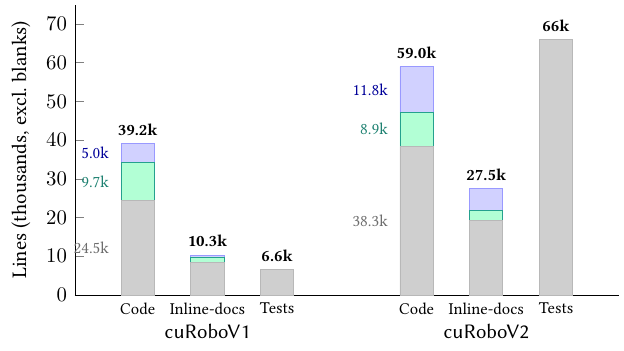}
    \end{minipage}%
    \hfill
    \begin{minipage}[b]{0.34\columnwidth}
    \centering
    \includegraphics[width=\linewidth]{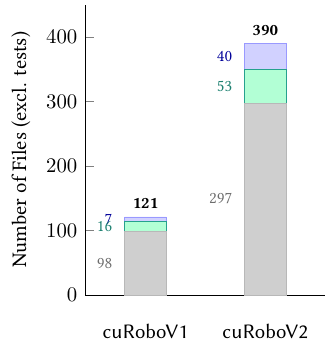}
    \end{minipage}
    \caption[Codebase size comparison across versions]{Codebase size comparison (excluding blank lines).
    \figloc{Left:} Non-blank lines split into Code, Inline-docs, and Tests,
    stacked by language.
    \figloc{Right:} File counts by language.}
    \label{fig:code_size}
\end{figure}
Fig.~\ref{fig:code_size} quantifies the resulting codebase growth.
Framework code grows $1.5{\times}$ while inline documentation grows $2.7{\times}$, reflecting
the investment in discoverability.
\CUDA{} C++ code stays nearly flat as scene collision migrated to \Warp{}, whose share of GPU code
increases from 34\% to 57\%.
Files grow from 121 to 390, reflecting decomposition into single-responsibility modules, and
the test suite grows from 264 to 3{,}978 tests ($15{\times}$, 6.6k to 66k lines).

\subsection{Quantifying LLM Contributions}

\begin{figure}
    \centering
    \includegraphics[width=0.99\linewidth]{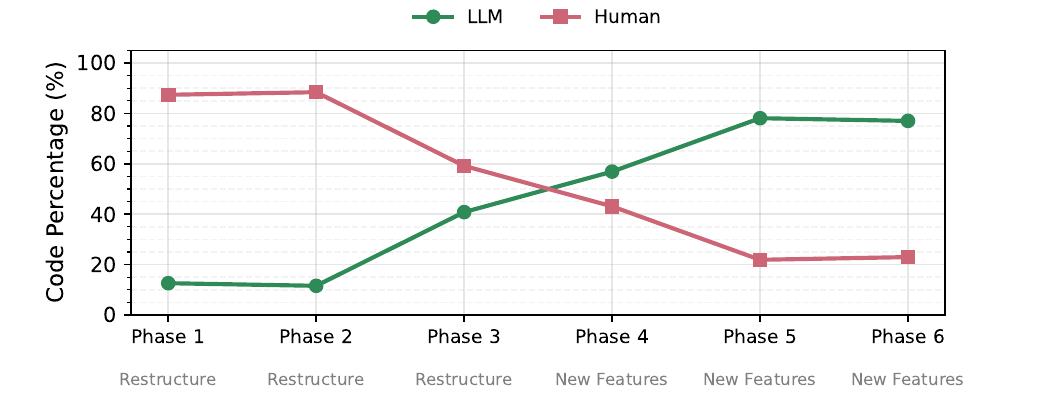}
    \caption[LLM vs.\ human code contributions per development phase]{Percentage of LLM-authored versus human-authored code additions per development
    phase.
    Early phases (R1--R3) focused on refactoring, with the human writing most code.
    Later phases (N1--N3) developed new modules using LLM-assisted workflows, with LLM
    contributions rising to 73\%.}
    \label{fig:llm_usage}
\end{figure}

Fig.~\ref{fig:llm_usage} shows LLM-authored code as a percentage of total additions over six
development phases.
During the initial refactoring (R1--R3), we used Claude Sonnet as the primary model;
the codebase was not yet structured enough for the LLM to contribute substantially, so
researchers wrote the majority of code while restructuring for discoverability, and LLM
contributions accounted for 6--18\% of additions.
Starting from N1, we switched to Claude Opus, and LLM contributions rose
sharply, reaching 50\% in N1 and 73\% in N2, driven by two factors: the discoverability
refactoring made the codebase legible enough for the LLM to take on increasingly complex
tasks, and the more capable model could better leverage the improved structure.
Overall, LLM and researcher contributions split nearly 50/50 across the entire development
period.

\subsection{Case Studies: RNEA, Scene Collision Migration, and PBA ESDF}

We detail three modules developed with substantial LLM assistance, illustrating the
human--LLM collaboration patterns that emerged.

\paragraph{Inverse Dynamics (RNEA) \CUDA{} Implementation \& Optimization}
\begin{figure}
    \centering
    \includegraphics{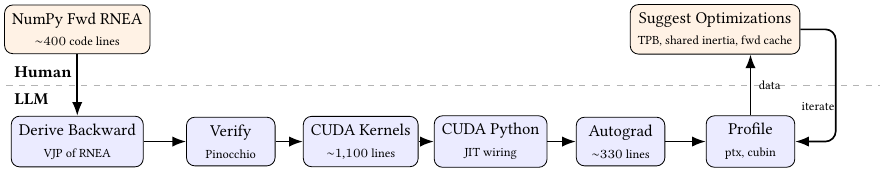}
    \caption[Human--LLM workflow for RNEA development]{Human--LLM workflow for RNEA development.
    The human (top) provides the \NumPy{} forward reference and suggests kernel
    optimizations; the LLM (bottom) handles each subsequent stage from derivation
    through profiling.}
    \label{fig:rnea_workflow}
\end{figure}

Adding differentiable inverse dynamics required implementing Featherstone's Recursive
Newton--Euler Algorithm as a \CUDA{} kernel with an analytical backward pass, a task
involving rigid-body physics, reverse-mode differentiation, and low-level GPU
optimization.
Rather than writing the kernel from scratch, we used a staged workflow in which a
human-written \NumPy{} reference served as the specification and the LLM carried out each
subsequent translation and optimization step (Fig.~\ref{fig:rnea_workflow}).

The human wrote a \NumPy{} forward pass (${\sim}400$ code lines) and tested it on a simple
two-DoF robot.
From this forward reference, the LLM derived the analytical backward pass, i.e., the
reverse-mode adjoint (vector-Jacobian product) of the two-pass RNEA algorithm, produced
a \NumPy{} implementation, and wrote tests verifying both passes against
Pinocchio~\cite{carpentier2019pinocchio}.
Next, the LLM read the docstrings on cuRobo's kinematics dataclasses to understand how
robot parameters are stored, then translated both passes into \CUDA{} kernels
(${\sim}800$ lines across forward and backward kernels, plus ${\sim}300$ lines of
spatial algebra helpers) and wired the module into the \CUDA{} Python backend with a
\PyTorch{} autograd wrapper (${\sim}330$ lines).

Kernel optimization followed an iterative cycle
(Fig.~\ref{fig:rnea_workflow}): the LLM profiled the compiled cubin and reported
metrics (register usage, memory transactions, and compute operations per phase), then
we suggested optimizations (tree-level parallelism for branched robots, block-shared
inertia parameters, a forward cache to avoid recomputation in the backward pass), and
the LLM implemented each one and re-profiled to confirm improvements.

Two episodes highlight the limits of this loop.
The LLM tried to optimize \texttt{sincos} calls in the forward kinematics;
inspecting the PTX, at our direction, revealed the compiler was already lowering them
to polynomial approximations.
Later, the LLM fixated on reducing register pressure based on high virtual register
counts in the PTX, when the cubin profile showed physical register usage was already low.
In both cases, the LLM misread intermediate compiler representations: it could not
distinguish virtual artifacts from real hardware behavior, a judgment call that required
human intervention.

\begin{figure}
    \centering
    \includegraphics{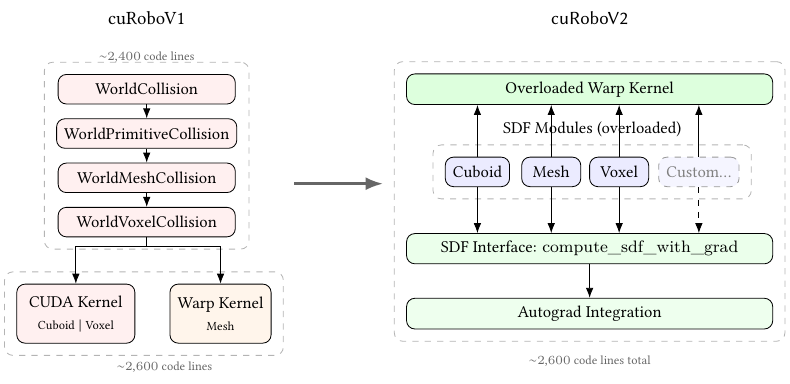}
    \caption[Unified scene collision architecture]{Unified scene collision architecture.
    \textbf{Left:} \cuRoboVone{} splits collision across two kernel backends and a deep
    inheritance chain.
    \textbf{Right:} \project{} unifies all obstacle types under a single type-generic
    Warp kernel, reducing total code by ${\sim}50\%$.}
    \label{fig:unified_collision}
\end{figure}

\paragraph{Scene Collision Checking Migration to \Warp{}.}
\cuRoboVone{} split collision checking across two backends: a monolithic \CUDA{} kernel
(${\sim}2{,}600$ code lines) for cuboids and voxels, with per-type code paths, template
specializations, and manual quaternion arithmetic, and a separate \Warp{} kernel for
triangle meshes.
Four Python wrapper classes (${\sim}2{,}400$ code lines) bridged this split through a
deep inheritance hierarchy, so adding a new obstacle type required changes across every
layer and both kernel backends.
Through an LLM-assisted migration, this was replaced by a unified \Warp{} system at roughly
half the original ${\sim}5{,}000$ lines (Fig.~\ref{fig:unified_collision}).

\begin{figure}
    \centering
    \begin{minipage}[t]{\columnwidth}
        \centering
        \includegraphics{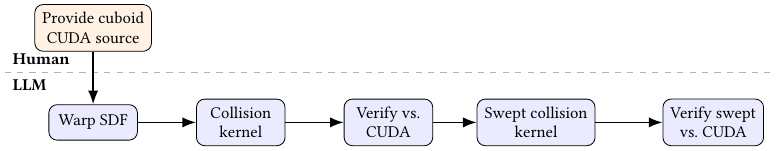}
        \subcaption{Phase~1: Cuboid, port \CUDA{} collision to \Warp{} with verification.}
        \label{fig:coll_phase1}
    \end{minipage}

    \vspace{0.6em}

    \begin{minipage}[t]{\columnwidth}
        \centering
        \includegraphics{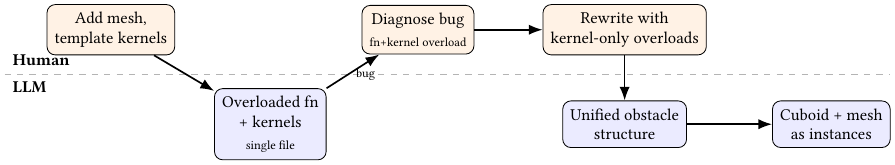}
        \subcaption{Phase~2: Generalize, add mesh support via kernel overloads.}
        \label{fig:coll_phase2}
    \end{minipage}

    \vspace{0.6em}

    \begin{minipage}[t]{\columnwidth}
        \centering
        \includegraphics{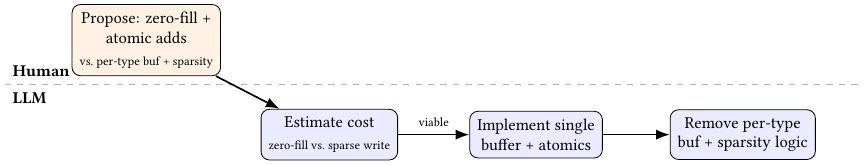}
        \subcaption{Phase~3: Simplify, replace per-type buffers with zero-fill and atomics.}
        \label{fig:coll_phase3}
    \end{minipage}

    \vspace{0.6em}

    \begin{minipage}[t]{\columnwidth}
        \centering
        \includegraphics{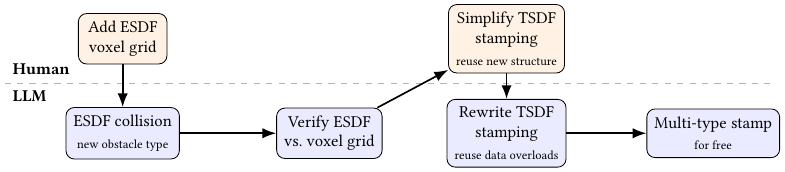}
        \subcaption{Phase~4: Extend, add ESDF voxel grid and simplify TSDF stamping.}
        \label{fig:coll_phase4}
    \end{minipage}
    \caption[Human--LLM workflow for scene collision migration]{Human--LLM workflow for migrating scene collision checking to \Warp{}.
    Orange boxes (\textcolor{orange!60}{$\blacksquare$}) denote human contributions;
    blue boxes (\textcolor{blue!30}{$\blacksquare$}) denote LLM contributions.}
    \label{fig:collision_workflow}
\end{figure}

The migration proceeded in four phases (Fig.~\ref{fig:collision_workflow}), starting
from a single cuboid type and progressively generalizing to a type-generic kernel
supporting cuboids, meshes, voxels, and ESDF grids.
Three episodes illustrate the human--LLM division of labor.
In Phase~2, the LLM generalized the cuboid kernel to support meshes using \Warp{}'s
function overloading, but the resulting code combined overloaded functions \emph{and}
overloaded kernels, triggering a silent \Warp{} caching bug.
The LLM could not diagnose the failure; we traced the root cause to a \Warp{} limitation
(either overloaded kernels or overloaded functions, but not both) and directed the LLM
to rewrite using kernel-only overloads.
In Phase~3, we proposed replacing per-type collision buffers with a single zero-filled
buffer and atomic accumulation.
The LLM evaluated the cost trade-off, implemented the change, and removed the old
buffer management logic, a case where the human supplied the design insight and the LLM
handled the mechanical refactoring.
In Phase~4, the overload-based architecture paid off: we asked the LLM to add ESDF
voxel grid support, and it implemented the new obstacle type by following the existing
overload pattern. We then asked it to simplify TSDF stamping by reusing the new
structure; the LLM rewrote the stamping logic to use the same data overloads, gaining
multi-type support for free.

\paragraph{PBA ESDF Implementation from a JFA Baseline.}
For ESDF propagation, we initially implemented Jump Flooding (JFA) because Parallel
Banding Algorithm (PBA) required nontrivial indexing and direct \CUDA{} integration
with the mapper.
After stabilizing interfaces with the JFA baseline, we asked the LLM to implement PBA
end-to-end.
The LLM generated banded indexing and kernel wiring, integrated PBA into the mapper,
and adapted it to our existing site pack/unpack representation.
Unlike common PBA implementations that assume uniform, power-of-2 grids, this version
supports non-uniform grid sizes and arbitrary dimensions.
The resulting PBA implementation is about $2\times$ faster than our JFA baseline while
producing exact signed distances.
This episode complements the RNEA and collision-migration case studies: the human
selected the staged strategy (JFA first, then PBA), while the LLM carried out low-level
implementation and system integration.

\section{Conclusion}
\label{sec:conclusion}

The central lesson of \project{} is that GPU-native computation makes it practical to enforce constraints that prior methods skip, and these constraints compound across the full stack. Fast inverse dynamics enables torque-aware B-spline optimization, keeping trajectories executable under payload where baselines fail. A dense ESDF, made tractable by decoupling TSDF and ESDF resolutions, provides $O(1)$ collision queries that accelerate planning and enable reactive MPC with live depth fusion on a real robot. Scalable kinematics and map-reduce self-collision enable collision-free motion optimization on a 48-DoF humanoid, producing physically valid reference motions whose quality propagates directly into downstream locomotion policies.
Each capability builds on the ones below it: B-splines provide a smooth trajectory space where torque constraints are well-conditioned, fast RNEA makes evaluating those constraints tractable, a dense ESDF enables real-time depth-driven collision avoidance, and scalable whole-body computation makes collision-free motion optimization practical from 7-DoF arms to 48-DoF humanoids.

Several of the modules above, including the RNEA \CUDA{} kernels, the unified scene collision system, and the PBA ESDF pipeline, were developed with substantial LLM assistance. This was only possible because the codebase was structured for discoverability: typed interfaces, predictable naming, small single-responsibility modules, and tests as executable documentation. These are standard software engineering practices, but they are necessary for LLMs in a way they are not for humans; we believe any robotics codebase that adopts them can unlock similar LLM-assisted productivity for GPU-accelerated systems.

\paragraph{Limitations \& Future Work}
Because planning and MPC share the same B-spline trajectory space, warm-starting MPC from a global plan is a natural next step toward reactive whole-body control. MPC could also serve as a safety layer that enforces collision-free, dynamically-feasible execution of learned policies.
On the perception side, our geometric robot segmentation is sensitive to depth estimation errors; learned RGB-based segmentation could improve robustness. A single camera provides only partial scene coverage; multi-camera fusion could improve reconstruction completeness.
Finally, our LLM case studies reveal that while LLMs can generate and profile GPU code, interpreting intermediate compiler representations (e.g., distinguishing virtual from physical register pressure) still requires human judgment; closing this gap is an open challenge for LLM-assisted systems development.

\paragraph{Acknowledgements}
We thank Neel Jawale for analysis of energy consumption in existing motion planners,
Ankur Handa for discussions on the challenges of human-to-robot retargeting, Erwin Coumans for discussions on real-robot perception-driven motion generation challenges,
and Dylan Turpin for building the tile-based LM solver in Warp~\citep{warp} (as part of Newton) that we adapt for inverse kinematics.
Finally, we gratefully acknowledge the software tools and libraries that made this work possible, including \PyTorch{}, \Warp{}, \Newton{}, \CUDA{} Python, \Viser{}, MimicKit, Morphit, the Cursor IDE, and Anthropic's Claude models.

\clearpage
\setcitestyle{numbers}
\bibliographystyle{plainnat}

\phantomsection
\addcontentsline{toc}{section}{References}
\bibliography{references}
\clearpage

\appendix
\section{Framework Design}
\label{sec:framework}

Beyond the algorithmic contributions presented in previous sections, \project{} introduces
substantial architectural changes to the framework itself.
The original \cuRobo{} release (hereafter \emph{v1}) was designed as a closed pipeline: users could
configure which robot to plan for, but the core optimization, collision checking, and cost
functions were monolithic and not intended for modification.
User feedback revealed that researchers wanted to add custom cost functions,
integrate new obstacle representations, and modify optimization strategies.
At the same time, onboarding a new robot required manually authoring configuration files, a
process that demanded expert knowledge of collision sphere placement and self-collision matrices.
\project{} addresses these concerns through a ground-up redesign that prioritizes
\emph{simplicity} and \emph{extensibility}, even accepting modest performance overhead where
it improves researcher experience.
This section details two areas of change: improving the researcher experience
(Sec.~\ref{sec:onboarding}) and software architecture
(Sec.~\ref{sec:code_structure}). These redesign choices also underpin the
LLM-assisted development workflow described in Sec.~\ref{sec:llm_dev}.

\subsection{Improving Researcher Experience}
\label{sec:onboarding}
A key learning from v1 was that trying a new motion-planning tool should be as
easy and quick as possible.
In practice, v1 fell short in four ways:
(1)~installation required matching \CUDA{} and \PyTorch{} versions and took roughly
    20\,minutes of compilation;
(2)~loading a new robot meant hand-authoring a configuration file, an
    error-prone process that could take another 30\,minutes even for an expert;
(3)~connecting the planner to real perception demanded additional libraries that
    did not yet exist in easy-to-use form; and
(4)~visualizing plans and debugging configurations required external tooling.
\project{} eliminates all four barriers.

\subsubsection{Simplified Installation}
\label{sec:cuda_core}

\cuRoboVone{} compiled all GPU kernels at install time through \pybindeleven{} C++ extensions.
This imposed three constraints on every user:
(1)~the Python environment's \CUDA{} toolkit version had to \emph{exactly} match the \CUDA{}
version that \PyTorch{} was built against;
(2)~the build imported \PyTorch{}'s C++ headers, resulting in compilation times of approximately
20~minutes; and
(3)~the resulting shared-object files (\code{.so}) were tied to a specific GPU architecture,
preventing portable distribution.
Taken together, these constraints made installation the single most common source of support
requests.

\begin{figure}[h]
    \centering
    \includegraphics{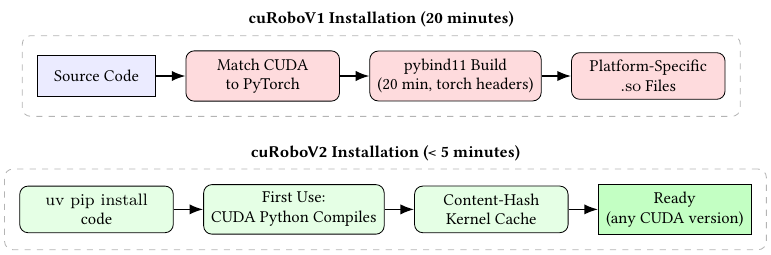}
    \caption[Installation workflow comparison]{Installation workflow comparison.
    \figloc{Top:} v1's \pybindeleven{} build required exact \CUDA{}--\PyTorch{} version matching
    and ${\sim}$20\,min of compilation.
    \figloc{Bottom:} \project{} installs via \code{pip}; kernels are compiled and
    cached on first use with no version constraints.}
    \label{fig:installation}
\end{figure}

\project{} replaces \pybindeleven{} with NVIDIA's \CUDA{} Python library~\cite{cuda-python}, which
compiles kernels at first use, automatically targets the host GPU, and caches results by
source-content hash (Fig.~\ref{fig:installation}).
Because \CUDA{} Python is fully decoupled from \PyTorch{}, the host \CUDA{} toolkit no longer needs
to match the version \PyTorch{} was compiled against, and \project{} ships as a standard
Python package (\code{pip install}), reducing installation from ${\sim}$20\,minutes to
seconds.

\subsubsection{Robot Configuration Builder}
\label{sec:robot_builder}

\begin{figure}
    \centering
    \includegraphics{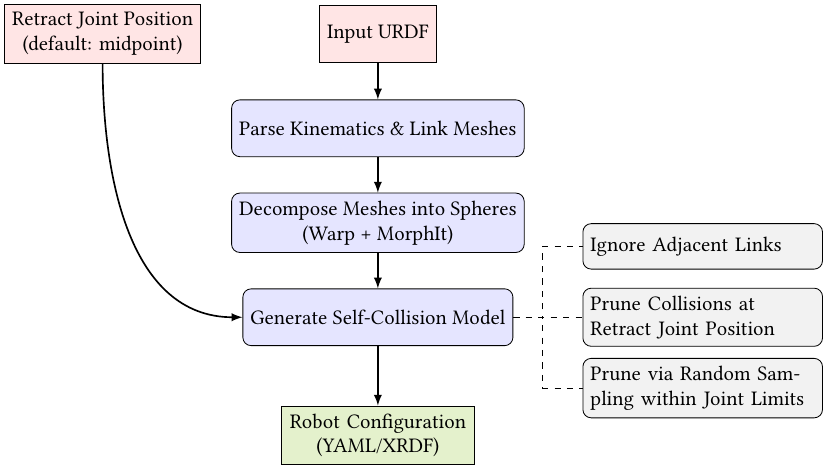}
    \caption[RobotBuilder pipeline from URDF to configuration]{The \texttt{RobotBuilder} pipeline converts a URDF into a complete, optimized
    robot configuration through four automated stages.}
    \label{fig:robot_builder}
\end{figure}

\cuRoboVone{} required users to manually author YAML configuration files specifying collision
sphere placements, self-collision ignore matrices, and joint-space parameters.
This was error-prone: there was no automated way to determine which link pairs could safely be
ignored, leading to configurations that were either overly conservative (wasting computation)
or dangerously permissive (producing unsafe trajectories).
\project{} introduces \code{RobotBuilder}, a pipeline that takes a standard URDF and produces
a complete, optimized configuration ready for planning (Fig.~\ref{fig:robot_builder}).

\paragraph{Collision Sphere Fitting.}
After parsing the URDF and identifying links with collision geometry, the builder computes a
sphere count per link from bounding-box volume scaled by a density parameter
(\code{spheres\_per\_cm}) and fits spheres using MorphIt~\cite{nechyporenko2025morphit}.
Density can be overridden per link (e.g., higher for grippers with fine features).

\paragraph{Self-Collision Matrix.}
The builder constructs the self-collision ignore matrix in three passes:
\begin{enumerate}[nosep]
    \item \textbf{Neighbor pairs:} adjacent links in the kinematic chain are unconditionally
    ignored.
    \item \textbf{Default configuration:} pairs that collide at the retract pose are ignored,
    as they represent permanent geometric overlaps that planning cannot resolve.
    \item \textbf{Sampling-based pruning:} Halton quasi-random joint configurations are
    evaluated on the GPU in large batches; pairs that \emph{never} collide are added to the
    ignore matrix.
\end{enumerate}

\paragraph{Export and Debugging.}
The final configuration can be saved as \cuRobo{} YAML or XRDF and inspected in a
web-based 3D viewer.
A load--modify--save workflow (\code{RobotBuilder.from\_config()}) and a companion
\code{RobotDebugger} class support iterative refinement without regenerating the entire
configuration.

\subsubsection{Eye-to-Hand Utilities}
\label{sec:eye_hand_utils}
Deploying a motion planner with camera-based perception requires two capabilities that
\cuRoboVone{} did not provide: calibrating the camera to the robot and filtering the robot
out of depth images before TSDF integration.

\paragraph{Eye-to-Hand Calibration}
Standard checkerboard-based calibration via OpenCV can achieve millimeter-level accuracy but
adds setup overhead.
To help users get started quickly, we provide a calibration tool inspired by
point-cloud registration methods~\cite{schmidt2015dart,huber2025hydra}.
A global ICP stage mean-centers both point clouds, evaluates a batch of candidate orientations
on the GPU with point-to-plane matching between points sampled on the robot mesh and a
user-specified bounding box in the scene, and selects the best alignment.
A local refinement step then minimizes signed-distance residuals similar to
DART~\cite{schmidt2015dart}.
The entire workflow is integrated into a web-based visualizer: the user supplies the
current joint state, adjusts a bounding box, and saves the calibrated pose to disk.
This approach proves sufficient in practice for collision-free motion planning and pick-and-place tasks.

\begin{figure*}
    \centering
    \includegraphics[width=0.99\linewidth]{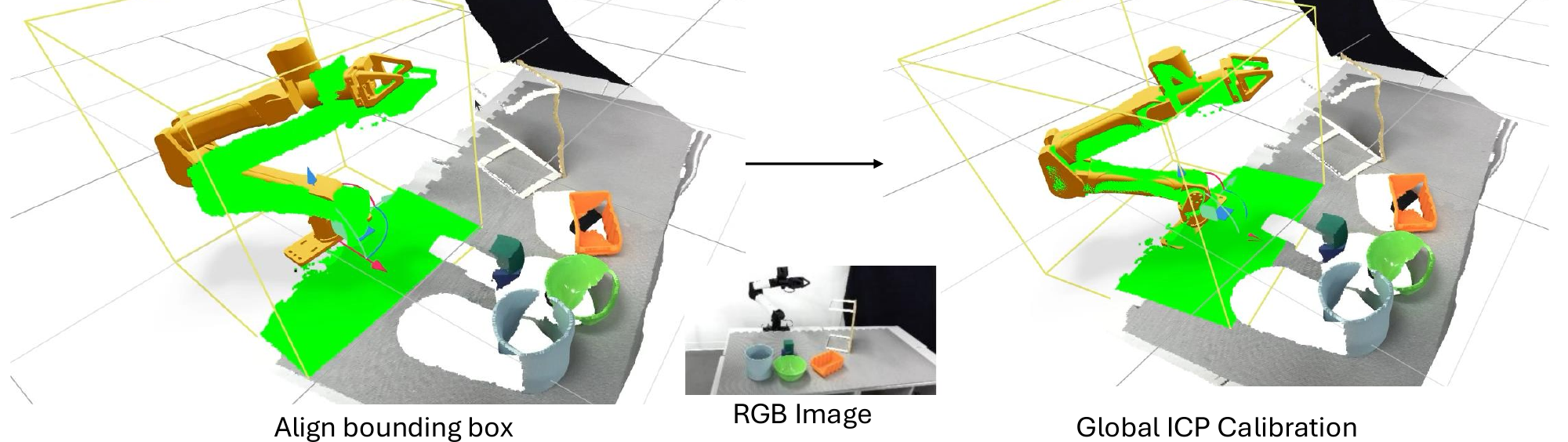}
    \caption[Camera-to-robot calibration via ICP]{Camera-to-robot calibration via ICP. The user specifies a bounding box in the scene (left), which is used to run global ICP, aligning the robot mesh to the observed point cloud (right).}
    \label{fig:calibration}
\end{figure*}

\paragraph{Robot Segmentation}
To remove the robot from depth images, we use our forward kinematics to position the collision spheres
at the current joint state, project depth pixels to 3D using the calibrated extrinsics and
camera intrinsics, and mask any point whose $\ell_2$ distance to the nearest sphere falls
within its radius.
This is approximate compared to mesh-based segmentation but significantly faster, providing a
practical default for TSDF-based scene reconstruction.

\subsubsection{Web Visualization via \Viser{}}
\label{sec:viser}

\cuRoboVone{} had no built-in visualization, requiring external tooling to inspect plans or
debug configurations.
\project{} integrates \Viser{}~\cite{yi2025viser}, a lightweight web-based 3D visualizer that
renders the robot URDF, provides interactive transform controls for IK/MPC targets, overlays
collision spheres and scene obstacles, and visualizes TSDF/ESDF
reconstructions, all accessible from any browser.

\subsection{Software Architecture}
\label{sec:code_structure}

We describe two new extension points in \project{} that let users add custom obstacle
types (Sec.~\ref{sec:unified_collision}) and cost functions
(Sec.~\ref{sec:cost_manager}) without modifying core code.
Cross-cutting design changes for codebase discoverability and the quantitative impact of
the restructuring are discussed in Sec.~\ref{sec:llm_dev}.

\subsubsection{Unified Scene Collision Checking}
\label{sec:unified_collision}

As described in Sec.~\ref{sec:llm_dev}, \cuRoboVone{}'s collision checking was split across
two kernel backends and a deep inheritance hierarchy
(Fig.~\ref{fig:unified_collision}, left).
\project{} consolidates all scene collision checking into a single \emph{type-generic \Warp{}~\cite{warp}
kernel} (Fig.~\ref{fig:unified_collision}, right).
Each obstacle type (cuboid, mesh, voxel) registers three function overloads into a shared
module:
\code{is\_obs\_enabled} (whether the obstacle is active),
\code{load\_obstacle\_transform} (world-frame pose), and
\code{compute\_local\_sdf\_with\_grad} (signed distance and analytic gradient in the
obstacle's local frame).
The unified kernel parallelizes over (sphere~$\times$~obstacle) pairs, calls the appropriate
overloads based on obstacle type, and accumulates results through atomic adds into a single
\code{CollisionBuffer}.
Adding a new obstacle representation (a point cloud, a neural SDF, or any user-defined
geometry) requires only implementing the three interface functions.
No existing kernel code needs to be modified.

\subsubsection{Cost Manager \& \Warp{}-First Kernels}
\label{sec:cost_manager}

\begin{figure}[h]
    \centering
    \includegraphics{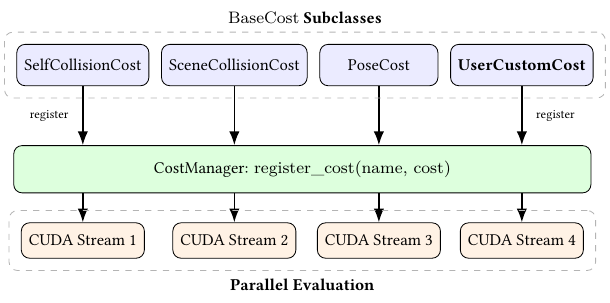}
    \caption[Composable cost manager architecture]{Composable cost manager.
    Each \texttt{BaseCost} subclass is registered by name and evaluated on its own \CUDA{} stream.
    Bold box indicates a user-defined custom cost.}
    \label{fig:cost_manager}
\end{figure}

\cuRoboVone{} computed all cost terms inside a monolithic rollout class, making it difficult to add
or modify individual costs.
\project{} introduces a \code{CostManager} (Fig.~\ref{fig:cost_manager}) where each cost is
a standalone \code{BaseCost} subclass registered by name.
Each subclass implements \code{forward(state, config)} and returns a cost tensor; gradients
flow automatically via \PyTorch{} backpropagation, and the manager evaluates all registered
costs in parallel on separate \CUDA{} streams.
Adding a new cost term (e.g., workspace constraints, learned cost functions) requires only
writing a new \code{BaseCost} subclass; no changes to the optimization pipeline or existing
costs.

The split between \CUDA{} C++ and \Warp{} is driven by \emph{stability} versus
\emph{modifiability}.
Forward kinematics, L-BFGS optimization, B-spline evaluation, and self-collision are mature
algorithms that change infrequently; these remain as hand-tuned \CUDA{} C++ kernels compiled via
the \CUDA{} Python backend.
Scene collision, pose distance, and swept-sphere collision are cost functions users may want
to modify; these are implemented as \Warp{} kernels.
In particular, \code{PoseCost}, the most commonly requested customization, was moved from
\cuRoboVone{}'s \CUDA{} C++ to \Warp{} so that users can read and edit it directly.
\Warp{}'s Python-like syntax and built-in automatic differentiation mean that custom or modified
costs require no C++ expertise and no recompilation.

\section{LLM Prompts}
\label{sec:prompt_rules}
Sec.~\ref{sec:discoverability} described the codebase properties that enable productive
LLM-assisted development.
To enforce these properties consistently, we authored a set of \emph{rule files}, i.e., persistent
system prompts loaded into every LLM session, that encode our design principles, formatting
conventions, file naming rules, and documentation standards.
Each rule file is shown below in its entirety.
Together, they ensure that LLM-generated code adheres to the same structure, naming, and
documentation standards as human-written code, reducing review friction and preventing
stylistic drift over time.

\begin{tcolorbox}[colback=lightgreen,colframe=lightgreenline,title=\textbf{\textcolor{black}{Design Principles (\texttt{design\_principles.mdc})}},breakable,enhanced]
You are an expert programmer that follow the below principles very strictly:

\begin{multicols}{2}
\footnotesize
\noindent \textbf{Software Development Principles}
\begin{enumerate}[leftmargin=2em]
\item \textbf{Don't Repeat Yourself (DRY)}: Avoid code duplication; use reusable components.
\item \textbf{Keep It Simple, Stupid (KISS)}: Keep designs simple and avoid complexity.
\item \textbf{You Aren't Gonna Need It (YAGNI)}: Implement only necessary features.
\item \textbf{Separation of Concerns}: Divide code into distinct sections.
\item \textbf{Law of Demeter}: Interact only with immediate dependencies.
\item \textbf{Composition Over Inheritance}: Prefer composition for code reuse.
\item \textbf{Encapsulation}: Hide internal state and expose only necessary interfaces.
\item \textbf{Cohesion}: Ensure related elements work together.
\item \textbf{Low Coupling}: Minimize dependencies between modules.
\item \textbf{Design by Contract}: Define clear interface specifications.
\item \textbf{Test-Driven Development (TDD)}: Write tests before code.
\item \textbf{Continuous Integration (CI)}: Integrate code changes frequently with automated tests.
\item \textbf{Continuous Deployment (CD)}: Automate deployment for quick releases.
\item \textbf{Agile Principles}: Use iterative development and collaboration.
\item \textbf{Clean Code}: Write readable and maintainable code.
\item \textbf{Refactoring}: Continuously improve code design.
\end{enumerate}

\vspace{0.5em}
\noindent \textbf{SOLID Principles}
\begin{enumerate}[leftmargin=2em]
\setcounter{enumi}{16}
\item \textbf{Single Responsibility Principle (SRP)}: One class, one responsibility.
\item \textbf{Open/Closed Principle (OCP)}: Open for extension, closed for modification.
\item \textbf{Liskov Substitution Principle (LSP)}: Subtypes must be substitutable for their base types.
\item \textbf{Interface Segregation Principle (ISP)}: Specific interfaces for different clients.
\item \textbf{Dependency Inversion Principle (DIP)}: Depend on abstractions, not concretions.
\end{enumerate}

\vspace{0.5em}
\textbf{Domain-Driven Design (DDD) Principles}
\begin{enumerate}[leftmargin=2em]
\setcounter{enumi}{21}
\item \textbf{Domain}: Model the business context.
\item \textbf{Entities}: Objects with distinct identities.
\item \textbf{Value Objects}: Objects with attributes but no identity.
\item \textbf{Aggregates}: Cluster of objects treated as a unit.
\item \textbf{Repositories}: Encapsulate storage and retrieval.
\item \textbf{Services}: Perform domain logic operations.
\item \textbf{Bounded Contexts}: Define explicit boundaries for models.
\item \textbf{Ubiquitous Language}: Use a common language for clear communication.
\end{enumerate}

\vspace{0.5em}
\textbf{Additional Principles}
\begin{enumerate}[leftmargin=2em]
\setcounter{enumi}{29}
\item \textbf{Optimize for Performance}: Use profiling tools to identify bottlenecks.
\item \textbf{Use Virtual Environments}: Isolate project dependencies.
\item \textbf{Document Your Code}: Write clear docstrings and comments.
\item \textbf{Leverage Libraries and Frameworks}: Use existing tools to speed up development.
\item \textbf{Keep Learning}: Stay updated with the latest advancements.
\item \textbf{Contribute to the Community}: Engage with and contribute to the community.
\end{enumerate}

\vspace{0.5em}
\textbf{Formatting:}
\begin{enumerate}[leftmargin=2em]
\setcounter{enumi}{35}
\item Follow @package\_format\_rules.mdc when writing code.
\item Follow @documenter.mdc when writing docstrings.
\item Follow @file\_naming.mdc when creating new files.
\end{enumerate}
\end{multicols}
\end{tcolorbox}
\clearpage

\begin{tcolorbox}[colback=lightgreen,colframe=lightgreenline,title=\textbf{\textcolor{black}{Package Format Rules (\texttt{package\_format\_rules.mdc})}},breakable,enhanced]
    \noindent \textbf{Docstrings}
    \begin{enumerate}[leftmargin=2em]
    \item follow flake8 rules from @setup.cfg
    \item Always docstring in the same line as \texttt{'''}
    \item Use :func:, :class:,  :attr: for references
    \item Write shape of inputs and outputs when possible
    \item For dataclass, always document attributes as they are declared. Do not document in the class
    docstring.
    \end{enumerate}

    \noindent \textbf{Python}
    \begin{enumerate}[leftmargin=2em]
    \item Follow flake8 rules from @setup.cfg
    \item Use descriptive names for functions, classess and other members
    \item Use type hints for all function signatures
    \item Use logging functions from @logging.py for print and raising
    warnigs, errors
    \item Avoid using deprecated functions
    \item Follow PEP 563 (Postponed Evaluation of Annotations)
    \end{enumerate}

    \noindent \textbf{Writing Tests}
    \begin{enumerate}[leftmargin=2em]
    \item tests should be src/curobo/tests folder.
    \item Do not write unit tests for wp.kernel decorators.
    \item When writing gradient checking using gradcheck, use float32 and not float64.
    \end{enumerate}
    \end{tcolorbox}

\begin{tcolorbox}[colback=lightgreen,colframe=lightgreenline,title=\textbf{\textcolor{black}{File Naming Convention (\texttt{file\_naming.mdc})}},breakable,enhanced]
    \footnotesize

    \noindent \textbf{Category-First File Naming Convention}

    \vspace{2mm}
    When organizing code in this project, strictly follow these category-first file naming conventions:

    \vspace{2mm}
    \noindent \textbf{File Naming Rules}

    \begin{enumerate}[leftmargin=2em]
    \item \textbf{Always use the format \texttt{category\_specific.py}} where:
       \begin{itemize}[leftmargin=1em]
       \item \texttt{category} identifies the broader component type or subsystem
       \item \texttt{specific} describes the particular implementation or variant
       \end{itemize}

    \item \textbf{Examples of correct file names}:
       \begin{itemize}[leftmargin=1em]
       \item \texttt{connector\_linear.py} (NOT \texttt{linear\_connector.py})
       \item \texttt{sampler\_node.py} (NOT \texttt{node\_sampler.py})
       \end{itemize}

    \item \textbf{Multiple adjectives} should appear in order from general to specific:
       \begin{itemize}[leftmargin=1em]
       \item \texttt{connector\_cubic\_spline.py} (NOT \texttt{connector\_spline\_cubic.py})
       \end{itemize}

    \item \textbf{Module directory structure} should follow the same categorical organization:
    \begin{itemize}[leftmargin=1em]
    \item \texttt{/connectors/connector\_linear.py}
    \end{itemize}
    \item \textbf{Import statements} should reflect this structure:
    \begin{itemize}[leftmargin=1em]
    \item \texttt{from curobo.connectors.connector\_linear import LinearConnector}
    \end{itemize}
    \end{enumerate}

    \noindent \textbf{Class Naming}

     Class names should remain in standard CamelCase format with specific-type first:
    \begin{itemize}[leftmargin=1.5em]
    \item File \texttt{connector\_linear.py} contains class \texttt{LinearConnector}
    \end{itemize}

    When refactoring existing code or creating new files, always audit and ensure compliance with this category-first naming convention. Consistency across the codebase is essential.
\end{tcolorbox}
\clearpage
\begin{tcolorbox}[colback=lightgreen,colframe=lightgreenline,title=\textbf{\textcolor{black}{Documentation Principles (\texttt{documenter.mdc})}},breakable,enhanced]
You write concise and clear documentation that strictly follows the below principles:
\begin{multicols}{2}
\footnotesize
\noindent \textbf{Understand the Audience:}

Tailor the documentation and tutorials to the knowledge level of new users. Provide clear explanations of domain concepts, APIs, and implementations.

\vspace{0.5em}
\noindent \textbf{Use Clear and Concise Language:}

Write in a clear, concise, and straightforward manner. Avoid jargon and complex language unless necessary, and provide definitions for any technical terms used.

\vspace{0.5em}
\noindent \textbf{Organize Content Logically:}

Structure the documentation with a logical flow. Start with an introduction to domain concepts, followed by API overviews, and then detailed implementation guides.

\vspace{0.5em}
\noindent \textbf{Follow reST and Sphinx Conventions:}

Adhere to reStructuredText syntax and Sphinx conventions for formatting. Use appropriate directives and roles for code blocks, references, and other elements.

\vspace{0.5em}
\noindent \textbf{Provide Comprehensive Tutorials:}

Create step-by-step tutorials that guide users through common tasks and use cases. Include code examples and explanations to help users understand how to use the framework.

\vspace{0.5em}
\noindent \textbf{Document Domain Concepts:}

Explain key domain concepts in detail. Use diagrams (using graphviz), examples, and analogies to make complex ideas more accessible.

\vspace{0.5em}
\noindent \textbf{Include API References:}

Provide comprehensive API documentation. Include descriptions of classes, methods, functions, parameters, return values, and exceptions. Use autodoc to generate API references from docstrings.

\vspace{0.5em}
\noindent \textbf{Explain Implementations:}

Offer insights into the implementation details of the framework. Explain the design decisions, algorithms, and data structures used.

\vspace{0.5em}
\noindent \textbf{Use Examples and Code Snippets:}

Include examples and code snippets to illustrate how to use the framework. Ensure that examples are relevant, well-documented, and easy to understand.

\vspace{0.5em}
\noindent \textbf{Highlight Best Practices:}

Provide guidance on best practices for using the framework. Include tips on performance optimization, error handling, and common pitfalls to avoid.

\vspace{0.5em}
\noindent \textbf{Ensure Consistency:}

Maintain consistency in terminology, formatting, and style throughout the documentation. Use a style guide to ensure uniformity.

\vspace{0.5em}
\noindent \textbf{Use Cross-References:}

Use cross-references to link related sections, tutorials, and API documentation. This helps users navigate the documentation easily.

\vspace{0.5em}
\noindent \textbf{Include Installation and Setup Instructions:}

Provide clear instructions for installing and setting up the framework. Include information on dependencies, virtual environments, and any required configurations.

\vspace{0.5em}
\noindent \textbf{Provide Troubleshooting Guides:}

Include troubleshooting guides to help users resolve common issues. Offer solutions to common errors and problems they may encounter.

\vspace{0.5em}
\noindent \textbf{Keep Documentation Up-to-Date:}

Regularly update the documentation to reflect changes in the framework. Ensure that tutorials, examples, and API references are current and accurate.

\vspace{0.5em}
\noindent \textbf{Encourage User Feedback:}

Encourage users to provide feedback on the documentation. Use this feedback to improve and refine the content.

\vspace{0.5em}
\noindent \textbf{Formatting:}

Stick to 99 as the line width

For graphviz, use style from @graphviz\_example.rst
\end{multicols}
\end{tcolorbox}

\end{document}